%% file: main.tex
\RequirePackage{fix-cm}
\documentclass[twocolumn,numbook]{svjour3}          
\smartqed  

\input{preload}

\usepackage{subfiles}
\usepackage{xr}

\journalname{A preprint}

\begin{document}

\hyphenation{segmentation-based}
\hyphenation{classification-based}
\hyphenation{data-sets}

\title{Intra- \& Extra-Source Exemplar-Based Style Synthesis for Improved Domain Generalization
}


\author{
    Yumeng Li \and
    Dan Zhang \and
    Margret Keuper \and
    Anna Khoreva
}

\institute{
Yumeng Li \at
Bosch Center for Artificial Intelligence \& University of Siegen \\
\email{yumeng.li@bosch.com}
\and
Dan Zhang \at
Bosch Center for Artificial Intelligence \& University of T\"ubingen\\
\email{dan.zhang2@bosch.com}
\and
Margret Keuper \at
University of Siegen \& Max Planck Institute for Informatics \\
\email{margret.keuper@uni-siegen.de}
\and
Anna Khoreva \at
Bosch Center for Artificial Intelligence \& University of T\"ubingen\\
\email{anna.khoreva@bosch.com}
}

\date{} 

\maketitle
\input{tex/0_abstract}

\input{tex/1_introduction}

\input{tex/2_related_work}

\input{tex/3_methods}

\input{tex/4_experiments}

\input{tex/5_conclusion}

\bibliographystyle{spbasic}   
\bibliography{references}

\end{document}

%% file: preload.tex
\usepackage[pagebackref=false,breaklinks=true]{hyperref}
\usepackage{amsmath,amssymb}
\usepackage[capitalize]{cleveref} 
\usepackage{cite}
\usepackage{breakcites}
\usepackage{microtype} 
\usepackage{url}
\usepackage{bbm} 
\usepackage{caption}
\usepackage{booktabs}
\usepackage{siunitx}
\usepackage{comment} 
\usepackage{pifont}%
\usepackage{makecell}
\usepackage{multirow}
\usepackage{times}
\usepackage{placeins}
\usepackage{colortbl}
\usepackage{rotating}
\usepackage{array}
\usepackage{color}

\usepackage[pdftex,dvipsnames]{xcolor}  
\usepackage{verbatim}
\usepackage{textcomp}
\usepackage{bbding}
\usepackage{mwe}
\usepackage{graphicx}
\usepackage{wrapfig}
\usepackage{subcaption}
\usepackage{listings}  
\usepackage{xr}
\usepackage{natbib}
\usepackage[misc]{ifsym}
\usepackage{nameref}

\PassOptionsToPackage{normalem}{ulem}
\usepackage{ulem}
\definecolor{lightgray}{gray}{0.4}
\definecolor{verylightgray}{gray}{0.7}
\definecolor{veryverylightgray}{gray}{0.9}

\definecolor{darkgreen}{rgb}{0, 0.4, 0}
\definecolor{darkred}{rgb}{0.7, 0, 0}
\definecolor{darkblue}{rgb}{0.0, 0.0, 0.7}

\definecolor{matplotlibBlue}{HTML}{1F77B4}
\definecolor{matplotlibOrange}{HTML}{FF7F0E}

\usepackage{xstring}
\usepackage{ifdraft}
\usepackage{lipsum}  
\usepackage{pifont} 
\usepackage[abs]{overpic}

\usepackage{tikz,pgf}
\usetikzlibrary{calc}
\usetikzlibrary{shapes,backgrounds}
\usepackage{float}

\crefname{section}{Sec.}{Secs.}
\Crefname{section}{Section}{Sections}
\Crefname{table}{Table}{Tabs.}
\crefname{table}{Table}{Tabs.}

\usepackage{chngcntr}
\counterwithout{figure}{section}
\counterwithout{table}{section}

\newcommand{\YL}[1]{[\textcolor{blue}{YL: #1}]}

\definecolor{RWTHblue}{RGB}{0,84,159}
\definecolor{RWTHbluedark}{RGB}{0,57,159}
\definecolor{purplemiddle}{RGB}{128,0,128}
\newcommand{\newchange}[1]{\textcolor{black}{#1}}

\renewcommand{\paragraph}[2][.]{\vspace{6pt}\noindent{\bf #2#1}}
\newcommand{\xmark}{\ding{55}}
\newcommand{\cmark}{\ding{51}}

\newcommand*{\newcite}[1]{~\citep{#1}}

\newcommand{\norm}[1]{\left\lVert#1\right\rVert}
\usepackage{pifont}
\usepackage{enumitem}

\usepackage{fontawesome} 

\newcommand{\hrnet}{\newcite{wang2020hrnet}}
\newcommand{\segformer}{\newcite{xie2021segformer}}

\newcommand{\acdc}{\newcite{sakaridis2021acdc}}
\newcommand{\ourstyle}{$\mathrm{ISSA}$}
\newcommand{\ourstylebf}{{$\mathbf{ISSA}$}}

\newcommand{\wplus}{$\mathcal{W^+}$}

\graphicspath{
{figs}{figs/semseg}
{figs/encoder_comparison}
{figs/intra_mix_bdd}
{figs/masking_ablation_2x2}
{figs/noise_vis}
{figs/compare_stylemix}
{figs/noise_map_rec}
{figs/landscape}
{figs/correlation}
{figs/interpolation}
{figs/stylized_proxy}
}

\usepackage{xspace}
\makeatletter
\DeclareRobustCommand\onedot{\futurelet\@let@token\@onedot}
\def\@onedot{\ifx\@let@token.\else.\null\fi\xspace}

\def\etal{\emph{et al}\onedot}

\def\@setref#1#2#3{%
  \ifx#1\relax
   \protect\G@refundefinedtrue
   \nfss@text{\reset@font\colorbox{yellow}{??$_\text{#3}$}}
   \@latex@warning{Reference `#3' on page \thepage \space
             undefined}%
  \else
   \expandafter#2#1\null
  \fi}
\makeatother


%% file: tex/0_abstract.tex
\begin{abstract}
The generalization with respect to domain shifts, as they frequently appear in applications such as autonomous driving, is one of the remaining big challenges for deep learning models. \newchange{Therefore, we propose an exemplar-based style synthesis pipeline  to improve domain generalization in semantic segmentation.
} Our method is based on a novel masked noise encoder for StyleGAN2 inversion. The model learns to faithfully reconstruct the image, preserving its semantic layout through noise prediction. Random masking of the estimated noise enables the style mixing capability of our model, i.e.~it allows to alter the global appearance without affecting the semantic layout of an image. Using the proposed masked noise encoder to randomize style and content combinations in the training set, i.e., intra-source style augmentation ({\ourstyle}) effectively increases the diversity of training data and reduces spurious correlation. 
As a result, we achieve up to $12.4\%$ mIoU improvements on driving-scene semantic segmentation under different types of data shifts, i.e., changing geographic locations, adverse weather conditions, and day to night. {\ourstyle} is model-agnostic and straightforwardly applicable with CNNs and Transformers. 
It is also complementary to other domain generalization techniques, e.g., it improves the recent state-of-the-art solution RobustNet by $3\%$ mIoU in Cityscapes to Dark Z\"urich.
In addition, we demonstrate the strong plug-n-play ability of the proposed style synthesis pipeline, which is readily usable for extra-source exemplars e.g., web-crawled images, without any retraining or fine-tuning. Moreover, we study a new use case to indicate neural network's generalization capability by building a stylized proxy validation set. This application has significant practical sense for selecting models to be deployed in the open-world environment. Our code is available at \url{https://github.com/boschresearch/ISSA}.
\keywords{
Domain Generalization \and GAN Inversion \and Data Augmentation \and Semantic Segmentation
}
\end{abstract}

%% file: tex/1_introduction.tex
\vspace{1em}
\section{Introduction}
\input{figs/intro_semseg}

The varying environment with potentially diverse illumination and adverse weather conditions makes challenging the deployment of deep learning models in an open-world\newcite{sakaridis2021acdc,zhang2021autonomous}.  
Therefore, improving the generalization capability of neural networks is crucial for safety-critical applications such as autonomous driving (see for example \cref{fig:intro-semseg}). While generally the target domains can be inaccessible or unpredictable at training time, it is important to train a generalizable model, based on the known (source) domain, which may offer only a limited or biased view of the real world\newcite{burton2017safety,shafaei2018uncertainty}.

Diversity of the training data is considered to play an important role for domain generalization, including natural distribution shifts\newcite{taori2020robustness}. Many existing works assume that multiple source domains are accessible during training\newcite{hu2020mda,li2018learning,balaji2018metareg,li2018domain,li2020domain,jin2020feature,zhou2020optimaltransport}. For instance, 
Li~\etal\newcite{li2018learning} applied meta-learning to better generalize to unseen domains, where source domains are divided into meta-source and meta-target domains to simulate domain shift; 
Hu~\etal\newcite{hu2020mda} propose multi-domain discriminant analysis to learn a domain-invariant feature transformation.
However, for pixel-level prediction tasks such as semantic segmentation, collecting diverse training data involves a tedious and costly annotation process\newcite{caesar2018coco}. Therefore, improving \newchange{and predicting} generalization from a \emph{single source domain} is exceptionally compelling, particularly for semantic segmentation.

One pragmatic way to improve data diversity is by applying data augmentation. It has been widely adopted in solving different tasks, such as image classification\newcite{zhang2018mixup,zhou2021mixstyle,hendrycks2019augmix,verma2019manifold,hong2021stylemix}, GAN training with limited data\newcite{stylegan2ada,jiang2021deceivegan}, or pose estimation\newcite{peng2018humanpose,bin2020adv_pose,wang2021human}. 
One line of data augmentation techniques focuses on increasing the content diversity in the training set, such as geometric transformation (e.g., cropping or flipping), CutOut\newcite{devries2017cutout}, and CutMix\newcite{yun2019cutmix}. However, CutOut and CutMix are ineffective on natural domain shifts, as reported in\newcite{taori2020robustness}. Style augmentation, on the other hand, only modifies the style - the non-semantic appearance such as texture and color of the image\newcite{gatys2016image} - while preserving the semantic content. 
By diversifying the style and content combinations, style augmentation can reduce overfitting to the style-content correlation in the training set, improving robustness against domain shifts. 
Hendrycks corruptions\newcite{hendrycks2018benchmarking} provide a wide range of synthetic styles, including weather conditions. However, they are not always realistic looking, thus being still far from resembling natural data shifts.
\newchange{In this work, we propose an exemplar-based style synthesis pipeline for semantic segmentation, aiming to improve the style diversity in the training and validation set without extra labeling effort.
} 

Our exemplar-based style synthesis technique is based on the inversion of StyleGAN2\newcite{stylegan2}, which is the state-of-the-art unconditional Generative Adversarial Network (GAN) and thus ensures high quality and realism of synthetic samples. 
GAN inversion allows encoding a given image to latent variables, and thus facilitates faithful reconstruction with style mixing capability. To realize the synthesis pipeline, we learn to separate semantic content from style information based on a single source domain. This allows to alter the style of an image while leaving the content unchanged. \newchange{In particular, we focus on intra-source style augmentation ({\ourstyle}). Namely, our exemplar-based style synthesis makes use of training samples from the source domain, extracting their styles and contents followed by randomly mixing them up.} %
In doing so, we can increase the data diversity and alleviate the spurious correlation in the given training data.

The faithful reconstruction of images with complex structures such as driving scenes is non-trivial. Prior methods\newcite{pspencoder,feature_style,roich2021pti,alaluf2022hyperstyle,dinh2022hyperinverter,vsubrtova2022chunkygan} are mainly tested on simple single-object-centric datasets, e.g., FFHQ\newcite{stylegan}, CelebA-HQ\newcite{progressivegan},  or LSUN\newcite{yu2015lsun}. 
As shown in \newcite{abdal2020image2stylegan++}, extending the native latent space of StyleGAN2 with a stochastic noise space can lead to improved inversion quality. However, all style \emph{and} content information will be embedded in the noise map, leaving the latent codes inactive in this setting.
Therefore, to enable the precise reconstruction of complex driving scenes as well as style mixing, we propose a masked noise encoder for StyleGAN2. The proposed random masking regularization on the noise map encourages the generator to rely on the latent prediction for reconstruction. Thus, it allows to effectively separate content and style information and facilitates realistic style mixing, as shown in \cref{fig:encoder-visual}.

\newchange{ 
We further discover an excellent plug-n-play ability of 
the proposed style synthesis pipeline, i.e., it can be directly applied to unseen domains without requiring the re-training of the encoder or generator. For instance, in \cref{fig:landscape-example}, we employ our pipeline directly on web-crawled images, where the model is only trained on Cityscapes. This appealing property opens up the opportunity to go beyond intra-source exemplar-based style mixing, and grants us more flexibility to harness extra-source data for style synthesis. %
Thus, we also experiment with extra-source style argumentation (ESSA) to further improve the generalization performance. }

\newchange{Besides data augmentation, we explore the usage of the proposed pipeline for assessing neural networks' generalization capability in \cref{sec:new_correlation}. By transferring styles from unannotated data samples of the target domain to existing labelled data, we can build a style-augmented proxy set for validation without introducing extra-labelling effort. We observe that performance on this proxy set has a strong correlation with the real test performance on unseen target data, which could be used in practice to select more suitable models for deployment.
}
 
In summary, we make the following contributions:
\begin{itemize}
    \item We propose a masked noise encoder for GAN inversion, which enables high quality reconstruction and style mixing of complex scene-centric datasets. 
    \item We exploit GAN inversion for intra-source data augmentation, which can improve generalization under natural distribution shifts on semantic segmentation. 
    \item Extensive experiments demonstrate that our proposed augmentation method {\ourstyle} consistently promotes domain generalization performance on driving-scene semantic segmentation across different network architectures, achieving up to $12.4\%$ mIoU improvement, even with limited diversity in the source data and without access to the target domain.  
    \item \newchange{We discover the plug-n-play ability of our masked noise encoder, and showcase its potential of direct application on extra-source data such as web-crawled images. 
    }
    \item \newchange{We further explore the usage of the proposed pipeline for assessing models' generalization performance on unseen data. By building a style-augmented proxy validation set on known labelled data, we observe that there is a strong correlation between the performance on the proxy validation set and the real test set, which offers useful insights for model selection without introducing any extra annotation effort.
    }
\end{itemize}
\newchange{
This paper is an extended version of our previous work \newcite{li2023issa} with more experimental evaluation and discussion on the potential and two new applications of the proposed method.
In particular, we provide a more detailed ablation study on the design of the proposed masked noise encoder (see \cref{tab:new-gan-inversion-mask-ablation,tab:new-noise-map},\cref{fig:new-noise-res-comp2}).
Furthermore, we add a discussion on the plug-n-play ability of the pipeline and go beyond intra-source domain to extra-source domain style mixing. We also conducted new experiments reported in \cref{tab:bddgan-cityscapes-dg-small,tab:ladnscape-dg}. Finally, the new application as model generalization performance indicator is introduced in \cref{sec:new_correlation}.
}

%% file: figs/intro_semseg.tex
\begin{figure}[t]
    \begin{centering}
    \setlength{\tabcolsep}{0.0em}
    \renewcommand{\arraystretch}{0}
    \par\end{centering}
    \begin{centering}
    \hfill{}
	\begin{tabular}{@{}c@{}c}
        \centering
		Unseen domain (snow) & Ground truth \tabularnewline
	    \includegraphics[width=0.47\linewidth]{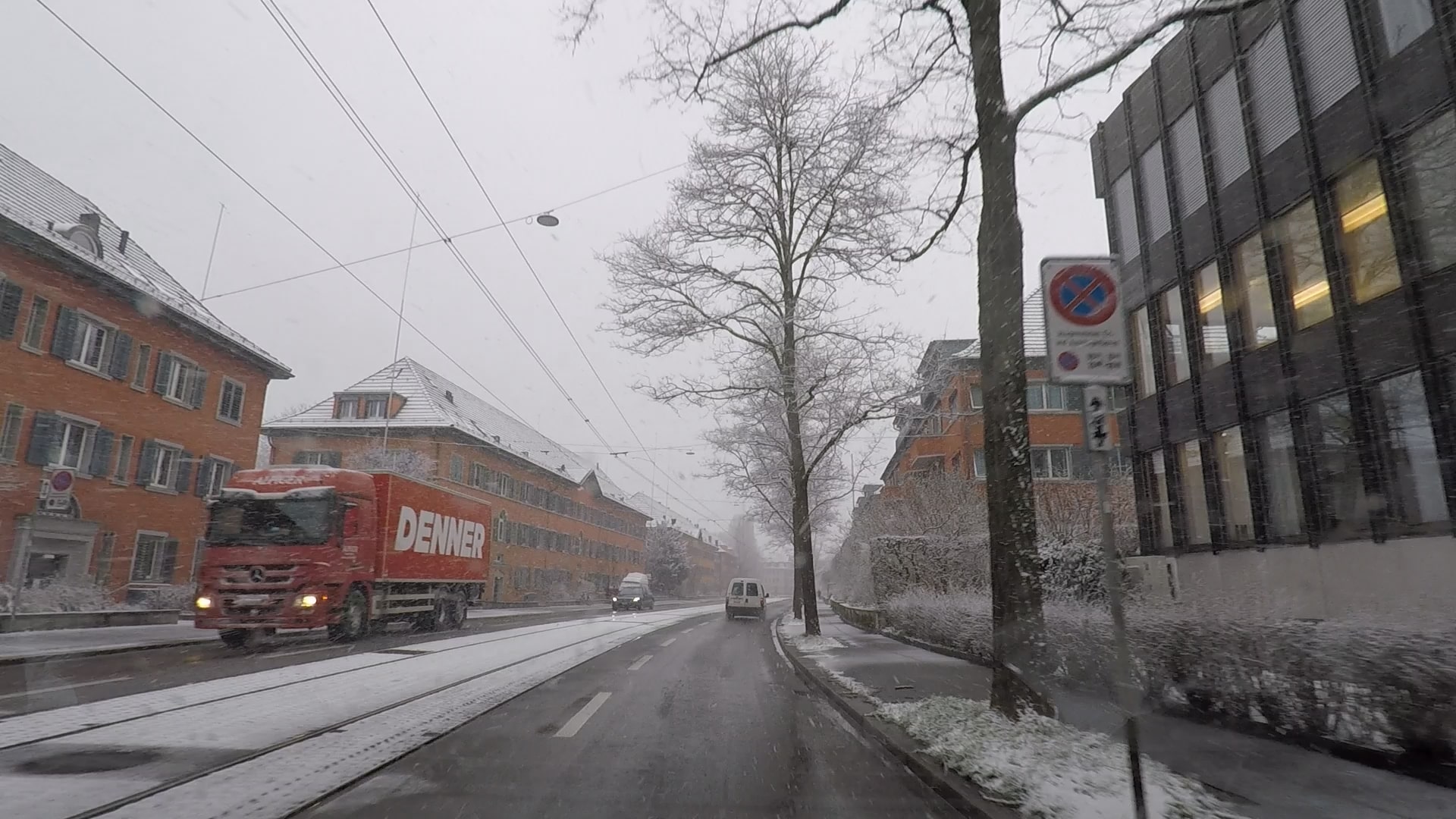} 
		&{\footnotesize{}} 
		\begin{tikzpicture}
            \node [
	        above right,
	        inner sep=0] (image) at (0,0) {\includegraphics[width=0.47\linewidth]{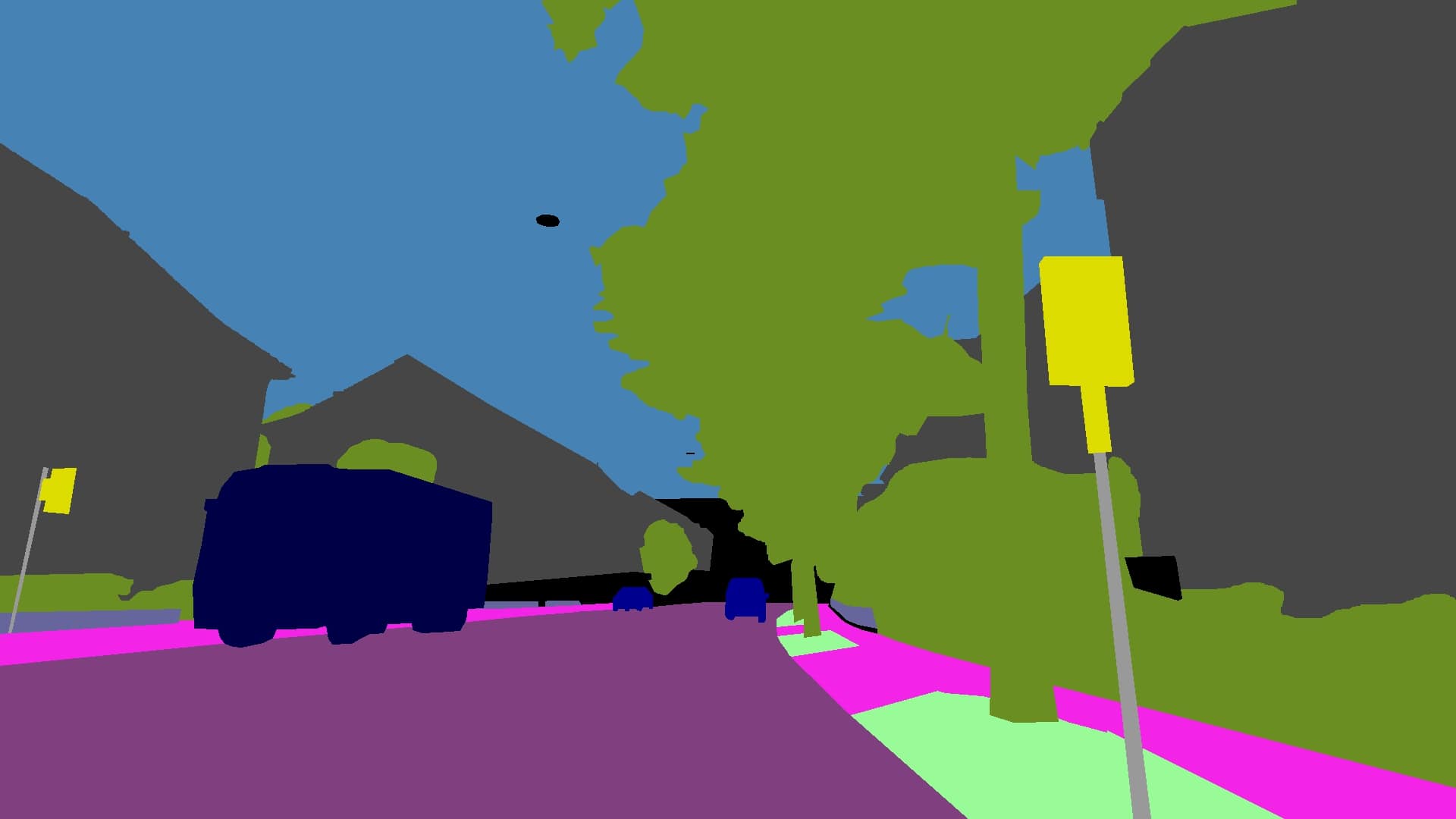} };
            \begin{scope}[
            x={($0.1*(image.south east)$)},
            y={($0.1*(image.north west)$)}]
            \draw[thick,green] (7,0.5) rectangle (9.8,7.2);
            \draw[thick,green] (1.1,2) rectangle (3.6,4.6);
        \end{scope}
    \end{tikzpicture}
		\tabularnewline
		Baseline & Ours \tabularnewline
	\begin{tikzpicture}
            \node [
	        above right,
	        inner sep=0] (image) at (0,0) {\includegraphics[width=0.47\linewidth]{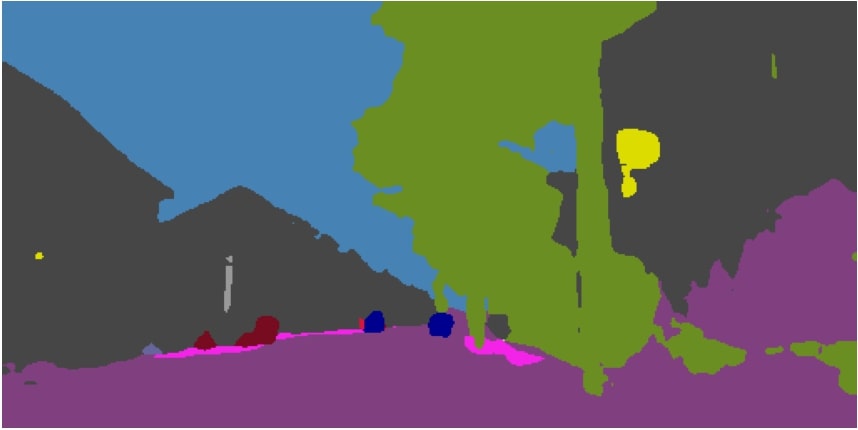}};
            \begin{scope}[
            x={($0.1*(image.south east)$)},
            y={($0.1*(image.north west)$)}]
            \draw[thick,red] (7,0.5) rectangle (9.8,7.2);
            \draw[thick,red] (1.1,2) rectangle (3.6,4.6);
        \end{scope}
    \end{tikzpicture} 
		&{\footnotesize{}} 
    \begin{tikzpicture}
            \node [
	        above right,
	        inner sep=0] (image) at (0,0) {\includegraphics[width=0.47\linewidth]{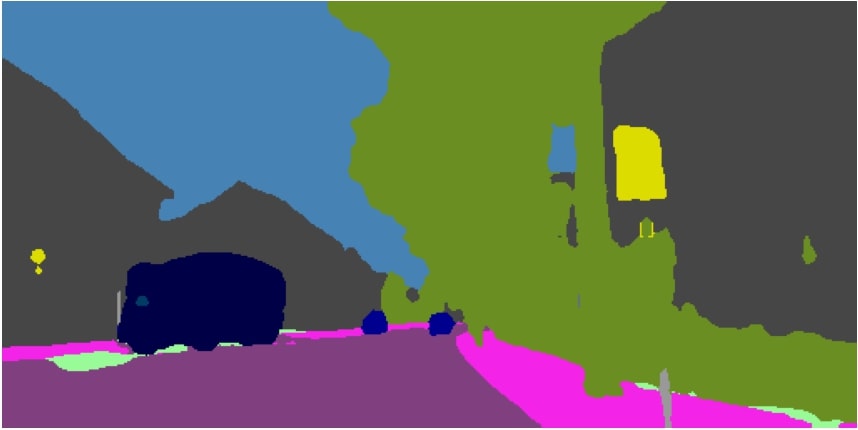}};
            \begin{scope}[
            x={($0.1*(image.south east)$)},
            y={($0.1*(image.north west)$)}]
            \draw[thick,green] (7,0.5) rectangle (9.8,7.2);
            \draw[thick,green] (1.1,1.5) rectangle (3.6,4.6);
        \end{scope}
    \end{tikzpicture}	
		\end{tabular}
\hfill{}
\par\end{centering}
\caption{
Semantic segmentation results of HRNet{\hrnet} on unseen domain (snow), trained on Cityscapes\newcite{cordts2016cityscapes} and tested on ACDC\newcite{sakaridis2021acdc}. The model trained with our {\ourstyle} can successfully segment the truck, while the baseline model fails completely.}
\label{fig:intro-semseg}
\end{figure}

%% file: tex/2_related_work.tex
\input{figs/encoder_comparison_supp} 

\section{Related Work}\label{sec:related-work}

\paragraph{Domain Generalization}
Domain generalization concerns the generalization ability of neural networks to a target domain that follows a different distribution than the source domain, and prior knowledge of the target domain is inaccessible at training. 
Various methods have been proposed to approach this problem from different angles, which employ data augmentation\newcite{khirodkar2019domain,somavarapu2020frustratingly,huang2021fsdr,zhou2021mixstyle,li2022dsu}, domain alignment\newcite{hu2020mda,li2020domain,jin2020feature,zhou2020optimaltransport}, 
adversarial training\newcite{li2018domain,multiad2019,RahmanFBS20adv,deng2020representation},
meta-learning\newcite{li2018learning,balaji2018metareg,Li_Episodic_2019,Zhao_Sebe_2021}, ensemble learning\newcite{d2018domain,Mancini_2018, Wu_Gong_2021,lee2022cross}, 
\newchange{ or feature decomposition\newcite{Wan_2022_metacnn,chen2022compound}.}  Particularly, \newcite{qiao2020single,wang2021single,jia2020single,ouyang2022causality} focus on single domain generalization problem.
While the majority focuses on image-level tasks, e.g., image classification or person re-identification, a few recent works \newcite{robustnet_2021,wildnet_2022_CVPR,kim2021wedge,kim2022pin,shade-style-hall} investigate pixel-level prediction tasks such as semantic segmentation. RobustNet\newcite{robustnet_2021} proposes an instance selective whitening loss to the instance normalization, aiming to selectively remove information that causes a domain shift while maintaining discriminative features. \newcite{kim2022pin} introduces a memory-guided meta-learning framework to capture co-occurring categorical knowledge across domains. \newcite{wildnet_2022_CVPR,kim2021wedge} make use of extra data in the wild for feature augmentation.
\newchange{
SHADE\newcite{shade-style-hall} proposed to use a style consistency constraint to learn a style-invariant representation and a retrospection consistency constraint to leverage knowledge from the pretrained backbone. To assist the training, they perturb features to simulate style variations.
}

Another line of work explores feature-level augmentation \newcite{zhou2021mixstyle,li2022dsu}. MixStyle\newcite{zhou2021mixstyle} and DSU \newcite{li2022dsu} add perturbation at the normalization layer to simulate domain shifts at test time. However, this perturbation can potentially cause a distortion of the image content, which can be harmful for semantic segmentation (see \cref{sec:exp_dg}). Moreover, these methods require a careful adaptation to the specific network architecture. 
In contrast, {\ourstyle} performs style mixing on the image-level, thus being model-agnostic, and can be applied as a complement to other methods in order to further increase the generalization performance.

\newchange{
Beyond data augmentation for improving domain generalization, we further explore the usage of our  exemplar-based style synthesis pipeline for assessing the generalization performance. Recently,  \newcite{zhang2021predictingGAN} proposed to predict generalization of image classifiers using performance on synthetic  data produced by a conditional GAN. While this is limited to the generalization in the source domain, and it is not straightforward how to apply it on semantic segmentation task. In contrast to generating image from scratch,  
we employ proposed exemplar-based style synthesis pipeline to augment labelled source data and build a stylized proxy validation sets.
We empirically show that such proxy validation sets can indicate generalization performance, without extra annotation required.
}

\paragraph{Data Augmentation}
Data augmentation techniques can diversify training samples by altering their style, content, or both, thus preventing overfitting and improving generalization.
Mixup augmentations\newcite{zhang2018mixup,dabouei2021supermix,verma2019manifold}  
linearly interpolate between two training samples and their labels, regularizing both style and content. Despite effectiveness shown on image-level classification tasks, they are not well suited for dense pixel-level prediction tasks. 
CutMix\newcite{yun2019cutmix} cuts and pastes a random rectangular region of the input image into another image, thus increasing the content diversity. Geometric transformation, e.g., random scaling and horizontal flipping, can also serve this purpose. 
In contrast, Hendrycks corruptions\newcite{hendrycks2018benchmarking} only affect the image appearance without modifying the content.
Their generated images look artificial, being far from resembling natural data, and thus offer limited help against natural distribution shifts\newcite{taori2020robustness}. 

StyleMix\newcite{hong2021stylemix} is conceptually closer to our method, which aims to decompose training images into content and style representations and then mix them up to generate more samples. Nonetheless, their AdaIN\newcite{huang2017adain} based style mixing method 
cannot fulfill the pixel-wise label-preserving requirement (see \cref{fig:stylemix-sample-1}). 
Another line of CycleGAN based style transfer methods\newcite{hoffman2018cycada,voreiter2020cycle}  require the access to both source and target domain during training, and thus cannot be employed for domain generalization problem.
Our {\ourstyle} is also a style-based data augmentation technique that leverages the capabilities of a state-of-the-art GAN to produce natural looking samples.  By modifying solely the style of the input images and maintaining their content intact, the original ground truth label maps can be reused. Furthermore, this model can be effectively trained on a single domain without necessitating target data.

\paragraph{GAN Inversion}
Showing good results, GAN inversion has been explored for many applications such as face editing \newcite{abdal2019image2stylegan,abdal2020image2stylegan++,zhu2020indomain}, image restoration\newcite{pan2021exploiting}, and data augmentation\newcite{face_da,golhar2022gan}. 
StyleGANs\newcite{stylegan, stylegan2, stylegan2ada} are commonly used for inversion, as they demonstrate high synthesis quality and appealing editing capabilities. 
Nevertheless, there is a known distortion-editability trade-off\newcite{tov2021e4e}. Thus, it is crucial to achieve a curated performance for a specific use case.  

GAN inversion approaches can be classified into three groups: optimization based methods\newcite{creswell2018inverting,abdal2019image2stylegan,abdal2020image2stylegan++,gu2020image,kang2021BDInvert,collins2020editing}, encoder based models\newcite{pspencoder,feature_style,bartz2020onenoise,tov2021e4e,wei2022e2style} methods, and hybrid approaches\newcite{dinh2022hyperinverter,roich2021pti,alaluf2022hyperstyle,chai2021ensembling,song2022editing}. 
Optimization methods generally have worse editability and need exhaustive optimization for each input. Thus,
in this paper, we use an encoder based method for our style mixing purpose.  
The representative encoder based work pSp encoder\newcite{pspencoder} 
embeds the input image in the extended latent space $\mathcal{W}^+$ of StyleGAN. 
\newchange{
The e4e encoder\newcite{tov2021e4e} improves editability of pSp while trading off detail preservation.
Yet, for the semantic segmentation augmentation task, it is crucial to assure the pixel-wise alignment with ground-truth label maps. 
To improve the reconstruction quality, the Feature-Style encoder\newcite{feature_style} further replaces the lower latent code prediction with a feature map prediction.
Recent works explored the usage of additional information such as labelled regions of interest\newcite{moon2022interestyle} and segment masks\newcite{vsubrtova2022chunkygan}, or involved the joint optimization of the generator\newcite{roich2021pti,hu2022invgan_gcpr}. Our method only requires RGB images and a frozen generator, meanwhile offers plug-n-play ability on web-crawled images (see \cref{sec:new_pnp}).
}

Despite much progress, most prior work only showcases applications on single object-centric datasets, such as CelebA-HQ\newcite{progressivegan}, FFHQ\newcite{stylegan},  LSUN\newcite{yu2015lsun}.
They still fail on more complex scenes, thus restricting their application in practice.
Our masked noise encoder can fulfil both the fidelity and the style mixing capability requirements, rendering itself well-suited for data augmentation for semantic segmentation. To the best of our knowledge, our approach is the first GAN inversion method which can be effectively applied as data augmentation for the semantic segmentation of complex scenes.

%% file: figs/encoder_comparison_supp.tex
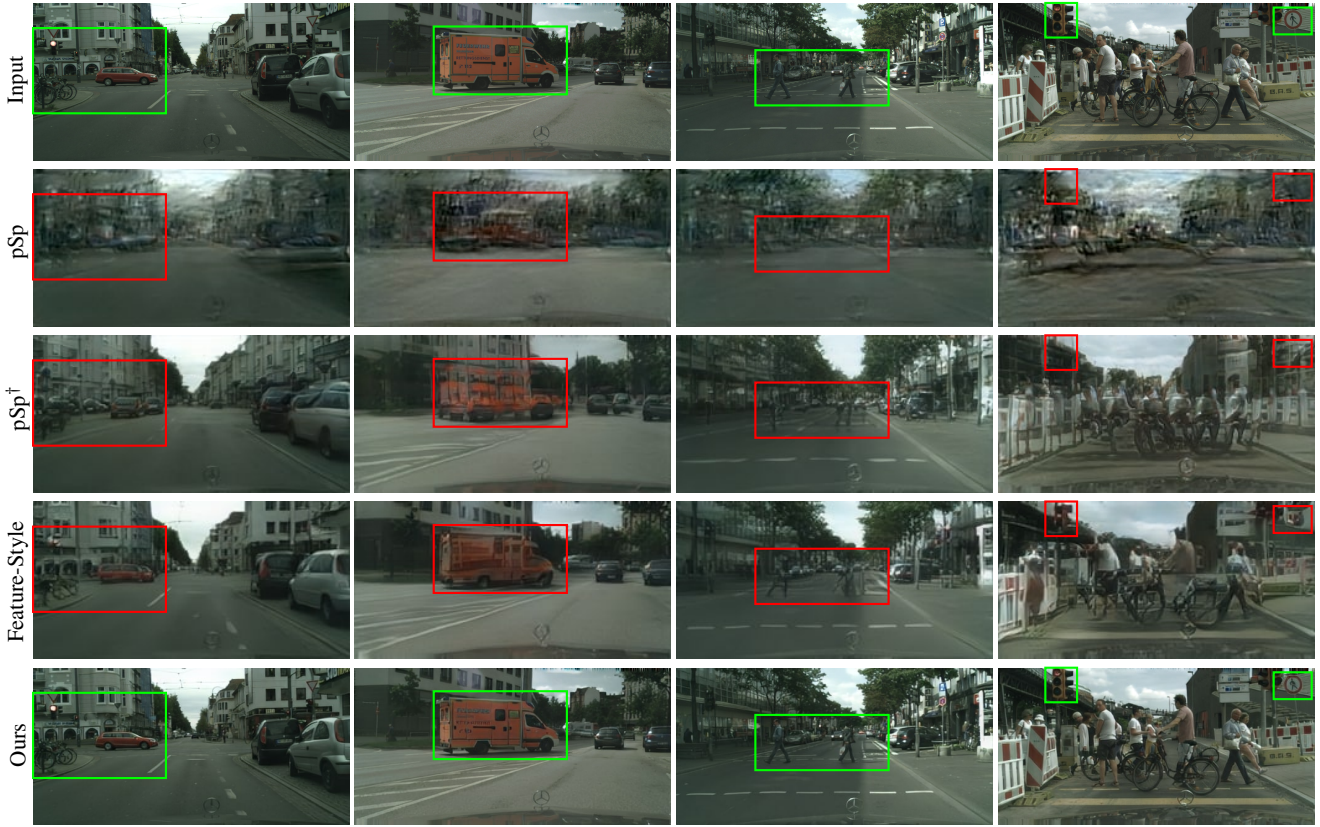
\begin{figure*}[t]
\begin{centering}
\setlength{\tabcolsep}{0.0em}
\renewcommand{\arraystretch}{0}
\par\end{centering}
\begin{centering}
\hfill{}
	\begin{tabular}{@{\hspace{-0.3em}}c@{\hspace{0.1em}}c@{}c@{}c@{}c}
			\centering
		&   &   &   &\vspace{0.01cm} \tabularnewline
		
	\multirow{1}{*}{ \rotatebox{90}{\hspace{3.5em}  Input \hspace{-4.3em} }} &	    
	\begin{tikzpicture}
            \node [
	        above right,
	        inner sep=0] (image) at (0,0) {\includegraphics[width=0.239\textwidth,]{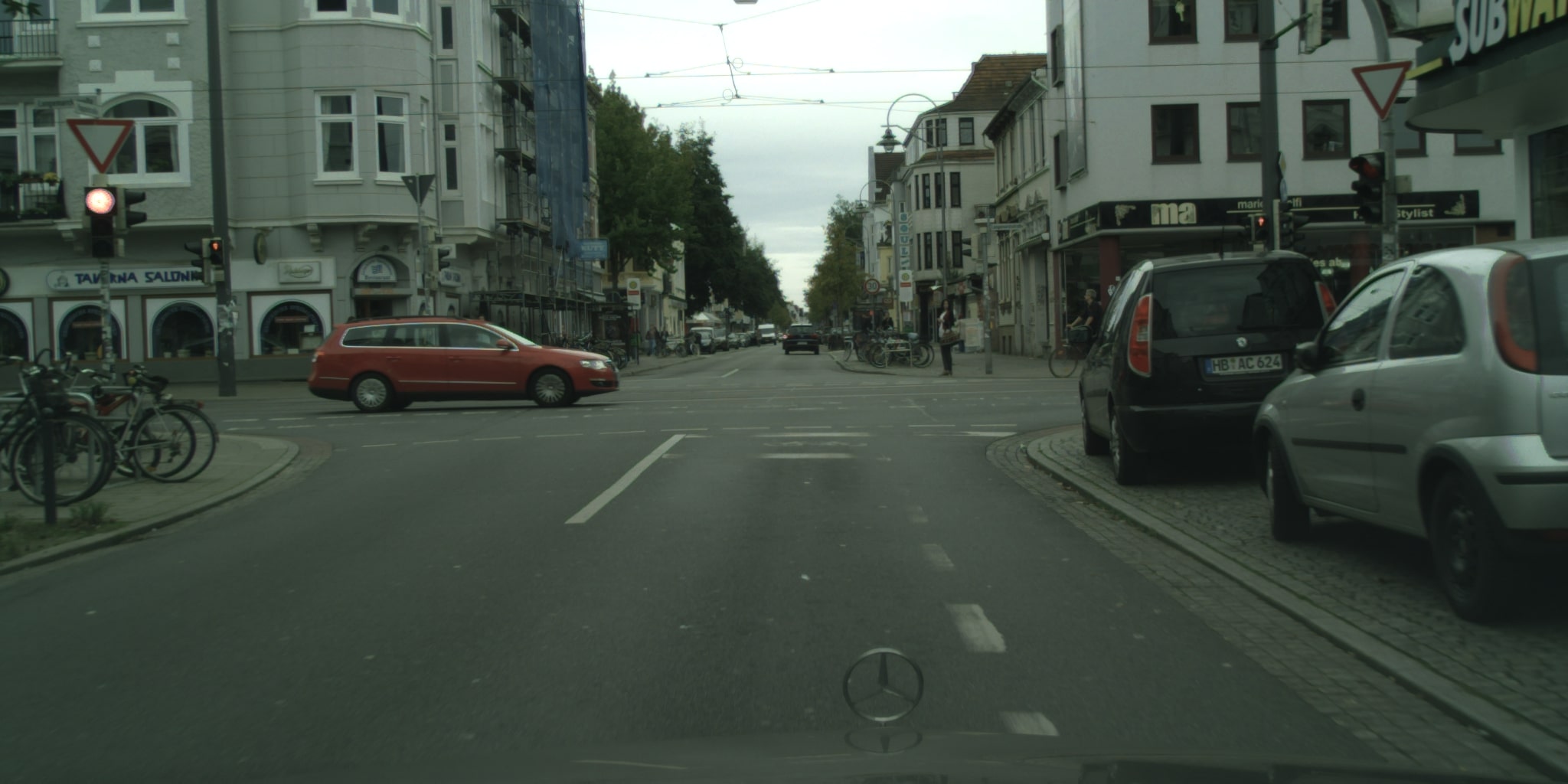} };
            \begin{scope}[
            x={($0.1*(image.south east)$)},
            y={($0.1*(image.north west)$)}]
            \draw[thick,green] (0.01,3) rectangle (4.2,8.4) ;
        \end{scope}
    \end{tikzpicture}
	        & {\footnotesize{}}
	 \begin{tikzpicture}
            \node [
	        above right,
	        inner sep=0] (image) at (0,0) {\includegraphics[width=0.239\textwidth,]{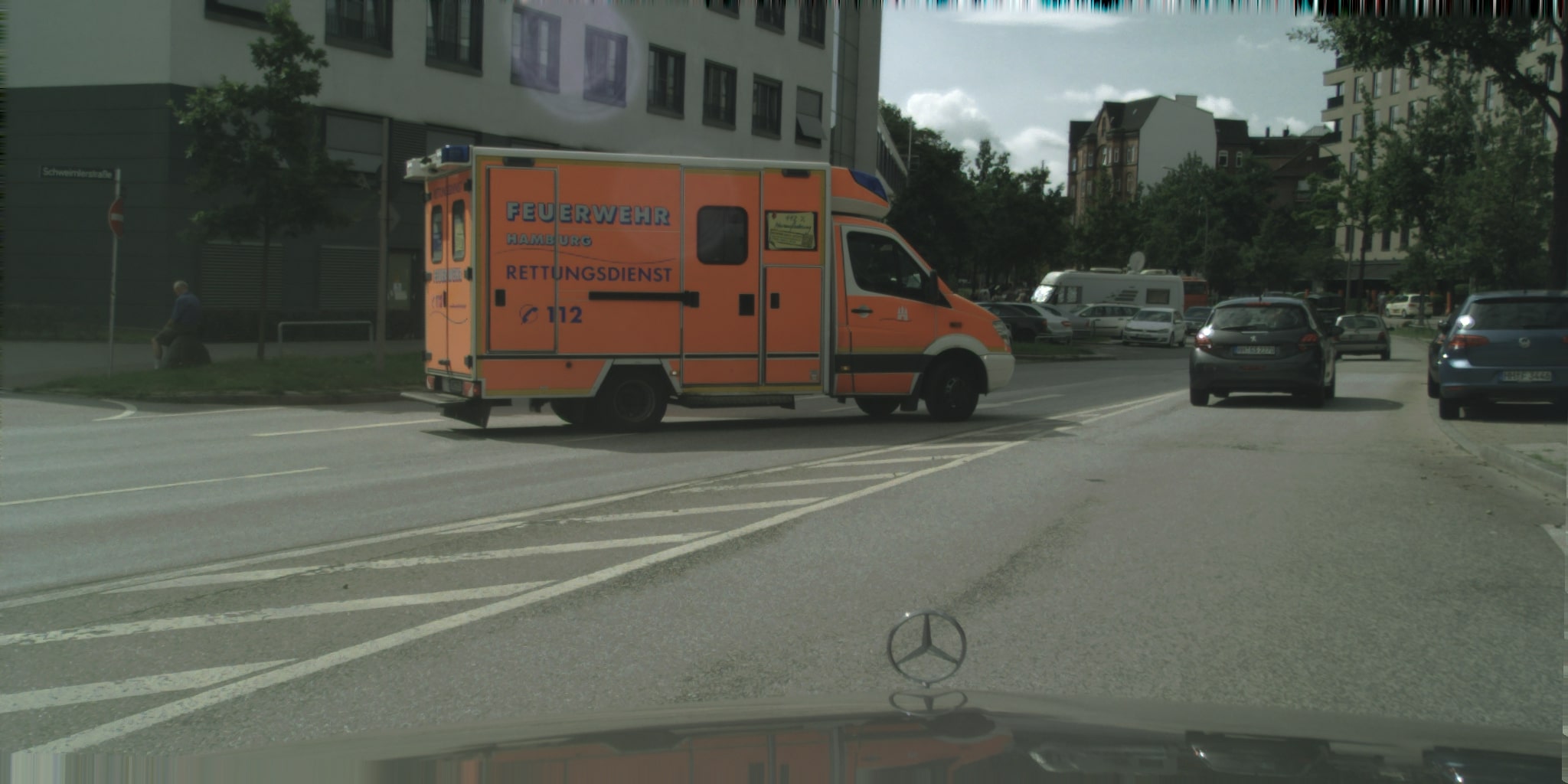}  };
            \begin{scope}[
            x={($0.1*(image.south east)$)},
            y={($0.1*(image.north west)$)}]
            \draw[thick,green] (2.5,4.2) rectangle (6.7,8.5) ;
        \end{scope}
    \end{tikzpicture}       
	        & {\footnotesize{}}
	        
	 \begin{tikzpicture}
            \node [
	        above right,
	        inner sep=0] (image) at (0,0) {\includegraphics[width=0.239\textwidth,]{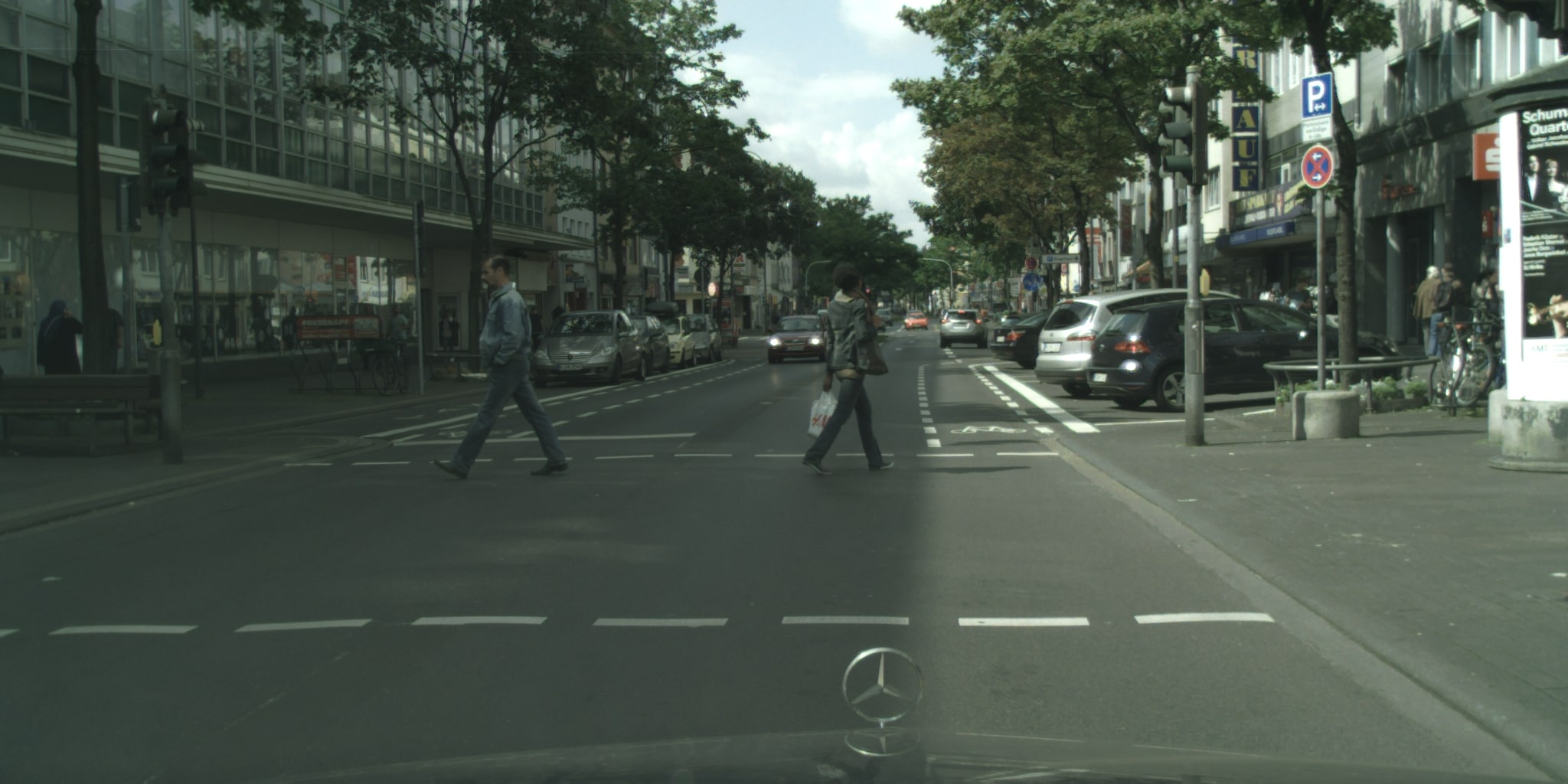}  };
            \begin{scope}[
            x={($0.1*(image.south east)$)},
            y={($0.1*(image.north west)$)}]
            \draw[thick,green] (2.5,3.5) rectangle (6.7,7) ;
        \end{scope}
    \end{tikzpicture}   
		    & {\footnotesize{}}
		    
    \begin{tikzpicture}
            \node [
	        above right,
	        inner sep=0] (image) at (0,0) {\includegraphics[width=0.239\textwidth,]{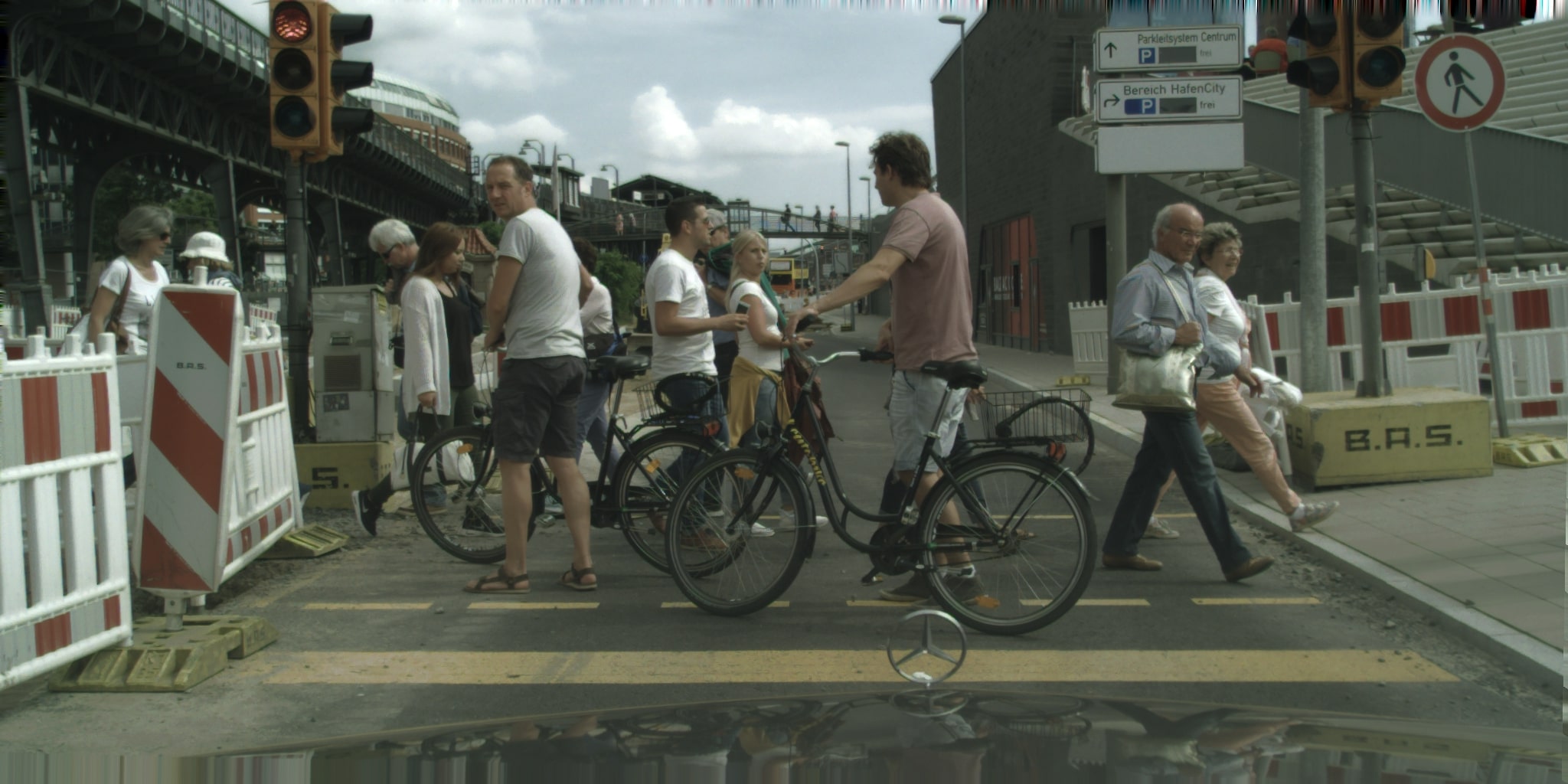}  };
            \begin{scope}[
            x={($0.1*(image.south east)$)},
            y={($0.1*(image.north west)$)}]
            \draw[thick,green] (1.5,7.8) rectangle (2.5,9.99) ;
            \draw[thick,green] (8.7,8) rectangle (9.9,9.7) ;
        \end{scope}
    \end{tikzpicture} 
	 \tabularnewline	
	 
	 \multirow{1}{*}{ \rotatebox{90}{\hspace{3.5em}  pSp \hspace{-4.2em} }} &	 
	 \begin{tikzpicture}
            \node [
	        above right,
	        inner sep=0] (image) at (0,0) {\includegraphics[width=0.239\textwidth,]{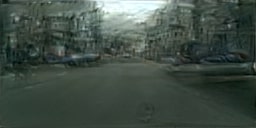}  };
            \begin{scope}[
            x={($0.1*(image.south east)$)},
            y={($0.1*(image.north west)$)}]
            \draw[thick,red] (0.01,3) rectangle (4.2,8.4) ;
        \end{scope}
    \end{tikzpicture}
	        & {\footnotesize{}}
\begin{tikzpicture}
            \node [
	        above right,
	        inner sep=0] (image) at (0,0) {\includegraphics[width=0.239\textwidth,]{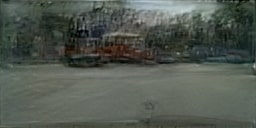}  };
            \begin{scope}[
            x={($0.1*(image.south east)$)},
            y={($0.1*(image.north west)$)}]
            \draw[thick,red] (2.5,4.2) rectangle (6.7,8.5) ;
        \end{scope}
    \end{tikzpicture}	  
	        & {\footnotesize{}}
\begin{tikzpicture}
            \node [
	        above right,
	        inner sep=0] (image) at (0,0) {\includegraphics[width=0.239\textwidth,]{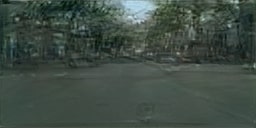} };
            \begin{scope}[
            x={($0.1*(image.south east)$)},
            y={($0.1*(image.north west)$)}]
            \draw[thick,red] (2.5,3.5) rectangle (6.7,7) ;
        \end{scope}
    \end{tikzpicture}
		    & {\footnotesize{}}
    \begin{tikzpicture}
            \node [
	        above right,
	        inner sep=0] (image) at (0,0) {		\includegraphics[width=0.239\textwidth,]{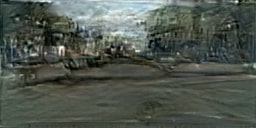}};
            \begin{scope}[
            x={($0.1*(image.south east)$)},
            y={($0.1*(image.north west)$)}]
            \draw[thick,red] (1.5,7.8) rectangle (2.5,9.99) ;
            \draw[thick,red] (8.7,8) rectangle (9.9,9.7) ;
        \end{scope}
    \end{tikzpicture}
	 \tabularnewline
	
	\multirow{1}{*}{ \rotatebox{90}{\hspace{3.5em}  pSp${}^\dagger$ \hspace{-4.5em} }} &
	\begin{tikzpicture}
            \node [
	        above right,
	        inner sep=0] (image) at (0,0) {	\includegraphics[width=0.239\textwidth,]{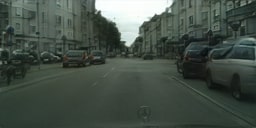} };
            \begin{scope}[
            x={($0.1*(image.south east)$)},
            y={($0.1*(image.north west)$)}]
            \draw[thick,red] (0.01,3) rectangle (4.2,8.4) ;
        \end{scope}
    \end{tikzpicture}
	        & {\footnotesize{}}
	  \begin{tikzpicture}
            \node [
	        above right,
	        inner sep=0] (image) at (0,0) {\includegraphics[width=0.239\textwidth,]{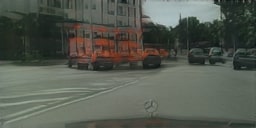} };
            \begin{scope}[
            x={($0.1*(image.south east)$)},
            y={($0.1*(image.north west)$)}]
            \draw[thick,red] (2.5,4.2) rectangle (6.7,8.5) ;
        \end{scope}
    \end{tikzpicture}
	        & {\footnotesize{}}
	 \begin{tikzpicture}
            \node [
	        above right,
	        inner sep=0] (image) at (0,0) {\includegraphics[width=0.239\textwidth,]{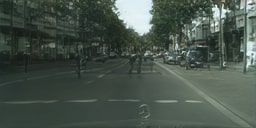}};
            \begin{scope}[
            x={($0.1*(image.south east)$)},
            y={($0.1*(image.north west)$)}]
            \draw[thick,red] (2.5,3.5) rectangle (6.7,7) ;
        \end{scope}
    \end{tikzpicture}
		    & {\footnotesize{}}
	\begin{tikzpicture}
            \node [
	        above right,
	        inner sep=0] (image) at (0,0) {	\includegraphics[width=0.239\textwidth,]{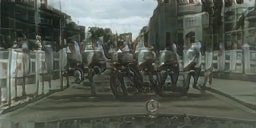}};
            \begin{scope}[
            x={($0.1*(image.south east)$)},
            y={($0.1*(image.north west)$)}]
            \draw[thick,red] (1.5,7.8) rectangle (2.5,9.99) ;
            \draw[thick,red] (8.7,8) rectangle (9.9,9.7) ;
        \end{scope}
    \end{tikzpicture}
	
	 \tabularnewline
		
	\multirow{1}{*}{ \rotatebox{90}{\hspace{3.5em}  \small{Feature-Style} \hspace{-5.8em} }} &
	\begin{tikzpicture}
            \node [
	        above right,
	        inner sep=0] (image) at (0,0) {\includegraphics[width=0.239\textwidth,]{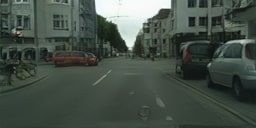}   };
            \begin{scope}[
            x={($0.1*(image.south east)$)},
            y={($0.1*(image.north west)$)}]
            \draw[thick,red] (0.01,3) rectangle (4.2,8.4) ;
        \end{scope}
    \end{tikzpicture}
	
	        & {\footnotesize{}}
	  \begin{tikzpicture}
            \node [
	        above right,
	        inner sep=0] (image) at (0,0) {\includegraphics[width=0.239\textwidth,]{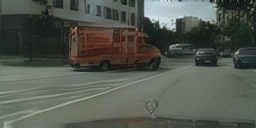} };
            \begin{scope}[
            x={($0.1*(image.south east)$)},
            y={($0.1*(image.north west)$)}]
            \draw[thick,red] (2.5,4.2) rectangle (6.7,8.5) ;
        \end{scope}
    \end{tikzpicture}
	        & {\footnotesize{}}
	 \begin{tikzpicture}
            \node [
	        above right,
	        inner sep=0] (image) at (0,0) {\includegraphics[width=0.239\textwidth,]{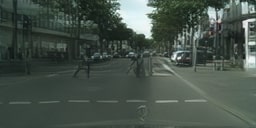}  };
            \begin{scope}[
            x={($0.1*(image.south east)$)},
            y={($0.1*(image.north west)$)}]
            \draw[thick,red] (2.5,3.5) rectangle (6.7,7) ;
        \end{scope}
    \end{tikzpicture}
		    & {\footnotesize{}}
	\begin{tikzpicture}
            \node [
	        above right,
	        inner sep=0] (image) at (0,0) {\includegraphics[width=0.239\textwidth,]{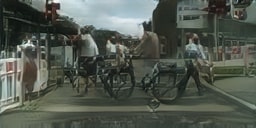} };
            \begin{scope}[
            x={($0.1*(image.south east)$)},
            y={($0.1*(image.north west)$)}]
            \draw[thick,red] (1.5,7.8) rectangle (2.5,9.99) ;
            \draw[thick,red] (8.7,8) rectangle (9.9,9.7) ;
        \end{scope}
    \end{tikzpicture}
		
	 \tabularnewline
	 
	\multirow{1}{*}{ \rotatebox{90}{\hspace{3.5em}  \small{Ours} \hspace{-4.2em} }} &	        
	\begin{tikzpicture}
            \node [
	        above right,
	        inner sep=0] (image) at (0,0) {\includegraphics[width=0.239\textwidth,]{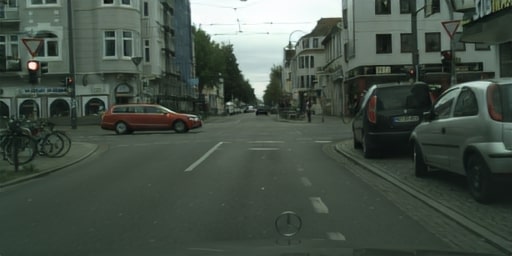}};
            \begin{scope}[
            x={($0.1*(image.south east)$)},
            y={($0.1*(image.north west)$)}]
            \draw[thick,green] (0.01,3) rectangle (4.2,8.4) ;
        \end{scope}
    \end{tikzpicture}
	        & {\footnotesize{}}
	 \begin{tikzpicture}
            \node [
	        above right,
	        inner sep=0] (image) at (0,0) {\includegraphics[width=0.239\textwidth,]{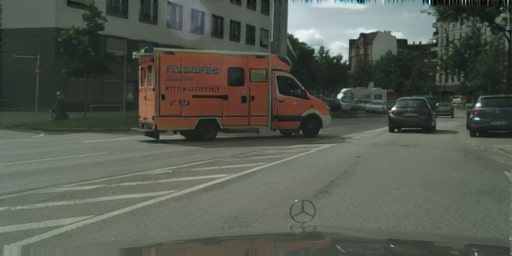} };
            \begin{scope}[
            x={($0.1*(image.south east)$)},
            y={($0.1*(image.north west)$)}]
            \draw[thick,green] (2.5,4.2) rectangle (6.7,8.5) ;
        \end{scope}
\end{tikzpicture}
	        & {\footnotesize{}}
	\begin{tikzpicture}
            \node [
	        above right,
	        inner sep=0] (image) at (0,0) {\includegraphics[width=0.239\textwidth,]{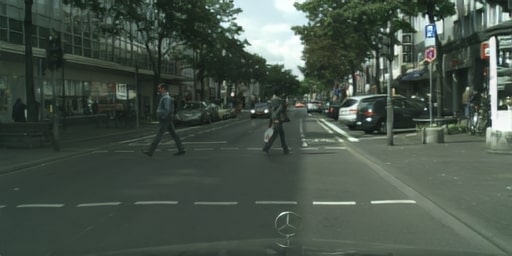} };
            \begin{scope}[
            x={($0.1*(image.south east)$)},
            y={($0.1*(image.north west)$)}]
            \draw[thick,green] (2.5,3.5) rectangle (6.7,7) ;
        \end{scope}
    \end{tikzpicture}
		    & {\footnotesize{}}
	\begin{tikzpicture}
            \node [
	        above right,
	        inner sep=0] (image) at (0,0) {\includegraphics[width=0.239\textwidth,]{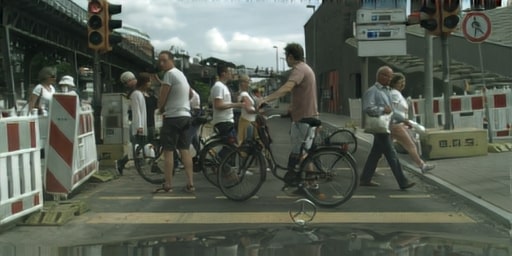}};
            \begin{scope}[
            x={($0.1*(image.south east)$)},
            y={($0.1*(image.north west)$)}]
            \draw[thick,green] (1.5,7.8) rectangle (2.5,9.99) ;
            \draw[thick,green] (8.7,8) rectangle (9.9,9.7) ;
        \end{scope}
\end{tikzpicture}
	 \tabularnewline
	\end{tabular}
\hfill{}
\par\end{centering}
\caption{
Qualitative results (best view in color and zoom in) of StyleGAN2 inversion methods on Cityscapes, i.e., pSp\newcite{pspencoder}, pSp${}^\dagger$, Feature-Style encoder\newcite{feature_style} and our masked noise encoder. Note, pSp${}^\dagger$ is an improved version of pSp\newcite{pspencoder} introduced by us, training pSp with an additional discriminator and incorporate synthesized images for better initialization.
pSp${}^\dagger$ can reconstruct the rough layout of the scene but still struggles to preserve details. The Feature-Style encoder shows a better reconstruction quality, yet it cannot faithfully reconstruct small objects (e.g. pedestrian), and some objects (e.g. the vehicle, bicycle) are rather blurry. Our masked noise encoder has highest image fidelity, preserving finer details in the inverted image. 
}
\label{fig:encoder-visual}
\end{figure*}

%% file: tex/3_methods.tex
\section{Method}\label{sec:method}
\input{figs/method_overview} %
We introduce our exemplar-based style synthesis pipeline
in \cref{method:style-mixing}, which relies on GAN inversion that can offer faithful reconstruction and style mixing of images. To enable better style-content disentanglement, we propose a masked noise encoder for GAN inversion in \cref{method:gan-inversion}. 
Its detailed training loss is described in \cref{methods: losses}.

\subsection{\newchange{Exemplar-Based Style Synthesis Pipeline}}
\label{method:style-mixing}
The lack of data diversity and the existence of spurious correlation in the training set often lead to poor domain generalization. To mitigate them, 
\newchange{the proposed style synthesis pipeline aims at 1) extracting styles from given exemplars, and 2) augmenting the training samples in the source domain with the new styles, while preserving their semantic content.} 
For data augmentation, it employs GAN inversion to randomize the style-content combinations. 
In doing so, it diversifies the source dataset and reduces spurious style-content correlations. Because the content of images is preserved and only the style is changed, the ground truth label maps can be re-used for training and validation, without requiring any further annotation effort.  

Our style synthesis pipeline is built on top of an encoder-based GAN inversion, given its fast inference. GANs, such as StyleGANs\newcite{stylegan,stylegan2,stylegan2ada}, have shown the capability of encoding rich semantic and style information in intermediate features and latent spaces. For encoder-based GAN inversion, an encoder is trained to invert an input image back into the latent space of a pre-trained GAN generator. The encoder is desired to separately encode the style and content information of the input image. With such an encoder, it can synthesize new training samples with new style-content combinations.
\newchange{In particular, we are interested in intra-source style augmentation (ISSA), where the encoder should take the content and style codes from different training samples within the source domain and feed them to the pre-trained generator. If this encoder-based GAN inversion can also handle unseen data, we will further make use the styles of exemplars outside the source domain, such as web-crawled images, enabling extra-source style augmentation (ESSA).} %
In both cases, since only the styles of the training samples in the source domain are modified, the newly synthesized training samples already have their ground truth label maps in place.

StyleGAN2 can synthesize natural looking images resembling scene-centric datasets such as Cityscapes\newcite{cordts2016cityscapes} and BDD100K\newcite{yu2020bdd100k}. However, existing GAN inversion encoders cannot provide the desired fidelity and style mixing capability to enable {\ourstyle} and ESSA for an improved domain generalization of semantic segmentation. Loss of fine details or 
inauthentic reconstruction of small-scale objects would even harm the model's generalization ability. Therefore, we propose a novel encoder design to invert StyleGAN2, termed \emph{masked noise encoder} (see \cref{fig:encoder-overview}).

\subsection{Masked Noise Encoder} \label{method:gan-inversion}

We build our encoder upon the pSp encoder\newcite{pspencoder}. It employs a feature pyramid\newcite{lin2017featurepyramid} to extract multi-scale features from a given image, see \cref{fig:encoder-overview}-(A). We improve over pSp by identifying in which latent space to embed the input image for the high-quality reconstruction of the images with complex street scenes. Further, we propose a novel training scheme to enable the style-content disentanglement of the encoder, thus improving its style mixing capability.

\paragraph{Extended Latent Space}
The StyleGAN2 generator takes the latent code $w\in\mathcal{W}$ generated by an MLP network and randomly sampled additive Gaussian noise maps $\{\epsilon\}$ as inputs for image synthesis. As pointed out in\newcite{abdal2019image2stylegan}, it is suboptimal to embed a real image into the original latent space $\mathcal{W}$ of StyleGAN2, due to the gap between the real and synthetic data distributions. A common practice is to map the input image into the extended latent space \wplus. 
The multi-scale features of the pSp feature pyramid are respectively mapped to the latent codes $\{w^k\}$ at the corresponding scales of the StyleGAN2 generator, i.e., $\mathrm{map2latent}$ in \cref{fig:encoder-overview}-(A).

\paragraph{Additive Noise Map}\label{para:noisemap}
The latent codes $\{w^k\}$ from the extended latent space {\wplus} alone are not expressive enough to reconstruct images with diverse semantic layouts such as Cityscapes\newcite{cordts2016cityscapes} as shown in \cref{fig:encoder-visual}-(pSp${}^\dagger$). 
The latent codes of StyleGAN2 are one-dimensional vectors that modulate the feature vectors at different spatial positions identically. Therefore, they cannot precisely encode the semantic layout information, which is spatially varying. To address this issue, our encoder additionally predicts the additive noise map $\varepsilon$ of the StyleGAN2 at an intermediate scale, i.e., $\mathrm{map2noise}$ in \cref{fig:encoder-overview}-(B). The noise map $\varepsilon$  has spatial
dimensions, making it inherently capable of encoding more information. It is particularly advantageous when dealing with content information that varies spatially, as the noise map can more readily accommodate such information.
As evidenced by the visualization presented in  \cref{fig:noise-vis}, the noise map is adept at capturing the semantic content of the scene.

\input{figs/masking_ablation_2x2} %
\input{figs/noise}                %
\input{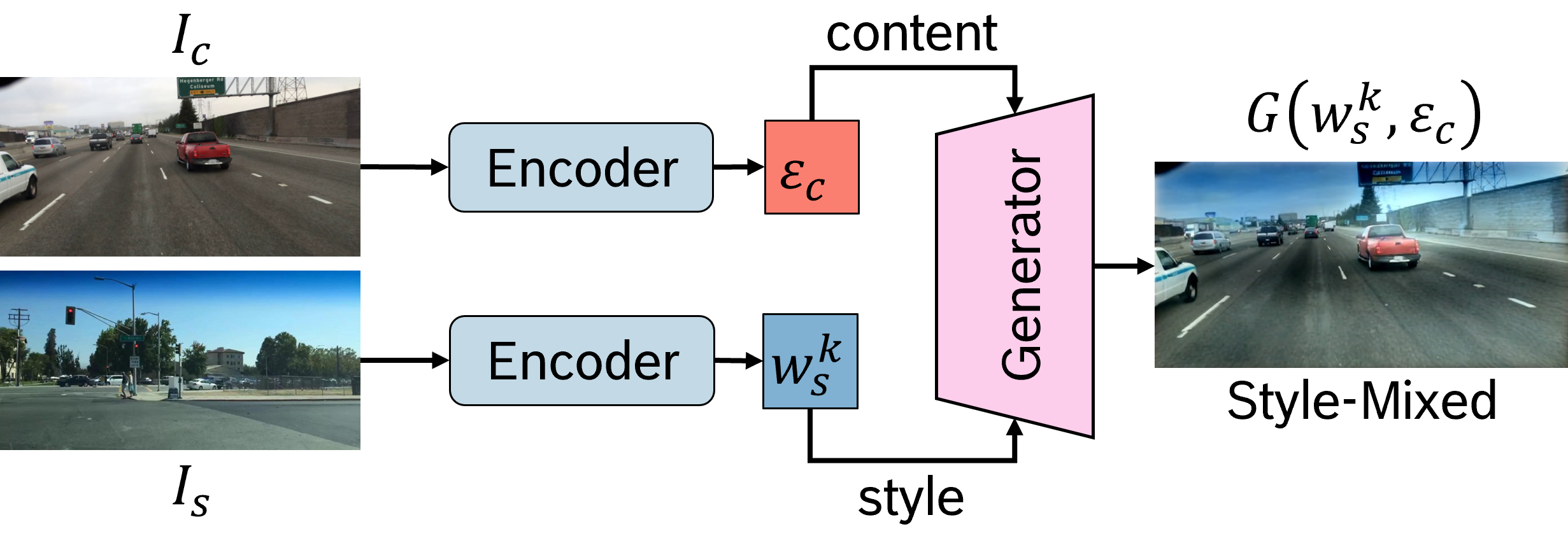}
\input{figs/intramix_bdd100k}     %

\paragraph{Random Noise Masking}\label{masking-page}
While offering high-quality reconstruction, the additive noise map can be too expressive so that it encodes nearly all perceivable details of the input image. This results in a poor style-content disentanglement and can damage the style mixing capability of the encoder (see \cref{fig:masking-ablation}). 
To avoid this undesired effect, we propose to regularize the noise prediction of the encoder by random masking of the noise map. Note that the random masking as a regularization technique has also been successfully used in reconstruction-based self-supervised learning\newcite{xie2022simmim,he2022mae}. 
In particular, we spatially divide the noise map into non-overlapping $P\times P$ patches, see $\fbox{M}$ in \cref{fig:encoder-overview}-(B). Based on a pre-defined ratio $\rho$, a subset of patches is randomly selected and replaced by patches of unit Gaussian random variables $\epsilon \sim N(0,1)$ of the same size. $N(0,1)$ is the prior distribution of the noise map at training the StyleGAN2 generator. 
We call this encoder \emph{masked noise encoder} as it is trained with random masking to predict the noise map.

The proposed random masking reduces the encoding capacity of the noise map, hence encouraging the encoder to jointly exploit the latent codes $\{w^k\}$ for reconstruction. 
\cref{fig:intra-mix-example} visualizes the style mixing effect. 
The encoder takes the noise map $\varepsilon_c$ and latent codes $\{w_s^k\}$ from the $\mathrm{content}$ image and $\mathrm{style}$ image, respectively. 
Then, they are fed into StyleGAN2 to synthesize a new image, i.e., $G(w_s^k, \varepsilon_c)$, as illustrated in \cref{fig:mix_up}. 
If the encoder is not trained with random masking, the new image does not have any perceptible difference with the $\mathrm{content}$ image. This means the latent codes $\{w^k\}$ encode negligible information of the image. In contrast, when being trained with masking, the encoder creates a novel image that takes the content and style from two different images. This observation confirms the enabling role of masking for content and style disentanglement, and thus the improved style mixing capability. The noise map no longer encodes all perceptible information of the image, including style and content. In effect, the latent codes $\{w^k\}$ play a more active role in controlling the style.  
In \cref{fig:noise-vis}, we further visualize the noise map of the masked noise encoder and observe that it captures well the semantic content of the scene. 

\newchange{
Additionally, we discover that our masked noise encoder is equipped with strong plug-n-play ability, i.e., readily usable on novel domains without retraining or fine-tuning. As shown in \cref{fig:landscape-example}, the masked noise encoder together with the generator which is trained on Cityscapes not only reconstruct unseen domain data (e.g., north polar bear), but also remain the style mixing capability (e.g., turning bright day into a sunset scene). This generalization capability allows us to further exploit extra-source data for style synthesis, i.e., ESSA. Except that the styles are extracted from external exemplars, the style synthesis process of ESSA is identical to ISSA.} 

\input{tex/3_method_training_loss}

%% file: figs/method_overview.tex
\begin{figure*}[t]
\centering
\includegraphics[width=0.97\linewidth]{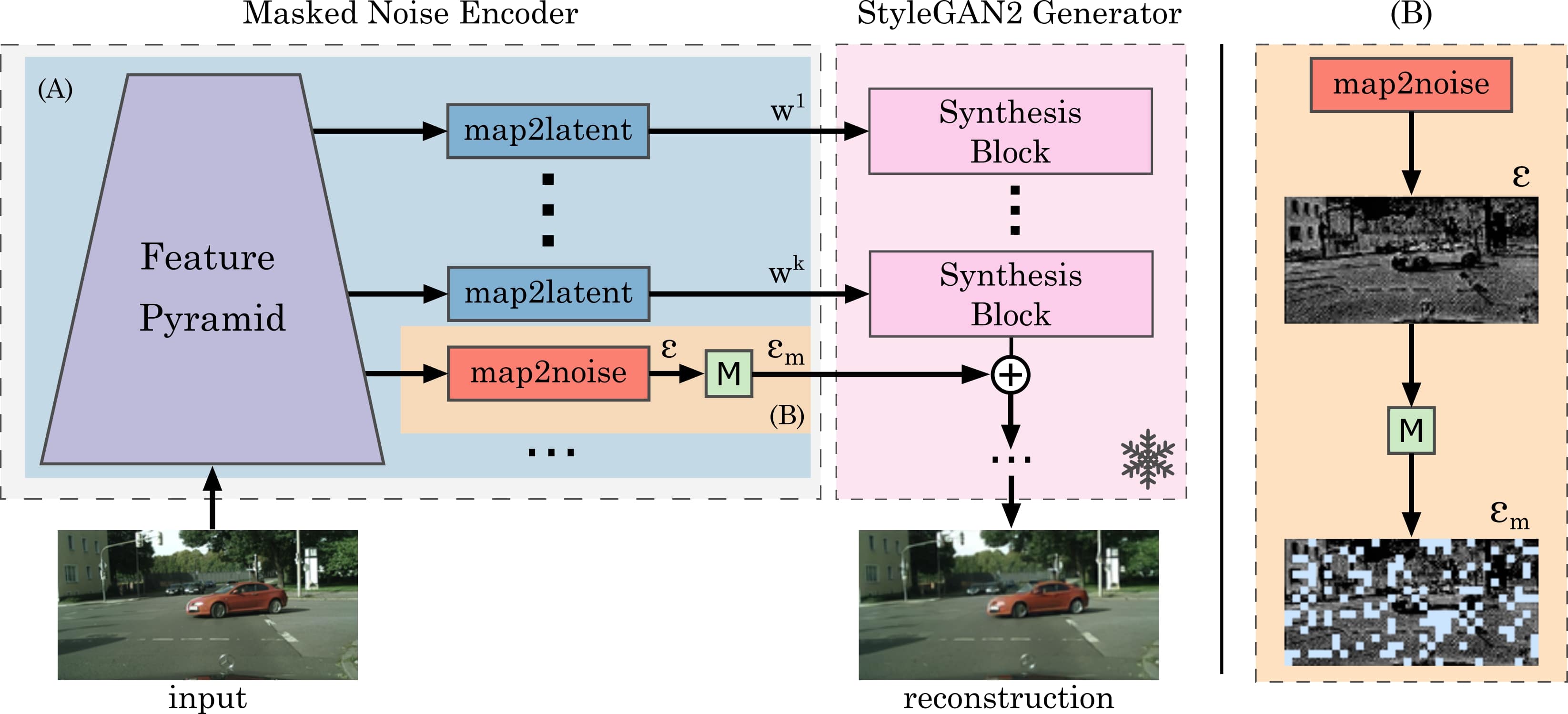}
\caption{\textbf{Method overview.} Our encoder is built on top of the pSp encoder\newcite{pspencoder}, shown in the blue area (A). It maps the input image to the extended latent space $\mathcal{W}^+$ of the pre-trained StyleGAN2 generator. To promote the reconstruction quality on complex scene-centric dataset, e.g., Cityscapes, our encoder additionally predicts the noise map at an intermediate scale, illustrated in the orange area (B). 
$\fbox{M}$ stands for random noise masking, regularization for the encoder training. Without it, the noise map overtakes the latent codes in encoding the image style, so that the latter cannot make any perceivable changes on the reconstructed image, thus making style mixing impossible. 
}
\label{fig:encoder-overview}
\end{figure*}

%% file: figs/masking_ablation_2x2.tex
\begin{figure}[t]
    \begin{centering}
    \setlength{\tabcolsep}{0.0em}
    \renewcommand{\arraystretch}{0}
    \par\end{centering}
    \begin{centering}
    \hfill{}
	\begin{tabular}{@{}c@{}c}
        \centering
		\small Content &  \small Style   \tabularnewline
		\includegraphics[width=0.47\linewidth]{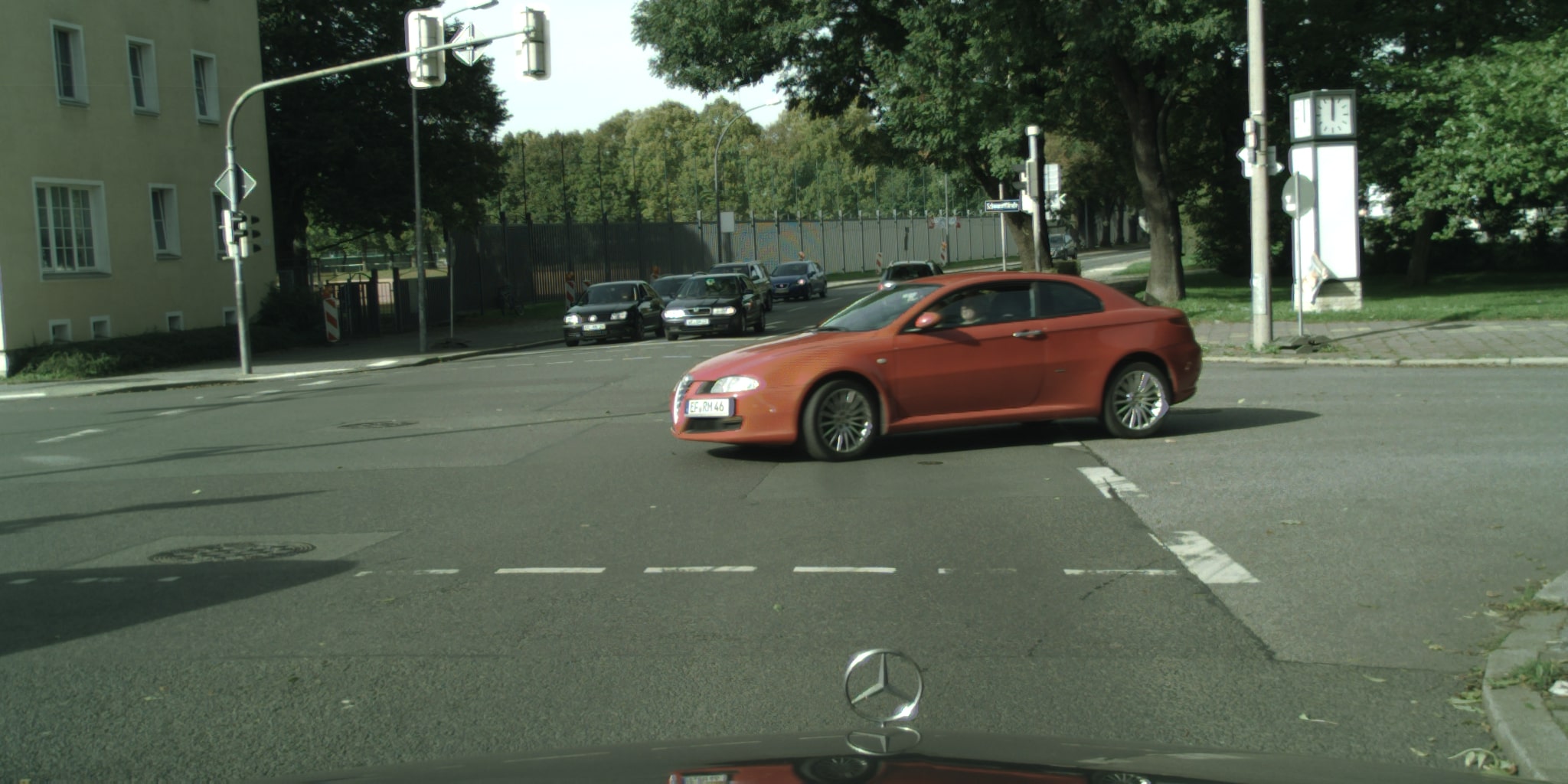} & {\footnotesize{}} 
		\includegraphics[width=0.47\linewidth]{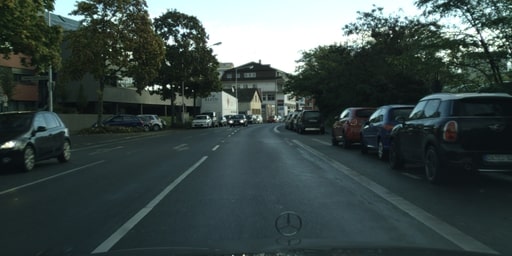} \tabularnewline
	     \small W/o masking & \small W/- masking (Ours)  \tabularnewline
		\includegraphics[width=0.47\linewidth]{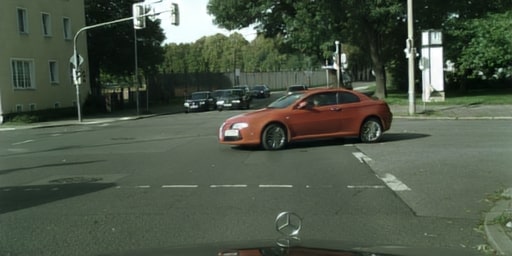} & {\footnotesize{}} 
		\includegraphics[width=0.47\linewidth]{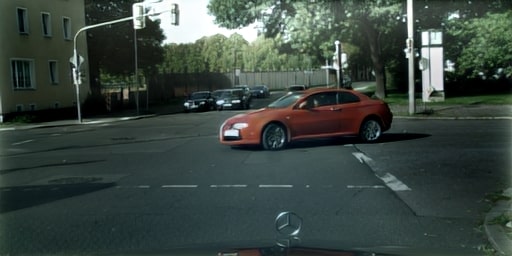} 
		\tabularnewline
		\end{tabular}
\hfill{}
\par\end{centering}
\caption{Style mixing effect enabled by random noise masking (best view in color). Despite the good reconstruction quality, the encoder trained without masking cannot change the style of the given $\mathrm{Content}$ image. In contrast, the encoder trained with masking can modify it using the style from the given $\mathrm{Style}$ image.}
\label{fig:masking-ablation}
\end{figure}

%% file: figs/noise.tex
\begin{figure}[t]
    \begin{centering}
    \setlength{\tabcolsep}{0.0em}
    \renewcommand{\arraystretch}{0}
    \par\end{centering}
    \begin{centering}
    \hfill{}
	\begin{tabular}{@{}c@{}c}
        \centering
		 & \tabularnewline
		\includegraphics[width=0.47\linewidth]{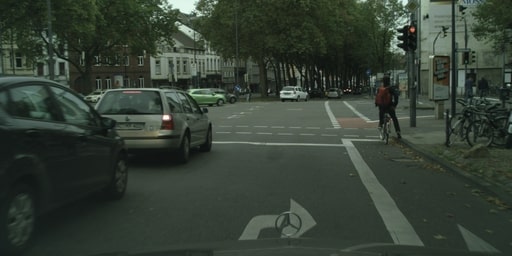} &
		{\footnotesize{}}
		\includegraphics[width=0.47\linewidth]{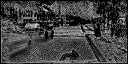}
		\tabularnewline
		
		\includegraphics[width=0.47\linewidth]{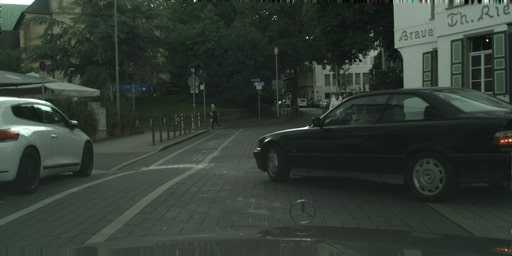} & {\footnotesize{}}
		\includegraphics[width=0.47\linewidth]{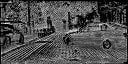}
        \tabularnewline
        
		\includegraphics[width=0.47\linewidth]{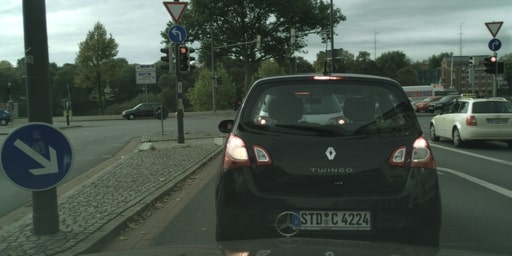} & {\footnotesize{}}
		\includegraphics[width=0.47\linewidth]{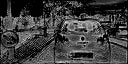} 
		\tabularnewline
		\end{tabular}
\hfill{}
\par\end{centering}
\caption{Noise map visualization of our masked noise encoder. The noise map encodes the semantic content of the image.}
\label{fig:noise-vis}
\end{figure}

%% file: figs/mix_up.tex
\begin{figure}[t]
\centering
\includegraphics[width=0.99\linewidth]{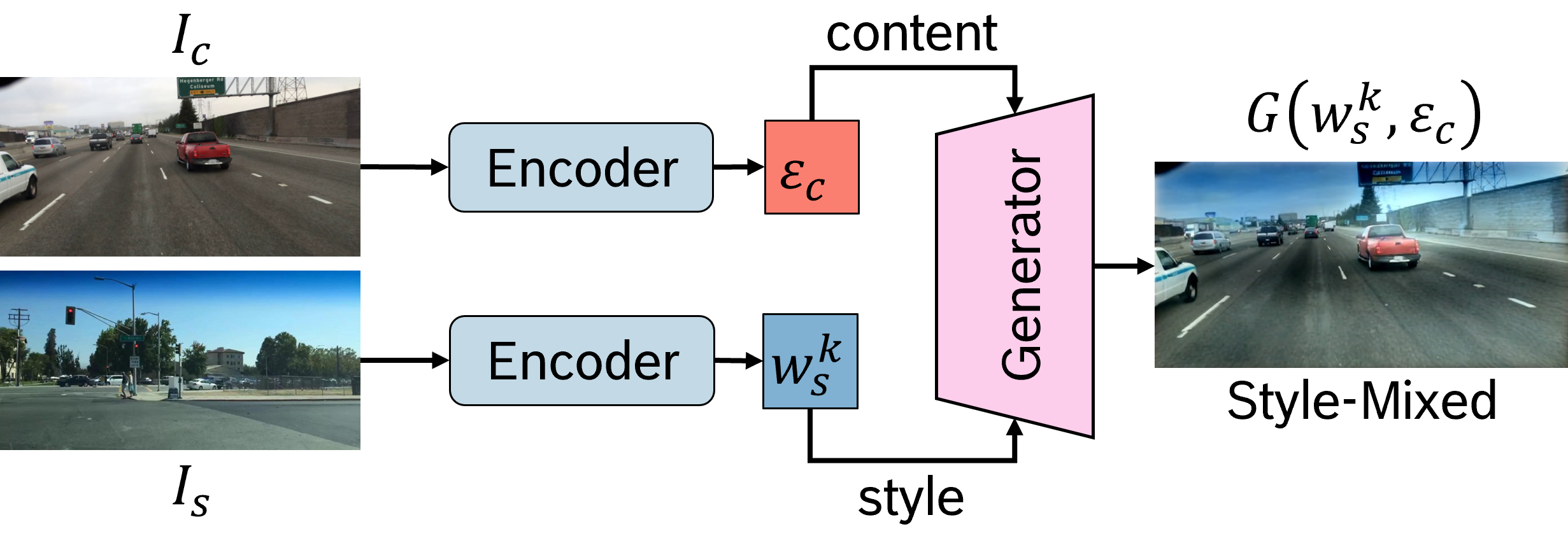}
\caption{
    \textbf{Style mixing process.} 
    The generator $G$ takes the 
    latent codes $\{w_s^k\}$ of $I_s$ and the noise map $\varepsilon_c$ of $I_c$, and produce the stylized image, i.e., $G(w_s^k, \varepsilon_c)$.
}
\label{fig:mix_up}
\end{figure}

%% file: figs/intramix_bdd100k.tex
\begin{figure*}[t]
    \begin{centering}
    \setlength{\tabcolsep}{0.0em}
    \renewcommand{\arraystretch}{0}
    \par\end{centering}
    \begin{centering}
    \hfill{}
	\begin{tabular}{@{}l@{}c@{}c@{}c}
        \centering
		 &   &  &  \tabularnewline
        \hspace{0.08\textwidth} Content $I_c$ \hspace{0.05\textwidth} \rotatebox{90}{\hspace{1.3em} Style $I_s$}

		& {\footnotesize{}}
		\includegraphics[width=0.235\textwidth]{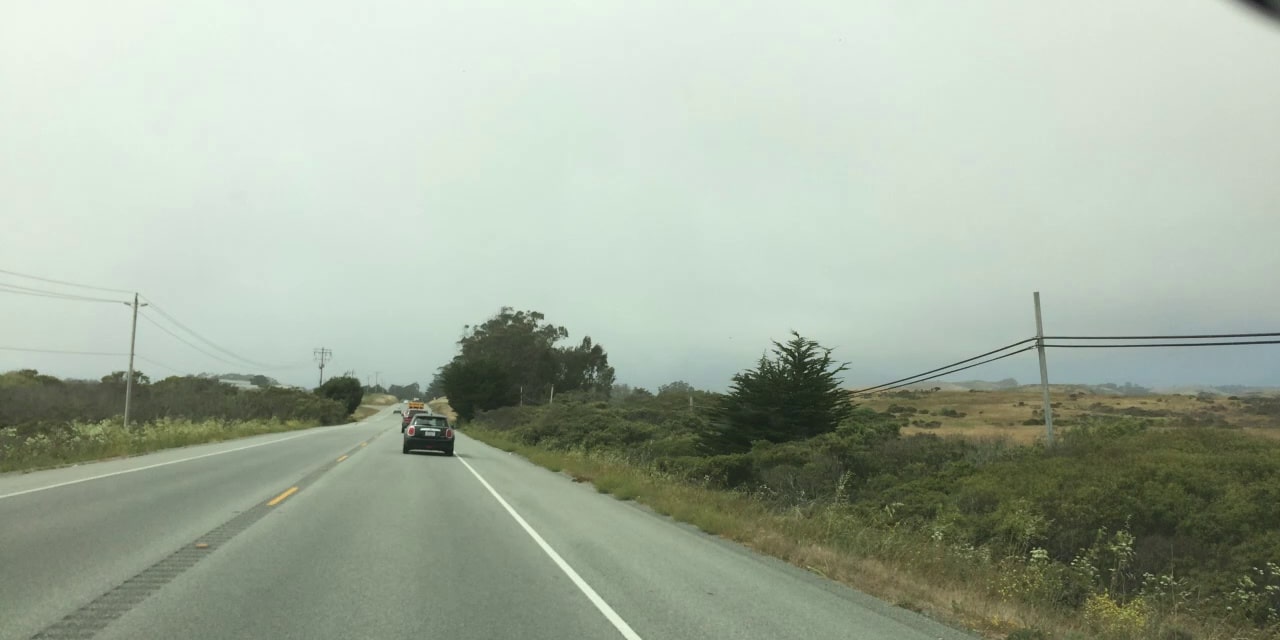} & {\footnotesize{}}
		\includegraphics[width=0.235\textwidth]{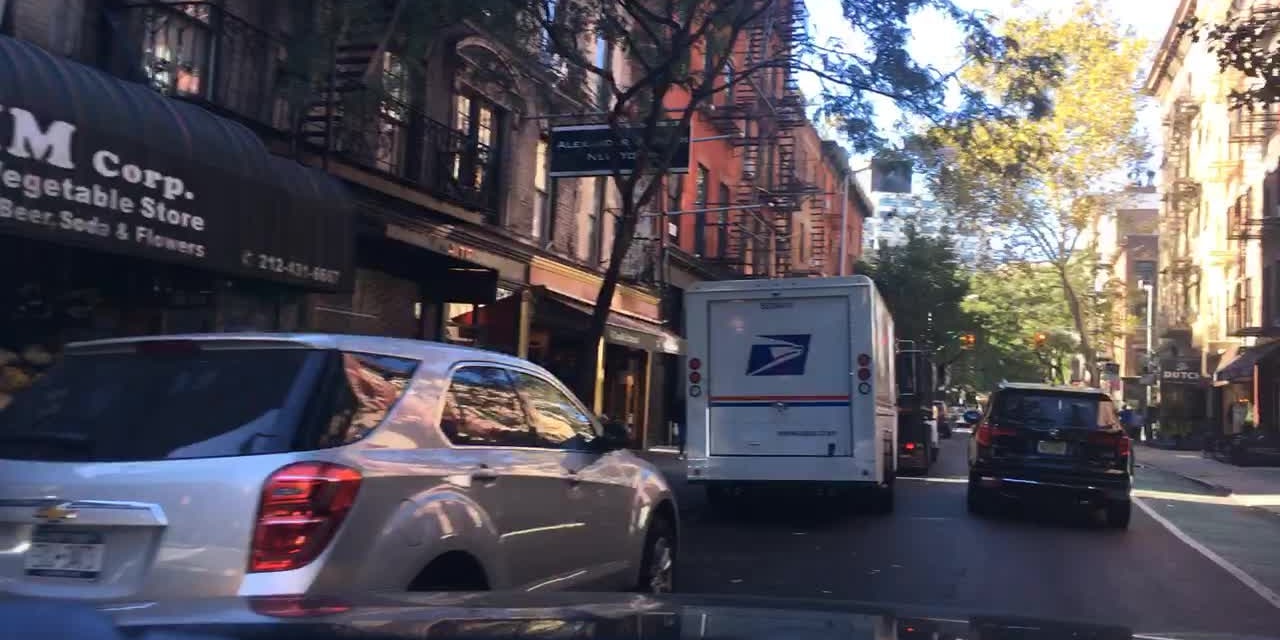} & {\footnotesize{}}
		\includegraphics[width=0.235\textwidth]{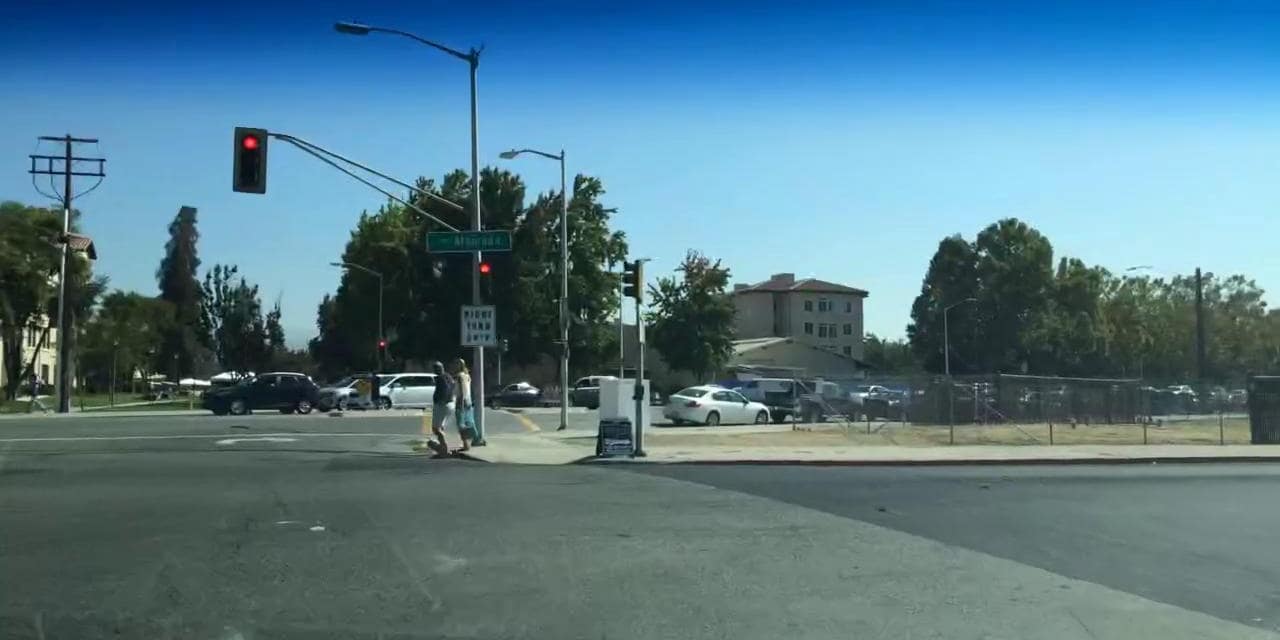}
		\tabularnewline
		
		\includegraphics[width=0.235\textwidth]{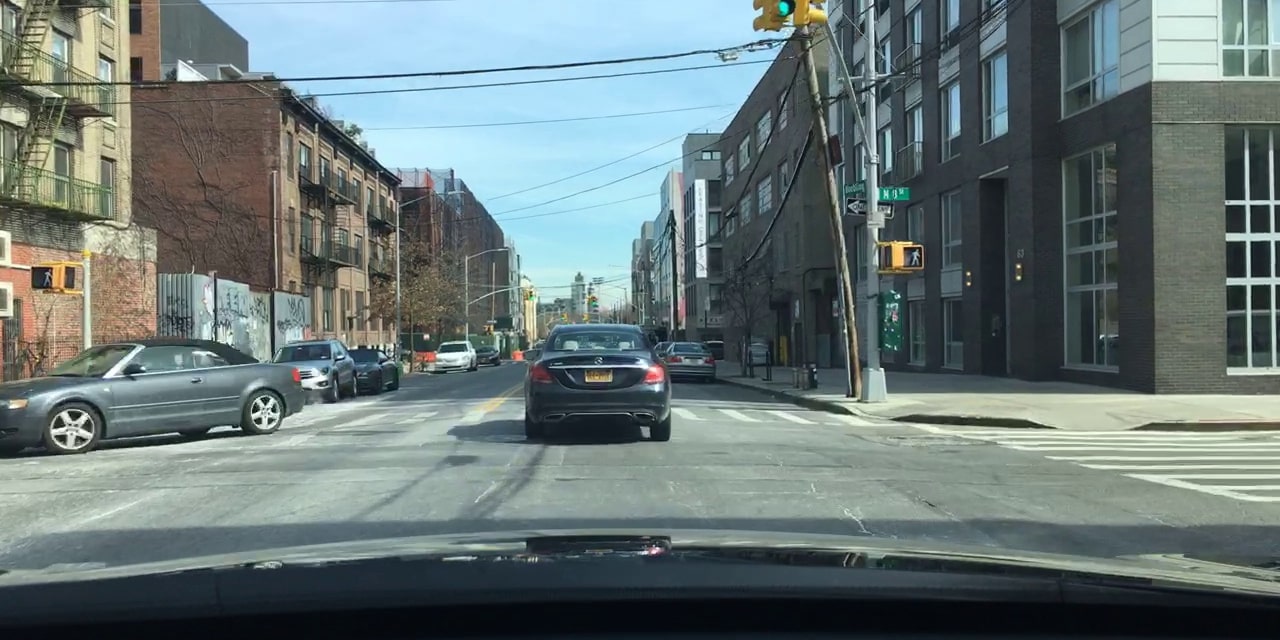} & {\footnotesize{}}
		\includegraphics[width=0.235\textwidth]{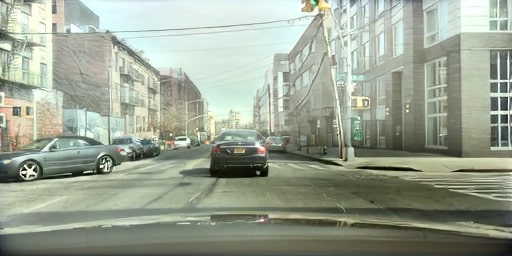} & {\footnotesize{}}
		\includegraphics[width=0.235\textwidth]{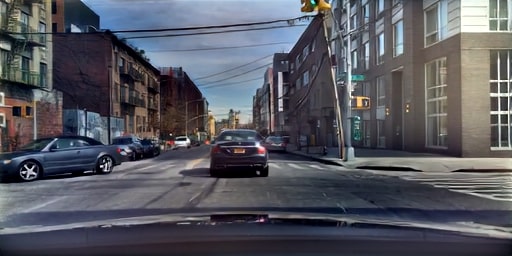} & {\footnotesize{}}
		\includegraphics[width=0.235\textwidth]{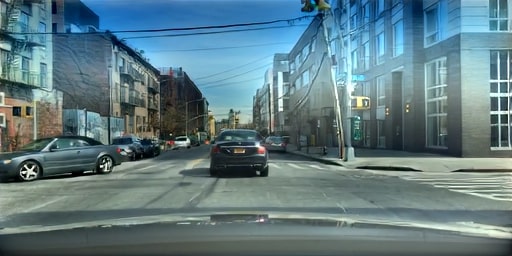}
		\tabularnewline
		
		\includegraphics[width=0.235\textwidth]{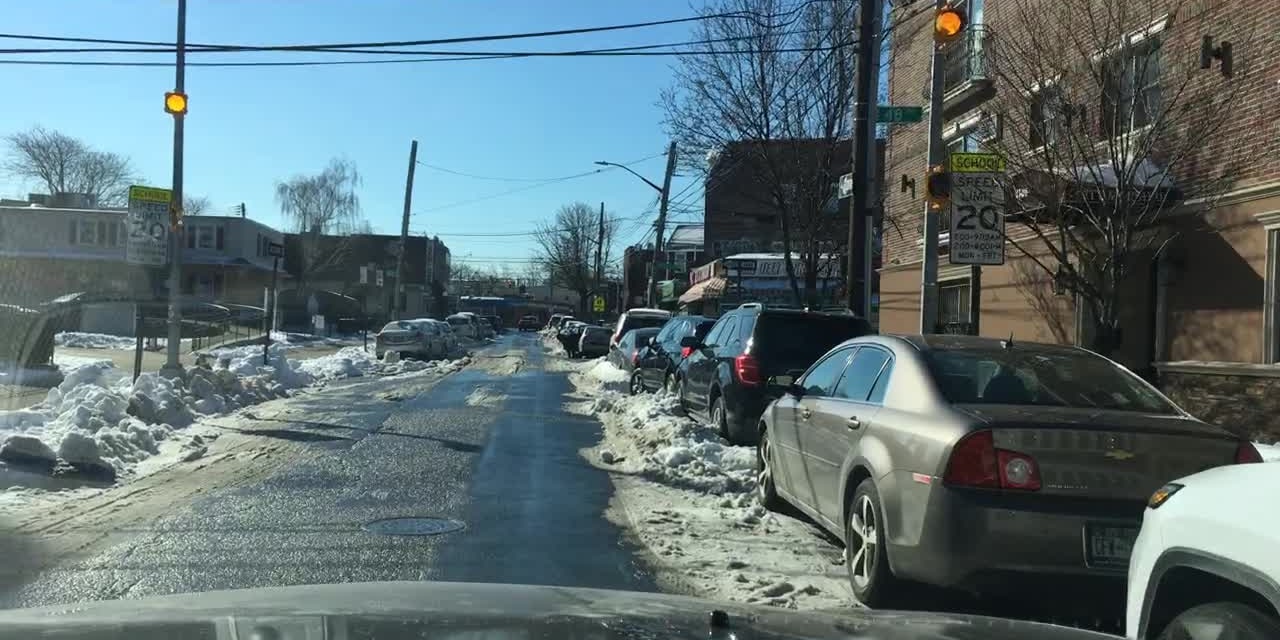} & {\footnotesize{}}
		\includegraphics[width=0.235\textwidth]{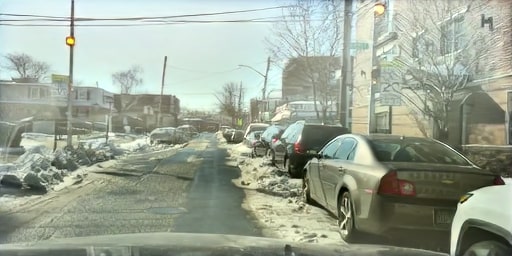} & {\footnotesize{}}
		\includegraphics[width=0.235\textwidth]{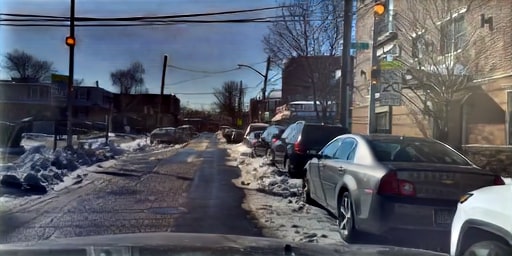} & {\footnotesize{}}
		\includegraphics[width=0.235\textwidth]{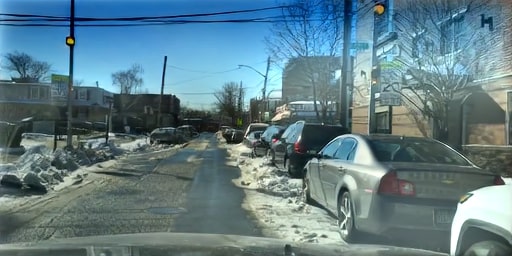}
		\tabularnewline
		
		\includegraphics[width=0.235\textwidth]{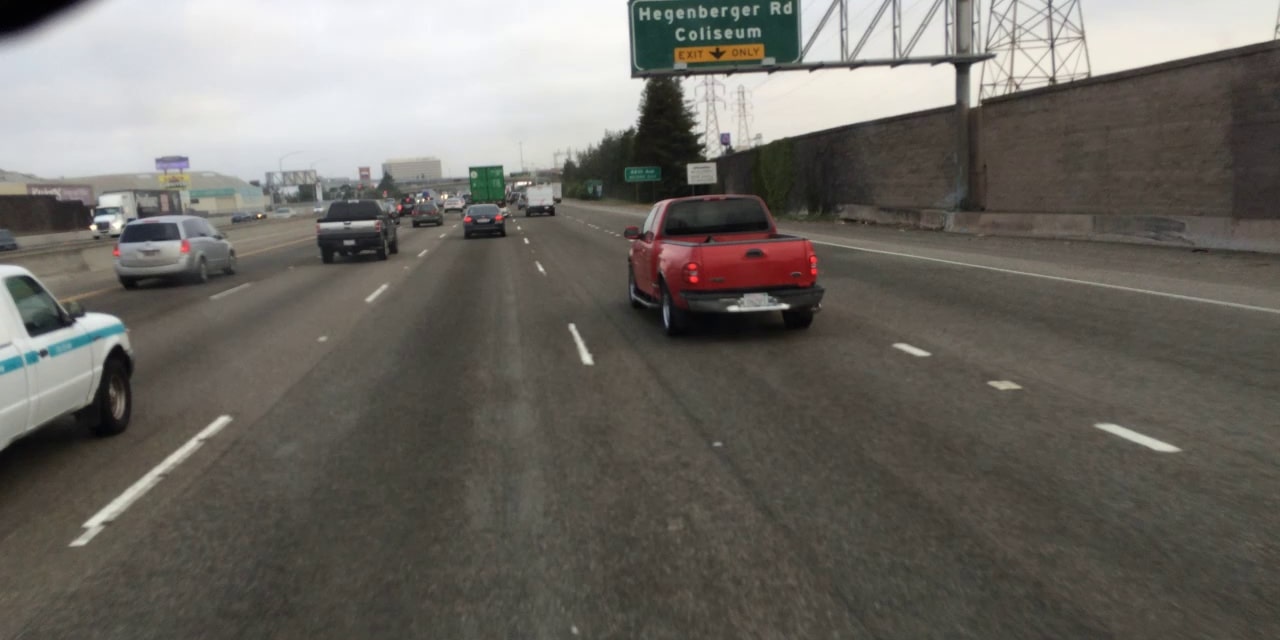} & {\footnotesize{}}
		\includegraphics[width=0.235\textwidth]{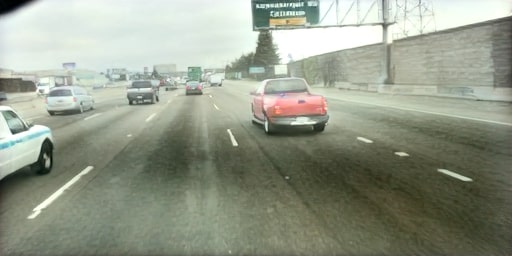} & {\footnotesize{}}
		\includegraphics[width=0.235\textwidth]{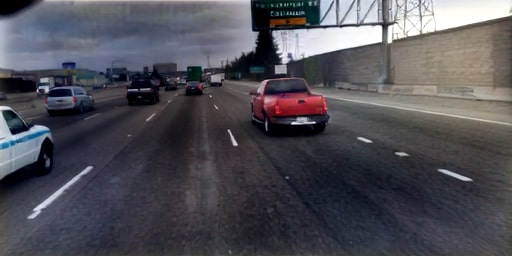} & {\footnotesize{}}
		\includegraphics[width=0.235\textwidth]{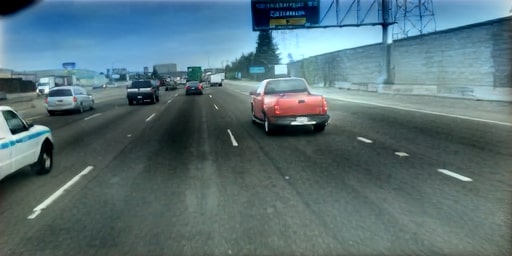}
		\tabularnewline
		
		\includegraphics[width=0.235\textwidth]{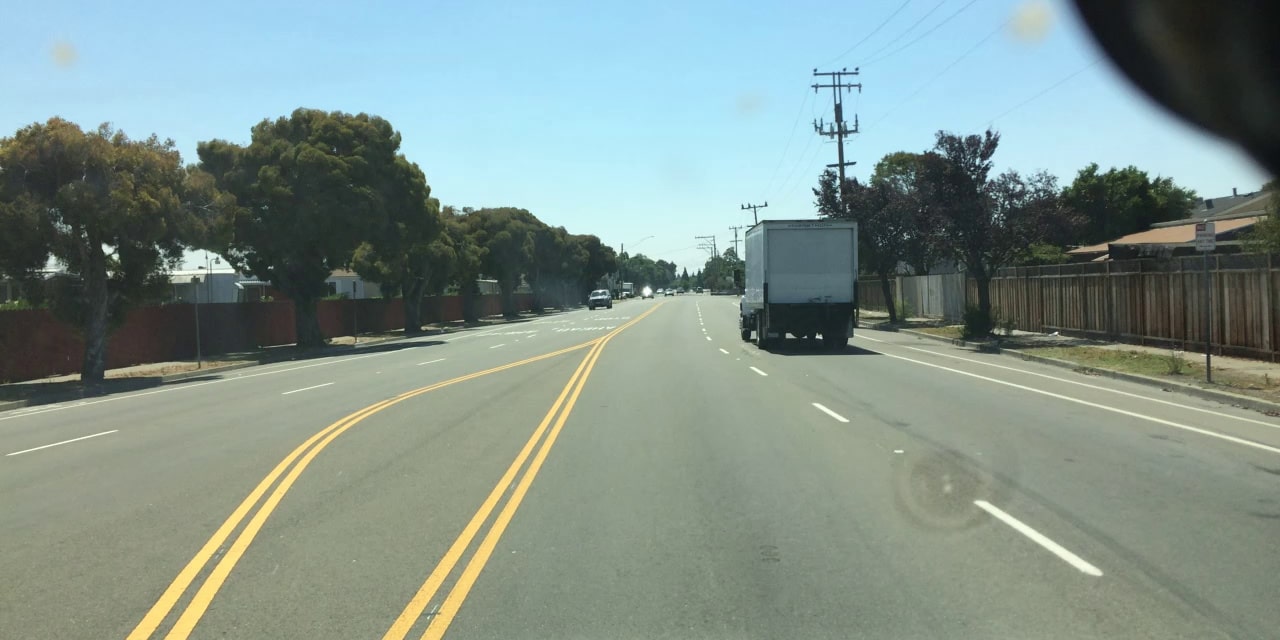} & {\footnotesize{}}
		\includegraphics[width=0.235\textwidth]{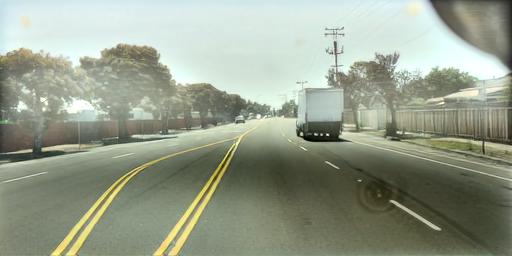} & {\footnotesize{}}
		\includegraphics[width=0.235\textwidth]{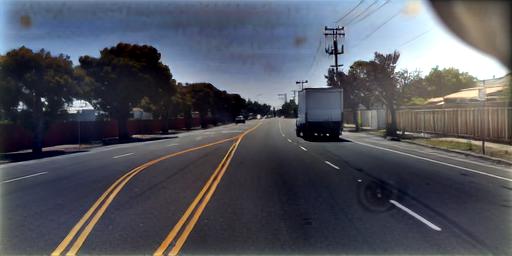} & {\footnotesize{}}
		\includegraphics[width=0.235\textwidth]{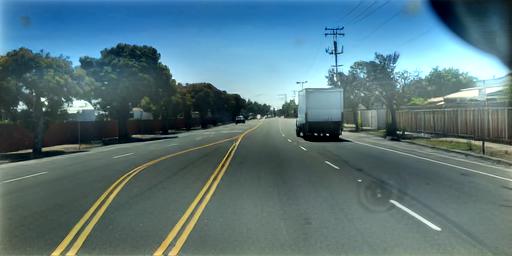}
		\tabularnewline
		
		\includegraphics[width=0.235\textwidth]{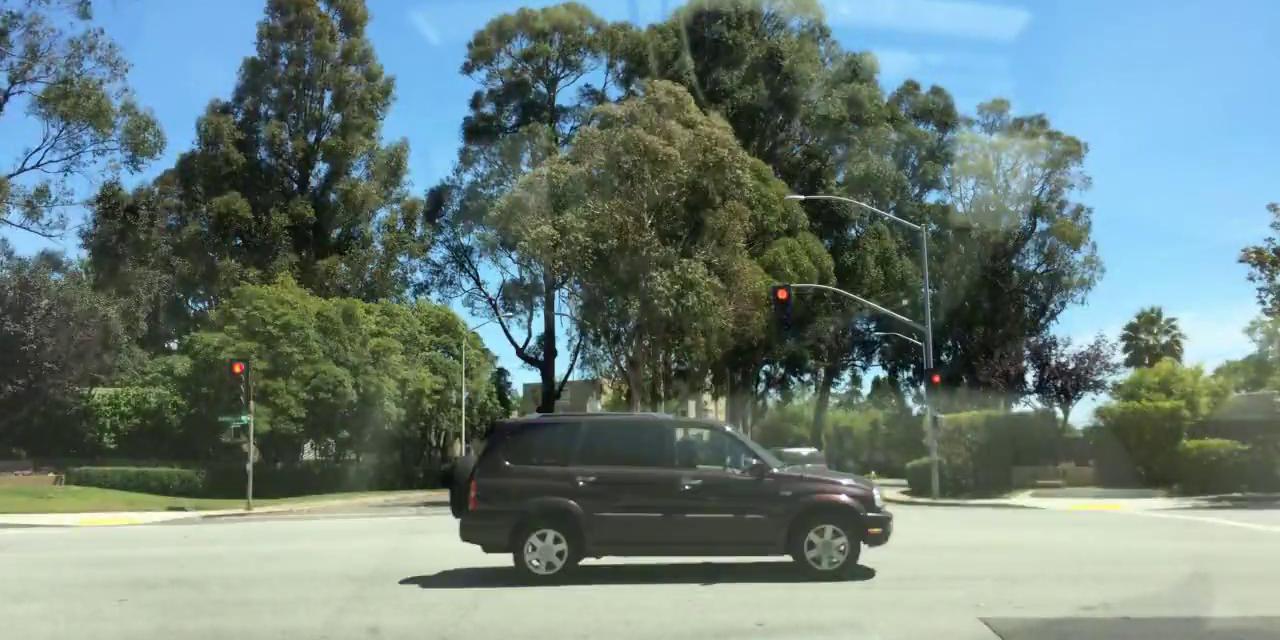} & {\footnotesize{}}
		\includegraphics[width=0.235\textwidth]{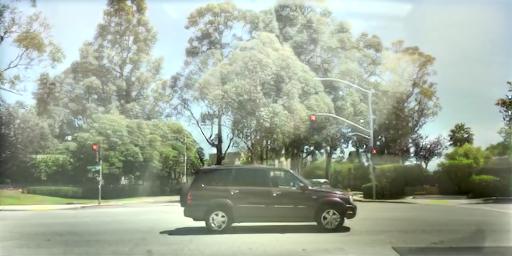} & {\footnotesize{}}
		\includegraphics[width=0.235\textwidth]{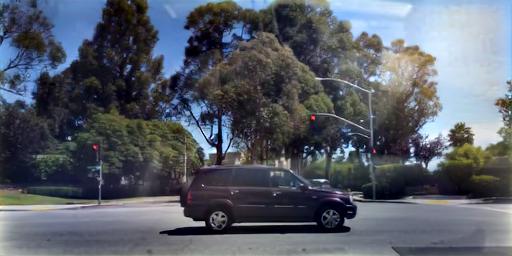} & {\footnotesize{}}
		\includegraphics[width=0.235\textwidth]{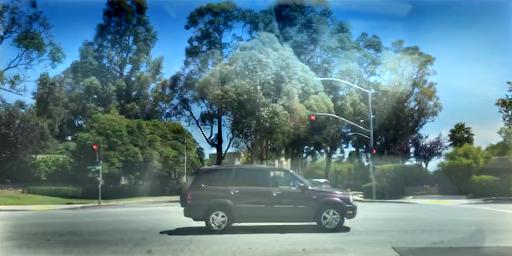}
		\tabularnewline
		
		\end{tabular}
\hfill{}
\par\end{centering}
\caption{Visual examples of style mixing on BDD100K (best view in color) enabled by our masked noise encoder. 
By combining the latent codes $\{w_s^k\}$ of $I_s$ and the noise map $\varepsilon_c$ of $I_c$, the synthesized images $G(w_s^k, \varepsilon_c)$ preserve the content of $I_c$ with a new style resembling $I_s$. 
}
\label{fig:intra-mix-example}
\end{figure*}

%% file: tex/3_method_training_loss.tex
\subsection{Encoder Training Loss}\label{methods: losses}
Mathematically, the proposed StyleGAN2 inversion with the masked noised encoder $E^M$ can be formulated as
\begin{align}
    \{w^1,\dots, w^K,\varepsilon\} &= E^M(x); \label{eq:encoder-1}\\
    x^* &=G\circ E^M(x)= G(w^1,\dots,w^K, \varepsilon).\nonumber 
\end{align}
The masked noise encoder $E^M$ maps the given image $x$ onto the latent codes $\{w^k\}$ and the noise map $\varepsilon$. The StyleGAN2 generator $G$ takes both $\{w^k\}$ and $\varepsilon$ as the input and generates $x^*$. Ideally, $x^*$ should be identical to $x$, i.e., a perfect reconstruction. 

When training the masked noise encoder $E^M$ to reconstruct $x$, the original noise map $\varepsilon$ is masked before being fed into the pre-trained $G$
\begin{align} 
\varepsilon_M &= (1 - M_{noise})  \odot \varepsilon + M_{noise} \odot \epsilon, \\
\tilde{x} &= G(w^1,\dots,w^K, \varepsilon_M), \label{eq:encoder}
\end{align}
where $M_{noise}$ is the random binary mask, $\odot$ indicates the Hadamard product, and $\tilde{x}$ denotes the reconstructed image with the masked noise $\varepsilon_M$. The training loss for the encoder is given as
\begin{align} 
\mathcal{L} =  \mathcal{L}_{mse} + 
    \lambda_1 \mathcal{L}_{lpips} + 
    \lambda_2 \mathcal{L}_{adv} + \lambda_3 \mathcal{L}_{reg} ,\label{eq:loss}
\end{align}
where $\{\lambda_i\}$ are weighting factors. %
The first three terms are the pixel-wise MSE loss, learned perceptual image patch similarity (LPIPS)\newcite{zhang2018lpips} loss and adversarial loss\newcite{goodfellow2014gan}, 
\begin{align} 
\mathcal{L}_{mse} &= \norm{(1 - M_{img})  \odot (x - \tilde{x})}_2, \\
\mathcal{L}_{lpips} &= \norm{(1 - M_{feat}) \odot
(\mathrm{VGG}(x) - \mathrm{VGG}(\tilde{x}))}_2, \\
\mathcal{L}_{adv} &= -\log D(G(E^M(x))).
\label{eq:encoder-loss-rec}
\end{align}
which are the common reconstruction losses for encoder training\newcite{pspencoder,zhu2020indomain}. Note that masking removes the information of the given image $x$ at certain spatial positions, the reconstruction requirement on these positions should then be relaxed. $M_{img}$ and $M_{feat}$ are obtained by up- and down-sampling the noise mask $M_{noise}$ to the image size and the feature size of the VGG-based feature extractor. The adversarial loss is obtained by formulating the encoder training as an adversarial game with a discriminator $D$ that is trained to distinguish between reconstructed and real images.

The last regularization term is defined as
\begin{align}
\mathcal{L}_{reg}=\norm{\varepsilon}_1 +  \norm{E^M_{w}(G(w_{gt}, \epsilon)) - w_{gt}}_2.
\end{align}
The L1 norm helps to induce sparse noise prediction. It is complementary to random masking, reducing the capacity of the noise map. The second term is obtained by using the ground truth latent codes $w_{gt}$ of synthesized images $G(w_{gt}, \epsilon)$ to train the latent code prediction $E^M_{w}(\cdot)$\newcite{feature_style}. It guides the encoder to stay close to the original latent space of the generator, speeding up the convergence.

%% file: tex/4_experiments.tex

\section{Experiments}\label{sec:exp}
We start from the experiment setup in \cref{sec:exp_setup}. Then, \cref{sec:exp_gan_inversion} and \cref{sec:exp_dg} respectively report our experiments on the masked noise encoder for StyleGAN2 inversion and ISSA for improved domain generalization of semantic segmentation.

\subsection{Experiment Setup}\label{sec:exp_setup}
\paragraph{Datasets}
We conduct extensive experiments on four driving scene datasets, which are Cityscapes~(CS)\newcite{cordts2016cityscapes}, BDD100K~(BDD)\newcite{yu2020bdd100k}, ACDC\newcite{sakaridis2021acdc} and Dark Z\"urich~(DarkZ)\newcite{sakaridis2019darkzurich}. Cityscapes is collected from different cities primarily in Germany, under good/medium weather conditions during daytime. BDD100K is a driving-scene dataset collected in the US, representing a geographic location shift from Cityscapes. Besides, it also includes more diverse scenes (e.g., city streets, residential areas, and highways) and different weather conditions captured at different times of the day. Both ACDC and Dark Z\"urich are collected in Switzerland. ACDC contains four adverse weather conditions (rain, fog, snow, night) and Dark Z\"urich contains night scenes. The default setting is to use Cityscapes as the source training data, whereas the validation sets of the other datasets represent unseen target domains with different types of natural shifts, i.e., used only for testing. 
\newchange{Additionally, we also study the challenging day-to-night generalization scenario, where BDD100K-Daytime is used as the source set, ACDC-Night and Dark Z\"urich are treated as unseen domains.}
In both cases, we consider a \emph{single source domain} for training.

\paragraph{Training details}
We experiment with two image resolutions: $128\times256$ and $256\times512$. 
The StyleGAN2\newcite{stylegan2ada} model is first trained to \emph{unconditionally} synthesize images and then fixed during the encoder training. To invert the pre-trained StyleGAN2 generator, the masked noise encoder predicts both latent codes in the extended {\wplus} space and the additive noise map. In accordance with the StyleGAN2 generator, {\wplus} space consists of $14$ and $16$ latent code vectors for the input resolution $128\times256$ and $256\times512$, respectively. The additive noise map is always at the intermediate feature space with one fourth of the input resolution. We use the same encoder architecture, optimizer, and learning rate scheduling as pSp\newcite{pspencoder}. Our encoder is trained with the loss function defined in \cref{eq:loss} with $\lambda_1=10$ and $\lambda_2= \lambda_3=0.1$. For our random noise masking, we use a patch size $P$ of $4$ with a masking ratio $\rho= 25\%$. \newchange{A detailed ablation study on the masking and noise map of the encoder can be found in \cref{sec:exp_gan_inversion}}.

We use the trained masked noise encoder to perform {\ourstyle} as described in \cref{method:style-mixing}. We experiment with several architectures for semantic segmentation, i.e., HRNet\hrnet, SegFormer\segformer, and DeepLab v2/v3+\newcite{chen2017deeplabv2,chen2018deeplabv3plus}. 
The baseline segmentation models are trained with their default configurations and using the standard augmentation, i.e., random scaling and horizontal flipping.

\input{figs/noise_map_res_ablation} 

\subsection{Masked Noise Encoder}\label{sec:exp_gan_inversion}
\paragraph{Reconstruction quality} 
\Cref{tab:gan-inversion-comparison} shows that our masked noise encoder considerably outperforms two strong \mbox{StyleGAN2} inversion baselines pSp\newcite{pspencoder} and Feature-Style encoder\newcite{feature_style} in all three evaluation metrics. The achieved low values of MSE, LPIPS\newcite{zhang2018lpips} and FID\newcite{heusel2017fid} indicate its high-quality reconstruction. 
Both the masked noise encoder and the Feature-Style encoder adopt the adversarial loss $\mathcal{L}_{adv}$  
and regularization using synthesized images with ground truth latent codes $w_{gt}$.
Therefore, we also add them to train pSp and note this version as $\text{pSp}^\dagger$. While $\text{pSp}^\dagger$ improves over pSp in MSE and FID, it still underperforms compared to the others. This confirms that inverting into the extended latent space {\wplus} only allows limited reconstruction quality on Cityscapes.
The Feature-Style encoder\newcite{feature_style} replaces the prediction of the low level latent codes with feature prediction, which results in better reconstruction without severely harming style editability. 
However, its reconstruction on Cityscapes is still not satisfying and underperforms to our masked noise encoder.
As noted in \newcite{feature_style},
the feature size of the Feature-Style encoder is restricted. Using a larger feature map to improve reconstruction quality can only be done as a replacement of more latent code predictions. Consequently, it largely reduces the expressiveness of the latent embedding and leads to extremely poor editability, being no longer suitable for downstream applications, e.g., style mixing data augmentation.

The visual comparison across $\text{pSp}^\dagger$, the Feature-Style encoder and our masked noise encoder is shown in \cref{fig:encoder-visual} and is aligned with the quantitative results in \Cref{tab:gan-inversion-comparison}.  
$\text{pSp}^\dagger$ has overall poor reconstruction quality. The Feature-Style encoder cannot faithfully reconstruct small objects and restore fine details. In comparison, our masked noise encoder offers high-quality reconstruction, preserving the semantic layout and fine details of each class. Having a high-quality reconstruction is an important requirement for using the encoder for data augmentation. Unfortunately, neither $\text{pSp}^\dagger$ nor the Feature-Style encoder achieve satisfactory reconstruction quality. For instance, they both fail at capturing the red traffic light in \cref{fig:encoder-visual}. Using such images for data augmentation can confuse the semantic segmentation model, leading to performance degradation. 

\input{tables/encoder_comparison}
\input{tables/masking_noise_ablation}
\input{tables/mask_size_ratio} 
\input{tables/noise_map_ablation} 
\input{tables/big_table_hrnet_segformer}
\input{tables/big_table_hrnet_segformer_all}


\paragraph{Ablation on the masking effect} 
In \cref{fig:masking-ablation} and \cref{fig:intra-mix-example}, we visually observe that random masking offers a stronger perceivable style mixing effect compared to the model trained without masking. 
Next, we test the effect of masking on improving the domain generalization for the semantic segmentation task. In particular, we employ the encoder that is trained with and without masking to perform {\ourstyle}. In \cref{tab:mask-cs-da}, while slightly degrading the source domain performance of the baseline model on Cityscapes, {\ourstyle} improves the domain generalization performance on BDD100K, ACDC and Dark Z\"urich. As {\ourstyle} with masked noise encoder is more effective at diversifying the training set and reducing the style-content correlation, it achieves more pronounced gains in \cref{tab:mask-cs-da}, e.g., more than $10\%$ improvement in mIoU from Cityscapes to Dark Z\"urich. 

\paragraph{Ablation on masking hyperparameters} 
We conduct an ablation study on the mask patch size $P$ and masking ratio $\rho$, shown in \cref{tab:new-gan-inversion-mask-ablation}. 
\newchange{
We observe that the mask patch size is a relatively insensitive hyperparameter, while higher masking ratio results in noticeable degradation on the reconstruction quality.}
Empirically, the patch size $P=4$ with a masking ratio $\rho= 25\%$ achieves the best reconstruction performance. Therefore, we use the encoder trained with this parameter combination for our data augmentation {\ourstyle}.

\paragraph{Ablation on the noise map resolution}
We investigate the effect of noise map size and experimentally observed that the reconstruction quality benefits the most from using the noise map at the intermediate feature space with one fourth of the input resolution.
As shown in~\cref{tab:new-noise-map}, using $32 \times 64 $ noise, i.e., one fourth of the image resolution, achieves better reconstruction quality than using lower resolution noise maps.
Higher resolution noise map, e.g., full image resolution, in contrast, can be too expressive and encode nearly all perceivable details. This results in worse style mixing capability, as shown in \cref{fig:new-noise-res-comp2}. 
Therefore, we employ the intermediate noise map at one fourth of the input resolution in all of our experiments.

\input{figs/semseg_hrnet}
\input{figs/compare_StyleMix}

\subsection{ISSA for Domain Generalization}\label{sec:exp_dg}

\paragraph{Comparison with data augmentation methods}
\cref{tab:hrnet-segformer-cityscapes-dg} reports the mIoU scores of Cityscapes to ACDC domain generalization using two semantic segmentation models, i.e., HRNet{\hrnet} and SegFormer\segformer. Qualitative visualization is illustrated in \cref{fig:dg-semseg}.
{\ourstyle} is compared with three representative data augmentations methods, i.e., CutMix\newcite{yun2019cutmix}, Hendrycks's weather and digital corruptions\newcite{hendrycks2018benchmarking}, and StyleMix\newcite{hong2021stylemix}. Remarkably, our {\ourstyle} is the top performing method, consistently improving mIoU in both models and across all four different scenarios of ACDC, i.e., rain, fog, snow and night. Compared to HRNet, SegFormer is more robust against the considered domain shifts.

In contrast to the others, CutMix mixes up the content rather than the style. It improves the in-distribution performance on Cityscapes, but this gain does not extend to domain generalization. Hendrycks's weather corruptions can be seen as the synthetic version of Cityscapes under the rain, fog, and snow weather conditions. While already mimicking ACDC at training, it can still degrade ACDC-Snow by more than $5.8\%$ in mIoU using HRNet. 
\newchange{Among the four Hendrycks' corruption types (i.e., noise, blur, digital and weather), Hendrycks-Digital, consisting of contrast, elastics transformation, pixelation and JPEG, is the best-performing one, but still underperforms ISSA.} 
StyleMix\newcite{hong2021stylemix} also seeks to mix up styles. However, it does not work well for scene-centric datasets, such as Cityscapes. Its poor synthetic image quality (see \cref{fig:stylemix-sample-1}) leads to the performance drop over the HRNet baseline in many cases, e.g., on Cityscapes to ACDC-Fog from $58.68\%$ to $49.11\%$ mIoU.

\newchange{
More evaluation on the generalization performance from Cityscapes to BDD100K and Dark Z\"urich is provided in \cref{tab:hrnet-segformer-cityscapes-all-dg-appendix}, where the observation is consistent with \cref{tab:hrnet-segformer-cityscapes-dg} explained above.
In addition to weather changes, we further compare different data augmentation methods under the more challenging day-to-night setting in \cref{tab:hrnet-bdd-daytime-da}. {\ourstyle} present consistent advantages over competing methods, which again justifies the effectiveness of {\ourstyle} on improving generalization performance. 
}

\input{tables/hrnet_bdd100k_daytime} 
\input{tables/dg_bn_aug}
\input{tables/robustnet}

\input{tables/uda_comparison} 

\input{figs/landscape_sample} 
\input{figs/interpolation} 

\paragraph{Comparison with domain generalization techniques}
We further compare {\ourstyle} with two advanced feature space style mixing methods designed to improve domain generalization performance: MixStyle\newcite{zhou2021mixstyle} and DSU\newcite{li2022dsu}. Both extract the style information at certain normalization layers of CNNs. MixStyle\newcite{zhou2021mixstyle} mixes up styles by linearly interpolating the feature statistics, i.e., mean and variance, of different images, while DSU\newcite{li2022dsu} models the feature statistics as a distribution and randomly draws samples from it. 

We adopt the experimental setting of DSU with default hyperparameters, using DeepLab v2\newcite{chen2017deeplabv2}
segmentation network with ResNet101 backbone.
\cref{tab:feature-aug-da} shows that {\ourstyle} outperforms both MixStyle and DSU by a large margin. 
We also observe that there is a slight performance drop on the source domain (i.e., CS) when applying DSU and MixStyle.
As they operate at the feature-level, there is no guarantee that the semantic content stays unchanged after the random perturbation of feature statistics. Thus, the changes in feature statistics might negatively affect the performance, as also indicated in\newcite{li2022dsu}. Note that, in contrast, {\ourstyle} operates on the image space.
Combining {\ourstyle} with MixStyle and DSU leads to a strong boost in performance of these methods.

Being model-agnostic, {\ourstyle} can be combined with other networks designed specifically for the domain generalization of semantic segmentation. 
\newchange{
To showcase its complementary nature, we add {\ourstyle} on top of two state-of-the-art domain generalization methods for semantic segmentation, RobustNet\newcite{robustnet_2021} and SHADE\newcite{shade-style-hall}. RobustNet proposed a novel instance whitening loss to selectively remove domain-specific style information. SHADE on the other hand aims to learn style-invariant representation and preserve knowledge from the pretrained backbone.
Although color transformation has already been used for augmentation in both methods and SHADE additionally employs feature-level style augmentation, 
{\ourstyle} can introduce more natural style shifts, thus is able to bring further improvements.
\Cref{tab:robustnet-cs-da} verifies the effectiveness of {\ourstyle}, which brings extra gains for RobustNet and SHADE. For RobustNet, the performance of the challenging day to night scenario, i.e., Cityscapes to Dark Z\"urich is boosted from $20.11\%$ to $23.09\%$ in mIoU. 
}

\paragraph{Comparison with unsupervised domain adaptation methods}\label{sec:appendix-uda}
We compare our method with multiple unsupervised domain adaptation (UDA) techniques, which not only have access to the source domain, but also use extra unlabeled samples of the target domain. The quantitative comparison of Cityscapes to ACDC adaptation/generalization is shown in \Cref{tab:uda-comparison}. Our method has presented competitive performance, even without using images from the target domain. 
\input{figs/acdc_stylized} 
\input{tables/bddgan_dg} 
\input{tables/landscape_style_dg} 

\section{Plug-n-Play Ability of the Exemplar-Based Style Synthesis Pipeline}\label{sec:new_pnp}
In \cref{sec:exp_dg}, we have focused on ISSA for improved domain generalization. Next, we investigate the plug-n-play ability of our exemplar-based style pipeline, which enables ESSA. Specifically, the generator and masked noise encoder which are trained on one dataset can be directly used for mixing styles from other datasets, thus avoiding retraining or fine-tuning the models. %
This ability is valuable in two perspectives: 1) harnessing external data for improved domain generalization via ESSA; and 2) saving computationally complexity. Compared to other data augmentation techniques such as CutMix\newcite{yun2019cutmix}, Hendrycks corruption\newcite{hendrycks2018benchmarking}, our style synthesis requires training GAN and an encoder, which could take considerable computational resources.
Therefore, it is of practical interest if the trained models can be readily useable for novel domains.

\paragraph{ISSA using arbitrary encoders} 
Favorably, thanks to the plug-n-play ability of the synthesis pipeline, 
we observe that ISSA can still be effective even when encoder and generator are trained on a different dataset of a similar task, and re-training is not required. Note that here the source is with respect to the segmenter training for domain generalization, not the encoder training.
As shown in \cref{tab:bddgan-cityscapes-dg-small}, when training the segmenter on Cityscapes using ISSA, we can directly use generator and encoder trained on BDD100K without fine-tuning. Even though the models have not seen any samples of Cityscapes, they can still reconstruct and augment styles within Cityscapes, and the effectiveness of ISSA is not compromised. This implies that, once the generator and encoder are trained on one dataset, they are also straightforwardly applicable for augmenting novel datasets.

\paragraph{Extra-source exemplar based style synthesis} 
Furthermore, we exploit the usage of extra-source data as the style exemplar. 
Visual examples in \cref{fig:landscape-example} showcase the plug-n-play style-mixing ability of our encoder on web-crawled images, where the model is only trained on Cityscapes. 
It can be observed that the style of unseen images can still be successfully transferred to the content images, which grants us the opportunity to further utilize images on the web to enhance the effectiveness of style augmentation beyond intra-source styles. Also, we illustrate the interpolation capability in the style latent space on both trained Cityscapes and unseen web-crawled image. This property enables more control on the style mixing strength.

To further explore the usage of images on the web, we take Landscape Pictures\footnote{\url{https://www.kaggle.com/datasets/arnaud58/landscape-pictures?resource=download}} dataset as the extra-source exemplars for style augmentation. \Cref{tab:ladnscape-dg} justifies that by exploiting additional image styles, ESSA can further improve the generalization performance of ISSA on unseen target domains.

\input{figs/performance_validation_id} 
\input{figs/performance_validation} 
\section{Stylized Proxy Validation Set Synthesis}\label{sec:new_correlation}
Beyond the usage of data augmentation for network training, we further explore if our exemplar-based style synthesis pipeline can be used to assess the generalization capability of semantic segmentation models for both source and target domain without extra data annotation effort. 
Prior work\newcite{zhang2021predictingGAN} has used conditional GAN synthesized samples to predict  generalization performance of image classifiers in the source domain. However, it remains unclear how to 
evaluate the generalization performance on unseen domains, and apply it on dense prediction tasks.
Given the fact that our masked noise encoder can transfer styles even from novel domains, we utilize this attractive property to generate a stylized proxy validation set, i.e., combining styles from the target domain with the contents from the source domain training samples. For getting their styles, exemplars from the target domain do not need to be labelled. The existing ground-truth label maps of the training samples in the source domain are reused as the ground-truth annotations of the stylized proxy validation set.
Visual examples of transferring ACDC style using one sample from each weather condition are provided in \cref{fig:stylized_proxy}.

\paragraph{Experimental Setup} 
We investigate the generalization performance of $95$ semantic segmentation models trained on Cityscapes, where $54$ models are obtained from MMSegmentation\newcite{mmseg2020} model zoo and the others are trained by ourselves. The models cover both CNN-based architectures, e.g., HRNet\hrnet, DeepLab\newcite{chen2017deeplabv3},  DANet\newcite{fu2019danet}, and transformer-based model, e.g., SegFormer\newcite{xie2021segformer}, SETR\newcite{zheng2021setr}. Besides, the models are trained using different strategies, e.g., various learning rate schedule, cropping size and data augmentation.
We consider generalization performance on both source and target domain for the correlation study. Specifically, we use the Cityscapes validation set as the source test set, ACDC and BDD100K validation sets as the target test data. To verify the generalization performance on the source domain, we apply intra-source style augmentation on the Cityscapes training set and use it as the proxy validation set. For the verification of target domain generalization performance, we build a proxy set by transferring styles from the corresponding target test dataset. Further, we study the correlation between the real test performance and performance on the proxy data.

\paragraph{Correlation Metrics} 
We compute Spearman’s Rank Correlation coefficient ($\rho$) and Kendall Rank Correlation Coefficient ($\tau$) to quantitatively measure the correlation strength.  The value of the correlation coefficient varies from $[-1, 1]$. A value closer to $\pm 1$ indicates strong positive/negative association between the two variables. As the coefficient goes towards $0$, the association becomes looser. Both correlation coefficients are non-parametric, i.e., no strict assumptions on the data distribution, and the assessment is based on the ranking of the data.

\paragraph{Observations} 
In \cref{fig:ID-correlation}, we show the correlation of performance on the intra-source style augmented proxy set and real Cityscapes test set across different network architectures. 
We clearly observe a strong correlation ($\rho > 0.95$), indicating that ISSA proxy set can serve as a good indicator for generalization in the source domain. 

Furthermore, we report the correlation results of target domain generalization on two datasets, i.e., ACDC and BDD100K in each row of \cref{fig:performance_correlation}. We compare
three different choices of the proxy set in each column, namely the original Cityscapes validation set, intra-source style augmented Cityscapes validation set and target data style augmented validation set.
\textcolor{matplotlibBlue}{Blue} and \textcolor{matplotlibOrange}{orange} dots represent CNN- and transformer-based backbones, respectively. 
Quantitatively, the correlation coefficients of \cref{csval-acdc,csval-bdd} are rather low. Also from \cref{csval-acdc}, some blue points in the upper right corner has stronger performance on Cityscapes validation set compared to the orange points, but worse on ACDC test data. This suggests that evaluation of the original Cityscapes (source) validation set cannot properly reflect the generalization performance on the target domain. Therefore, this raises the concern that by following the traditional way, selecting the best model based on the source validation performance could be problematic when the deploying environment involves data of unknown target domains. 
By applying intra-source style augmentation on the Cityscapes validation set, the correlation coefficient has been improved (see \cref{intra-acdc,intra-bdd}). We hypothesize that style mixing results in better data coverage and thus can better represent model's generalization ability under style shifts.
Furthermore, whenever it is possible to have access to images of the target domain, even though without annotation, we can utilize styles of the unlabeled target data and achieve the strongest correlation in \cref{acdc-acdc,bdd-bdd}.
In addition to the correlation metric, in general models have higher mIoU on the Cityscapes validation set, compared with the intra-source style and target domain style augmented proxy set. And the mIoU range on the intra-source proxy set is closer to the one of using target styles, which also justifies our hypothesis above.

Besides, we also observe an interesting phenomenon from \cref{fig:performance_correlation}: all transformer-based models (\textcolor{matplotlibOrange}{orange dots}) are above the linear fit. This suggests that transformer-based models present better generalization ability under natural shifts compared with CNN-based models (\textcolor{matplotlibBlue}{blue dots}). This is consistent with the acknowledgement on transformers from prior works \newcite{naseer2021vitintriguing,bai2021transformersRobust,zhang2022delving}.

To sum up, we present a new use case of proposed exemplar-based style synthesis pipeline, and demonstrate that stylized samples can be used as a proxy validation set and a strong indicator for model's generalization capability without introducing additional annotation efforts. Based on this observation, we can better utilize existing annotated data together with our exemplar-based style synthesis pipeline, to select models in practice especially when deployment in an open-world environment, where unknown target data commonly exists.

%% file: figs/noise_map_res_ablation.tex
\begin{figure*}[t]
\begin{centering}
\setlength{\tabcolsep}{0.0em}
\renewcommand{\arraystretch}{0}
\par\end{centering}
\begin{centering}
\hfill{}
	\begin{tabular}{@{}c@{}c@{}c@{}c@{}c}
			\centering
	Content & Style 
	& $\frac{H}{16}\times \frac{W}{16}$  
	& $\frac{H}{4}\times \frac{W}{4}$ (Ours)
	& $H\times W$
	\vspace{0.02cm} \tabularnewline
	\includegraphics[width=0.195\linewidth]{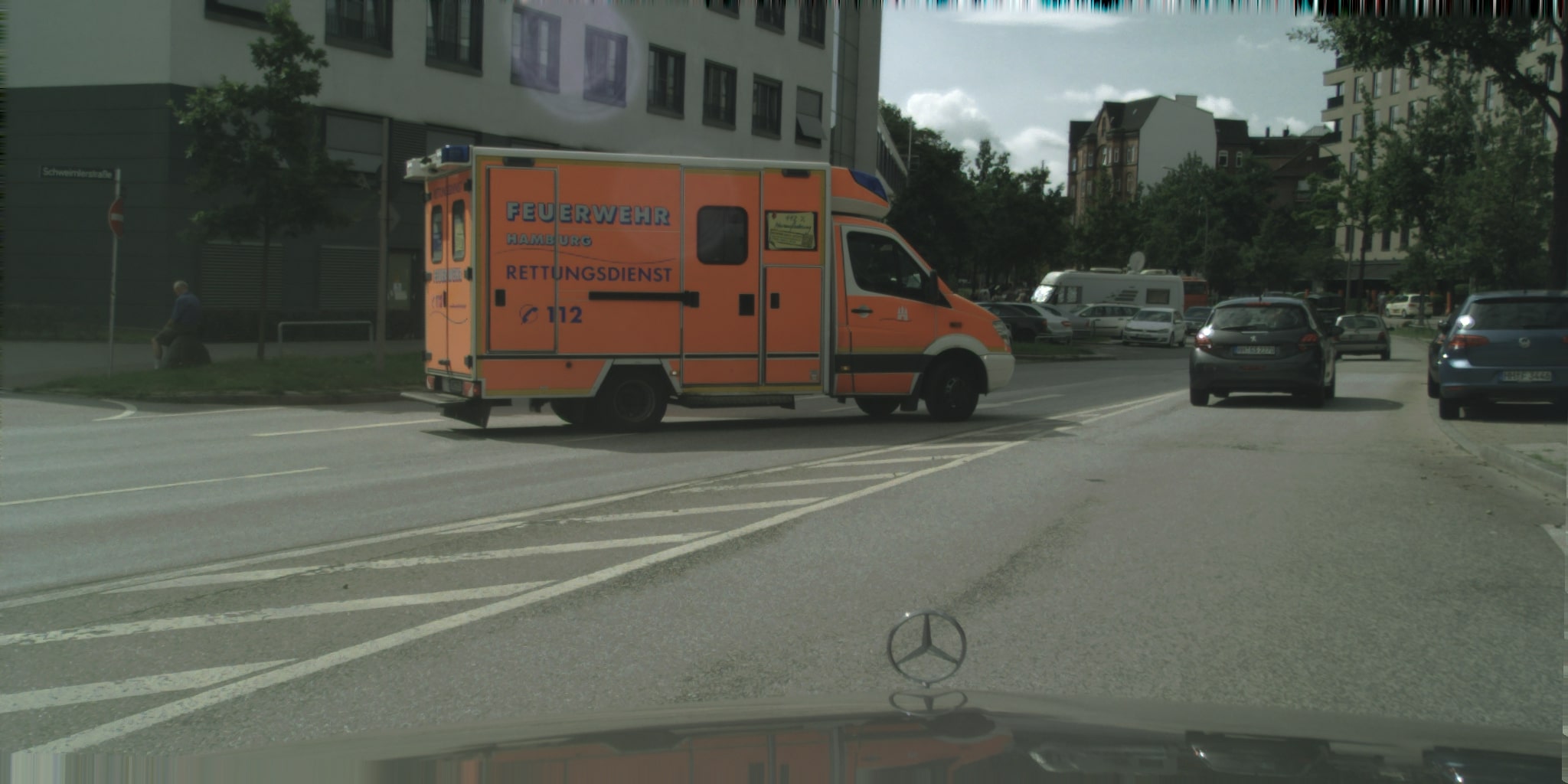}
	      & {\footnotesize{}}
	\includegraphics[width=0.195\linewidth]{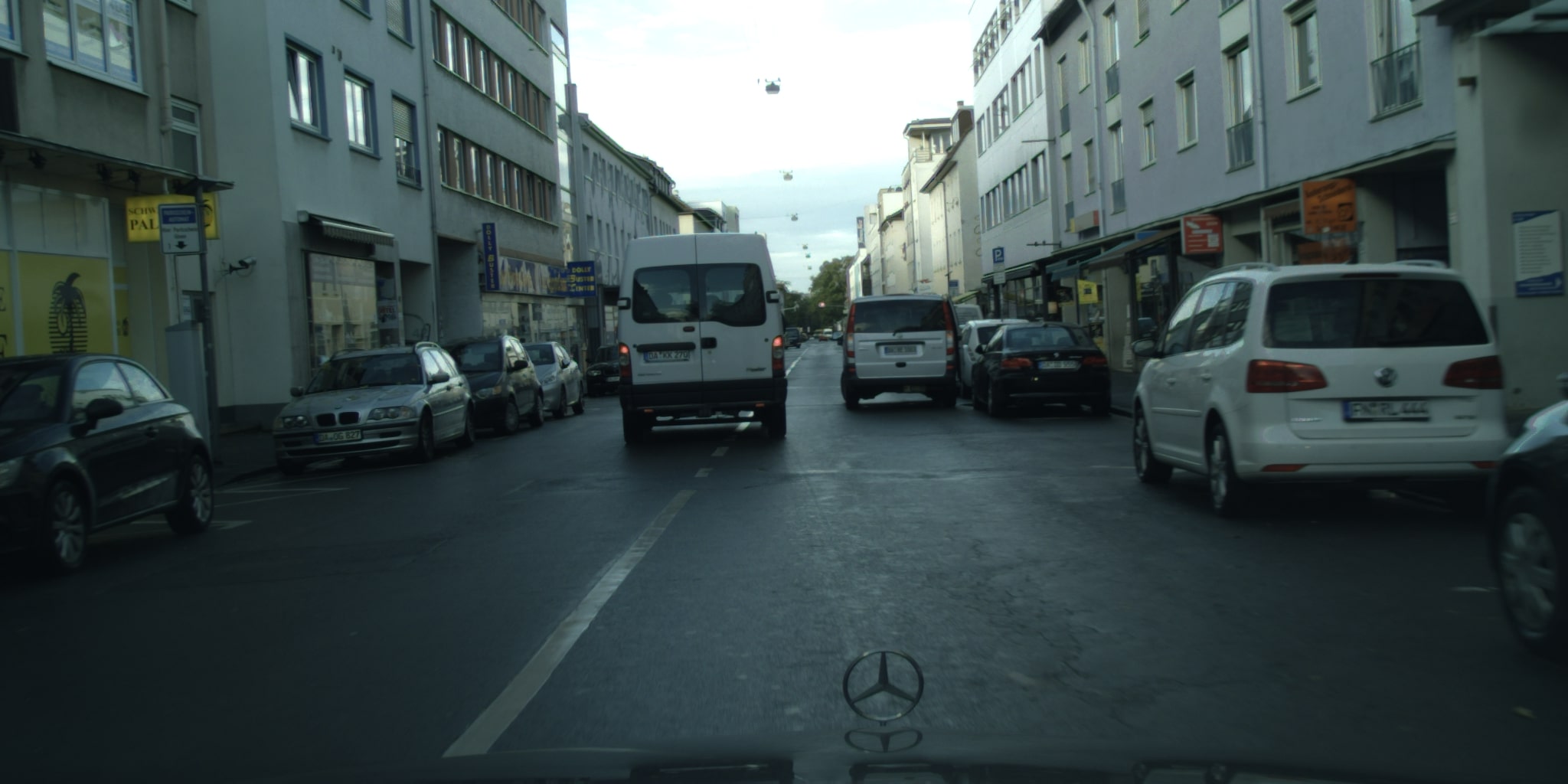}
	        & {\footnotesize{}}
	\includegraphics[width=0.195\linewidth]{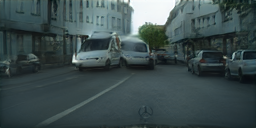} 
		    & {\footnotesize{}}
	\includegraphics[width=0.195\linewidth]{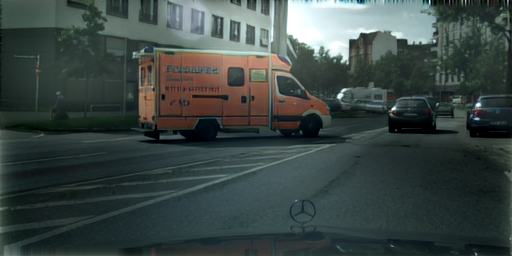}
		& {\footnotesize{}}
	\includegraphics[width=0.195\linewidth]{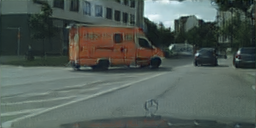}
	 \tabularnewline
	\end{tabular}
\hfill{}
\par\end{centering}
\caption{Influence of the  noise map resolution on style-mixing ability. Using higher resolution noise map, e.g., $H\times W$, leads to poor style-mixing ability. While too low resolution, e.g., $\frac{H}{16}\times \frac{W}{16}$, cannot reconstruct the scene faithfully.
} 
\label{fig:new-noise-res-comp2}
\end{figure*}

%% file: tables/encoder_comparison.tex
\begin{table}[t]
\begin{center}
{\small
    {
    \begin{tabular}{l|lcc}
        Method & MSE $\downarrow$ & LPIPS $\downarrow$  & FID $\downarrow$  \\
    \midrule
        pSp\newcite{pspencoder} & 0.078  & 0.348 & 130.62  \\
        pSp${}^\dagger$\newcite{pspencoder} & 0.049  & 0.339 & 14.60  \\
        Feature-Style\newcite{feature_style} & 0.025  & 0.220 & 7.14   \\
        \textbf{Ours} & \textbf{0.011}  & \textbf{0.124} & \textbf{3.94}  \\
    \end{tabular}
    }
}
\end{center}
\caption{Reconstruction quality on Cityscapes at the resolution $128\times 256$. MSE, LPIPS\newcite{zhang2018lpips} and FID\newcite{heusel2017fid} respectively measure the pixel-wise reconstruction difference, perceptual difference, 
and distribution difference between the real and reconstructed images. The proposed masked noise encoder (Ours) consistently outperforms pSp, $\text{pSp}^\dagger$ and the feature-style encoder. Note, $\text{pSp}^\dagger$ is introduced by us, by training pSp with an additional discriminator and incorporating synthesized images for better initialization. 
}
\label{tab:gan-inversion-comparison}
\end{table}

%% file: tables/masking_noise_ablation.tex
\begin{table}[t]
\begin{center}
{\small
    {
    \begin{tabular}{@{}l|c|cccc@{}}
        Method & CS & ACDC & BDD & DarkZ  \\ \midrule
        Baseline   & \textbf{70.47}  & 41.48 & 45.66  & 15.25  \\
        {\ourstyle} w/o masking  & 69.68 & 44.63 & 46.45 & 17.36  \\
        {\ourstyle} w/- masking  & 69.48  &\textbf{ 47.43} & \textbf{47.87} & \textbf{26.10}  \\
    \end{tabular}
    }
}
\end{center}
\caption{The effect of random noise masking on improving domain generalization via {\ourstyle}. We report the mean Intersection over Union (mIoU) of HRNet\newcite{wang2020hrnet} trained on Cityscapes at the resolution $256\times 512$. BDD100K (BDD), ACDC, and Dark Z\"urich (DarkZ) represent different domain shifts from Cityscapes.}
\label{tab:mask-cs-da}
\end{table}

%% file: tables/mask_size_ratio.tex
\begin{table}[t]
\begin{center}
{\small
    \begin{tabular}{@{}cc|ccc@{}}
        \multicolumn{1}{l}{Patch size} & Ratio & MSE $\downarrow$ & LPIPS $\downarrow$ & FID $\downarrow$ \\ \midrule
        \multirow{2}
        {*}{2} & 25\% & 0.005 & 0.090 & 1.50 \\
               & 50\% & 0.008 & 0.127 & 2.02 \\
        \midrule
        \multirow{2}
        {*}{4} & 25\% & \textbf{0.004} & \textbf{0.089} & \textbf{1.41} \\
               & 50\% & 0.009 & 0.129 & 2.01 \\ 
    \end{tabular}
}
\end{center}
\caption{Ablation on the mask patch size and masking ratio. The influence of patch size on the reconstruction is minor, while masking ratio is more important, i.e., higher masking ratio has negative impact. }
\label{tab:new-gan-inversion-mask-ablation}
\end{table}

%% file: tables/noise_map_ablation.tex
\begin{table}[t]
\begin{center}
{\small
    {
    \begin{tabular}{l|lcc}
        Noise scale & MSE $\downarrow$ & LPIPS $\downarrow$  & FID $\downarrow$  \\
    \midrule
        $4\times8 \sim 8\times16$ & 0.041 & 0.317 & 14.90  \\
        $32\times64$ & 0.008  & 0.101 & 2.30  \\
    \end{tabular}
    }
}
\end{center}
\caption{Effect of noise map resolution on reconstruction quality. Experiments are done on Cityscapes, $128\times 256$ resolution.
}
\label{tab:new-noise-map}
\end{table}

%% file: tables/big_table_hrnet_segformer.tex
\begin{table*}[t]
\begin{center}
{\small 
    \begin{tabular}{@{}l|c|ccccc||c|ccccc @{}}
         & \multicolumn{6}{c||}{HRNet\hrnet} & \multicolumn{6}{c}{SegFormer\newcite{xie2021segformer}} \\ 
        Method & CS & Rain & Fog & Snow & Night & Avg. & CS & Rain & Fog & Snow & Night & Avg.\\ \midrule
        Baseline  & 70.47 & 44.15 & 58.68 & 44.20 & 18.90 & 41.48  & 67.90 & 50.22 & 60.52 & 48.86 & 28.56 & 47.04\\ 
        \midrule
        ColorTransform &  69.90 & 49.35 & {65.14} & {52.63} & {26.56} & {48.42} &
        {68.50} & {51.58} & {66.45} & {52.87} & {30.33} & {50.31}
        \\
        CutMix\newcite{yun2019cutmix}   & \textbf{72.68} & \underline{42.48} & \underline{58.63} & 44.50 & \underline{17.07} & \underline{40.67} & \textbf{69.23} & \underline{49.53} & 61.58 & \underline{47.42} & \underline{27.77} & \underline{46.57} \\
        Hendrycks-Weather 
        & 69.25 & \textbf{50.78} & 60.82 & \underline{38.34} & 22.82 & 43.19  & 67.41 & 54.02 & 64.74 & 49.57 & 28.50 & 49.21 \\
        Hendrycks-Digital & 69.13 & 50.13 & 65.71 & 49.22 & 24.81 & 47.47 & 67.57 & 55.53 & 66.46 & 49.92 & 30.33 & 50.56\\
        {
        FDA\newcite{yang2020fda}} & {70.43} & {49.68} & {65.19} & {50.65} & {26.41} & 
        {47.98} &  {67.92} &
        {51.28} & {67.03} & 
        {51.30} & 
        {28.28} & 
        {49.47}  
        \\
        StyleMix\newcite{hong2021stylemix} & 57.40 & \underline{40.59} & \underline{49.11}  & \underline{39.14} & 19.34 & \underline{37.04} & 65.30 & 53.54 & 63.86 & 49.98 & 28.93 & 49.08\\ 
        \textbf{{\ourstylebf} (Ours)}    & 70.30 & 50.62 & \textbf{66.09} & \textbf{53.30} & \textbf{30.18} & \textbf{50.05} & 67.52 & \textbf{55.91} & \textbf{67.46} & \textbf{53.19} & \textbf{33.23} & \textbf{52.45}  \\
        \midrule
        Oracle & 70.29 & 65.67 & 75.22 & 72.34 & 50.39 & 65.90  & 68.24 & 63.67 & 74.10 & 67.97 & 48.79 & 63.56 \\
    \end{tabular}
}
\end{center}
\caption{
Comparison of data augmentation for improving domain generalization, i.e., from Cityscapes (train) to ACDC (unseen). The mean Intersection over Union (mIoU) is reported on Cityscapes (CS), four individual scenarios of ACDC (Rain, Fog, Snow and Night) and the whole ACDC (Avg.).
{
ColorTransform consists of various color transformations such as altering the contrast, brightness, saturation; luma flip and hue rotation.
}
Hendrycks-Weather\newcite{hendrycks2018benchmarking} simulates weather conditions in a synthetic manner for data augmentation, and Hendrycks-Digital is composed of contrast, elastics transformation, pixelation and JPEG corruption. 
Oracle indicates the supervised training on both Cityscapes and ACDC, serving as an upper bound on ACDC for the other methods. Note, it is not supposed to be an upper bound on Cityscapes. Underline denotes worse results than the baseline on ACDC. {\ourstyle} performs the best and consistently improves the mIoU in all four scenarios of ACDC using both HRNet and SegFormer.
}
\label{tab:hrnet-segformer-cityscapes-dg}
\end{table*}

%% file: tables/big_table_hrnet_segformer_all.tex
\begin{table*}[t]
\begin{center}
{\small 
    \begin{tabular}{@{}l|c|ccc||c|ccc @{}}
         & \multicolumn{4}{c||}{HRNet\hrnet} & \multicolumn{4}{c}{SegFormer\newcite{xie2021segformer}} \\ 
        Method & CS & ACDC & BDD100K & Dark Z\"urich  & CS & ACDC & BDD100K & Dark Z\"urich \\ \midrule
        Baseline & 70.47 & 41.48  & 45.66 & 15.50 & 67.90 &  47.04 & 49.35 & 24.20 \\ 
        \midrule
        {ColorTransform} &  {69.90} & {48.42} & {50.22} & {24.13} & {68.50} & {50.31} & {51.09}
        & {25.04}
        \\
        CutMix\newcite{yun2019cutmix}   & \textbf{72.68} & 40.67 &45.57 & 15.34 & \textbf{69.23} & 46.57 & 48.93 & 22.98\\
        Hendrycks-Weather
        & 69.25 & 43.19 & 44.53  & 18.71   & 67.41 & 49.21 & 49.84 & 23.44 \\
        Hendrycks-Digital
        & 69.13 &  47.47 & 47.60 & 22.32 & 67.57 & 50.56 & 51.11 & 25.11\\
        {
        FDA\newcite{yang2020fda}} & {70.43} & {47.98} & {48.74} &  {22.46} & 
        {67.92} &
        {49.47} &
        {50.47} &
        {22.45} 
        \\
        StyleMix\newcite{hong2021stylemix} & 57.40 & 37.04 & 39.30 & 15.85 & 65.30 &  49.08 & 50.49 & 23.50 \\ 
        {\ourstylebf} \textbf{(Ours)}    & 70.30 &  \textbf{50.05} & \textbf{50.29} & \textbf{27.24} & 67.52 & \textbf{52.45} & \textbf{51.92} & \textbf{27.39}\\
    \end{tabular}
}
\end{center}
\caption{
Comparison of data augmentation for improving domain generalization, i.e., from Cityscapes (train) to ACDC, BDD100K and Dark Z\"urich (unseen). 
{\ourstyle} consistently outperforms the other data augmentation techniques across different datasets and network architectures, which is consistent with the \cref{tab:hrnet-segformer-cityscapes-dg}.
}
\label{tab:hrnet-segformer-cityscapes-all-dg-appendix}
\end{table*}

%% file: figs/semseg_hrnet.tex
\begin{figure*}[t]
\begin{centering}
\setlength{\tabcolsep}{0.0em}
\renewcommand{\arraystretch}{0}
\par\end{centering}
\begin{centering}
\hfill{}%
	\begin{tabular}{@{}c@{}c@{}c@{}c}
			\centering
		 Image  & Ground truth  & Baseline & Ours \vspace{0.01cm} \tabularnewline
	\includegraphics[width=0.241\textwidth,height=0.1205\textwidth,]{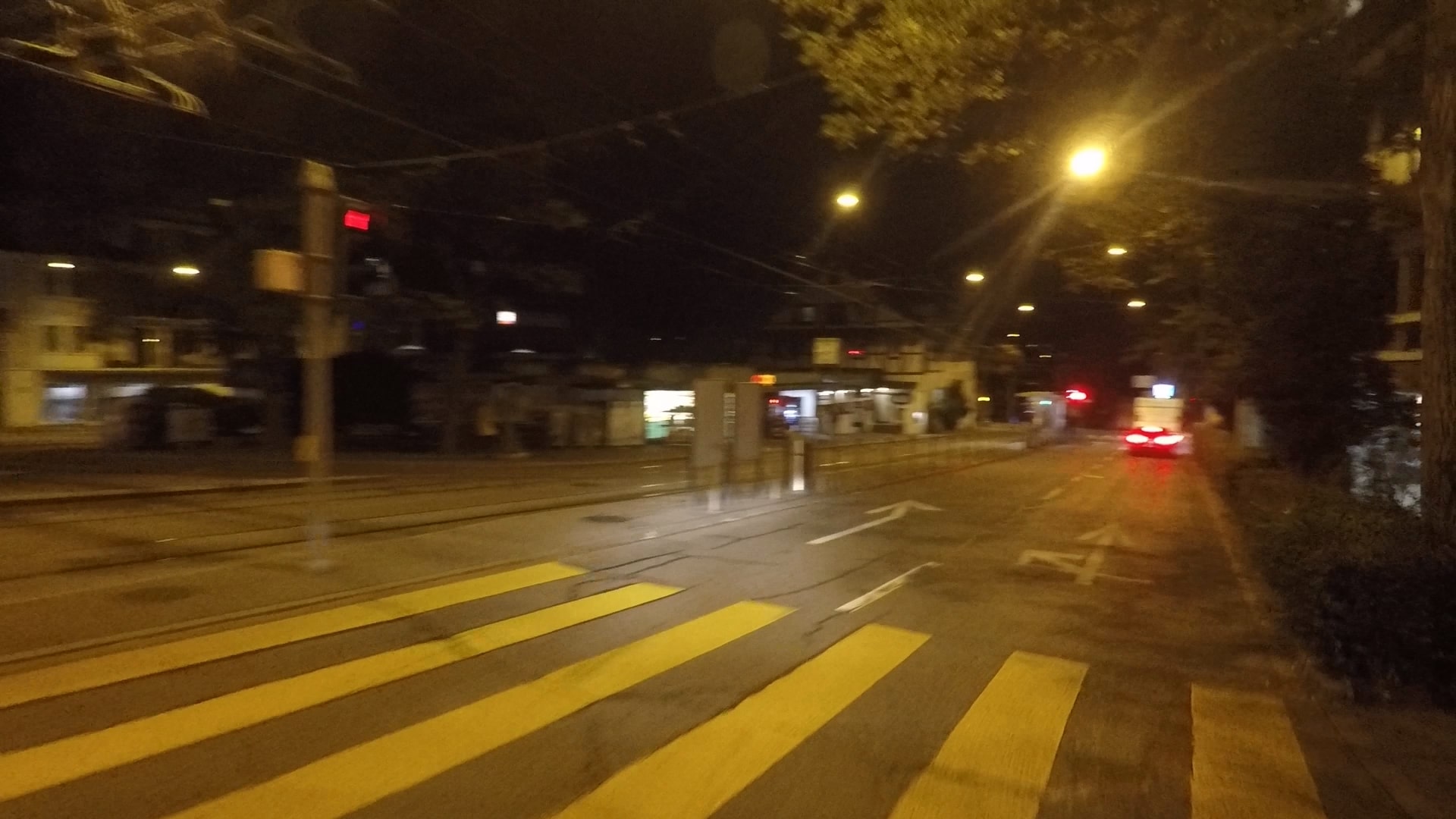} 
	        & {\footnotesize{}}
	  \begin{tikzpicture}
            \node [
	        above right,
	        inner sep=0] (image) at (0,0) {\includegraphics[width=0.241\textwidth,height=0.1205\textwidth]{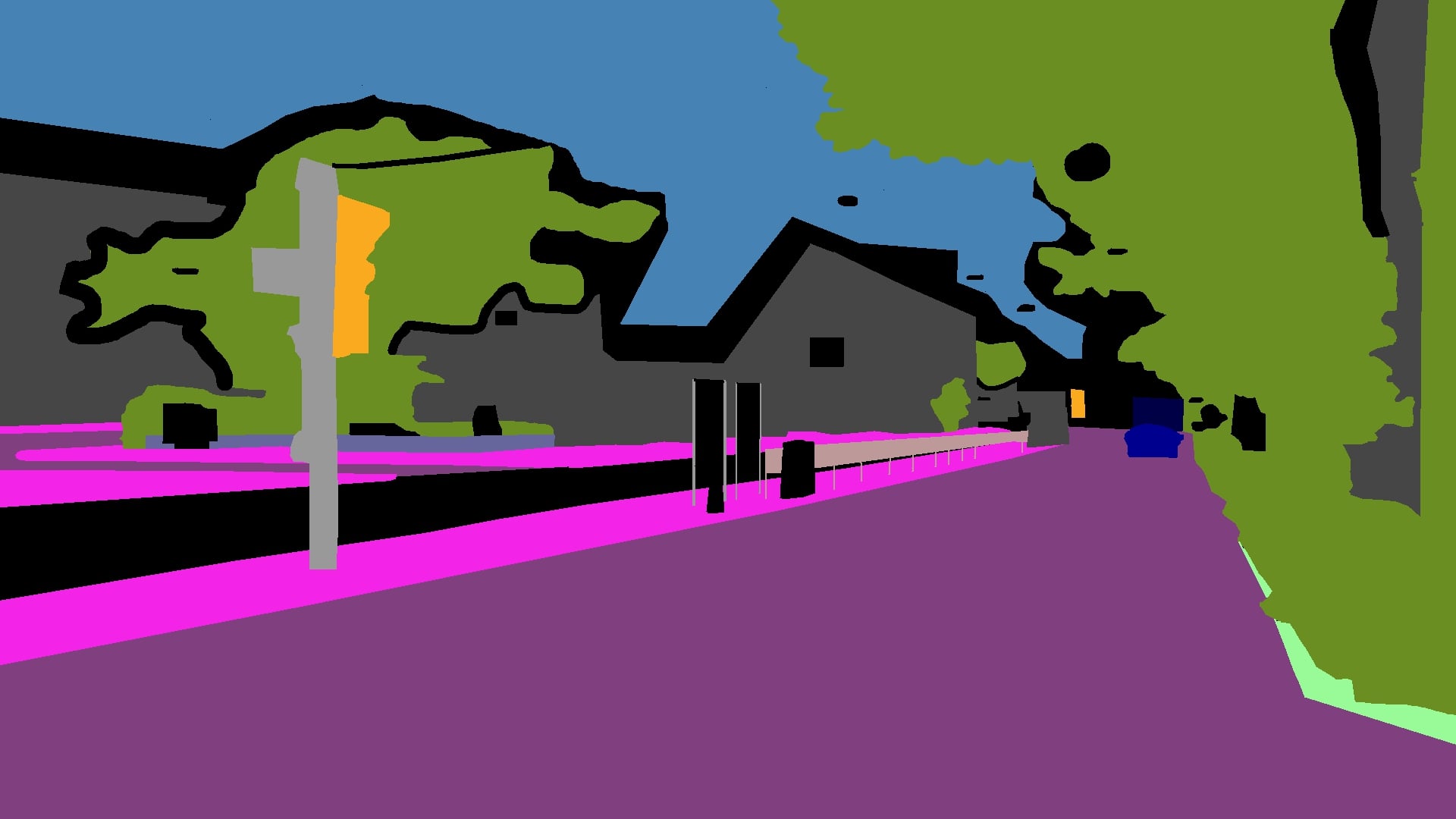} };
            \begin{scope}[
            x={($0.1*(image.south east)$)},
            y={($0.1*(image.north west)$)}]
            \draw[thick,green] (1.5,2.5) rectangle (3,8.4) ;
        \end{scope}
    \end{tikzpicture}
	        & {\footnotesize{}}
	  \begin{tikzpicture}
            \node [
	        above right,
	        inner sep=0] (image) at (0,0) {\includegraphics[width=0.241\textwidth,]{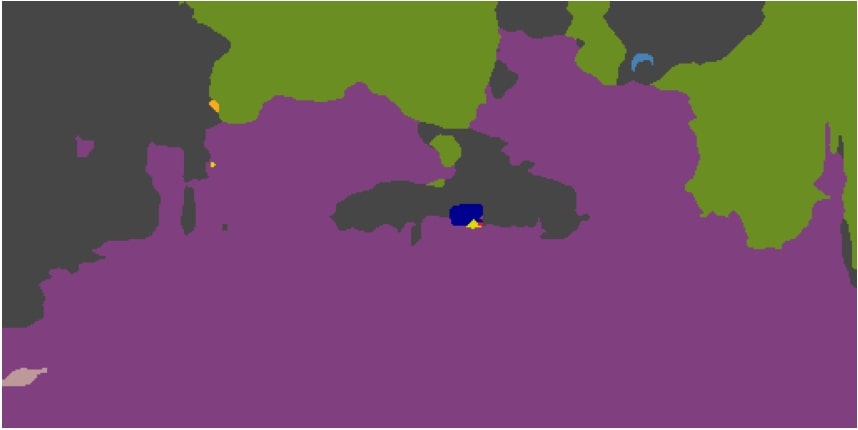} };
            \begin{scope}[
            x={($0.1*(image.south east)$)},
            y={($0.1*(image.north west)$)}]
            \draw[thick,red] (1.5,2.5) rectangle (3,8.4) ;
        \end{scope}
    \end{tikzpicture}
		    & {\footnotesize{}}
    \begin{tikzpicture}
            \node [
	        above right,
	        inner sep=0] (image) at (0,0) {\includegraphics[width=0.241\textwidth,]{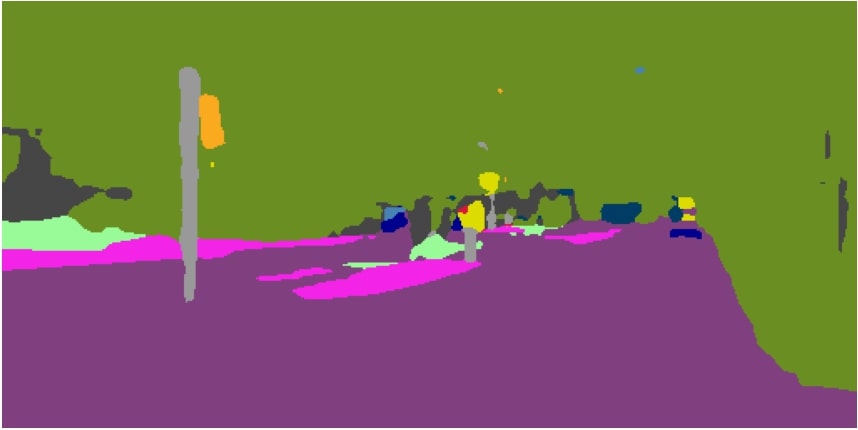}  };
            \begin{scope}[
            x={($0.1*(image.south east)$)},
            y={($0.1*(image.north west)$)}]
            \draw[thick,green] (1.5,2.5) rectangle (3,8.4) ;
        \end{scope}
    \end{tikzpicture}
	 \tabularnewline	
	 
	 	\includegraphics[width=0.241\textwidth,height=0.1205\textwidth]{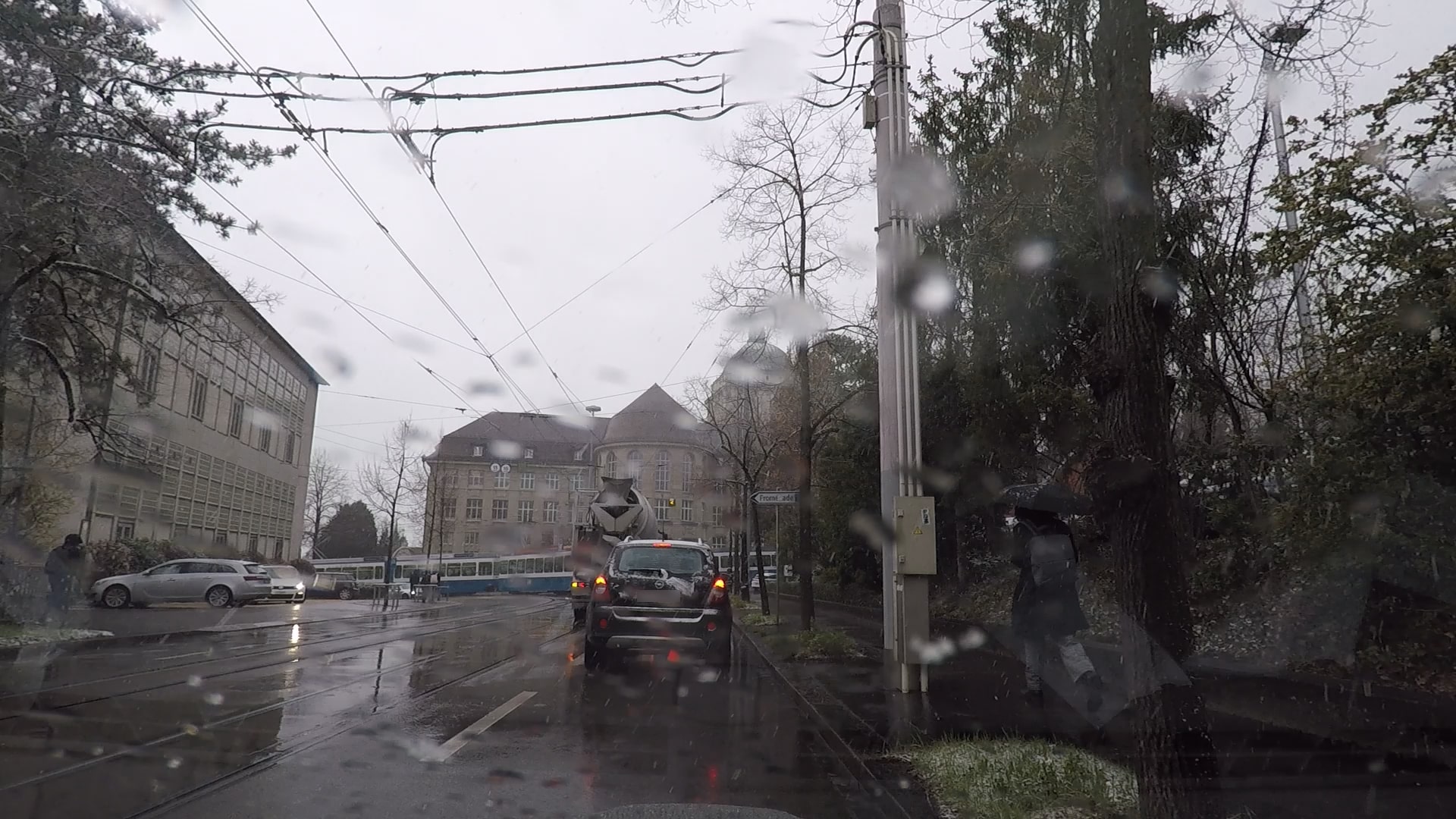} 
	        & {\footnotesize{}}
	  \begin{tikzpicture}
            \node [
	        above right,
	        inner sep=0] (image) at (0,0) {\includegraphics[width=0.241\textwidth,height=0.1205\textwidth]{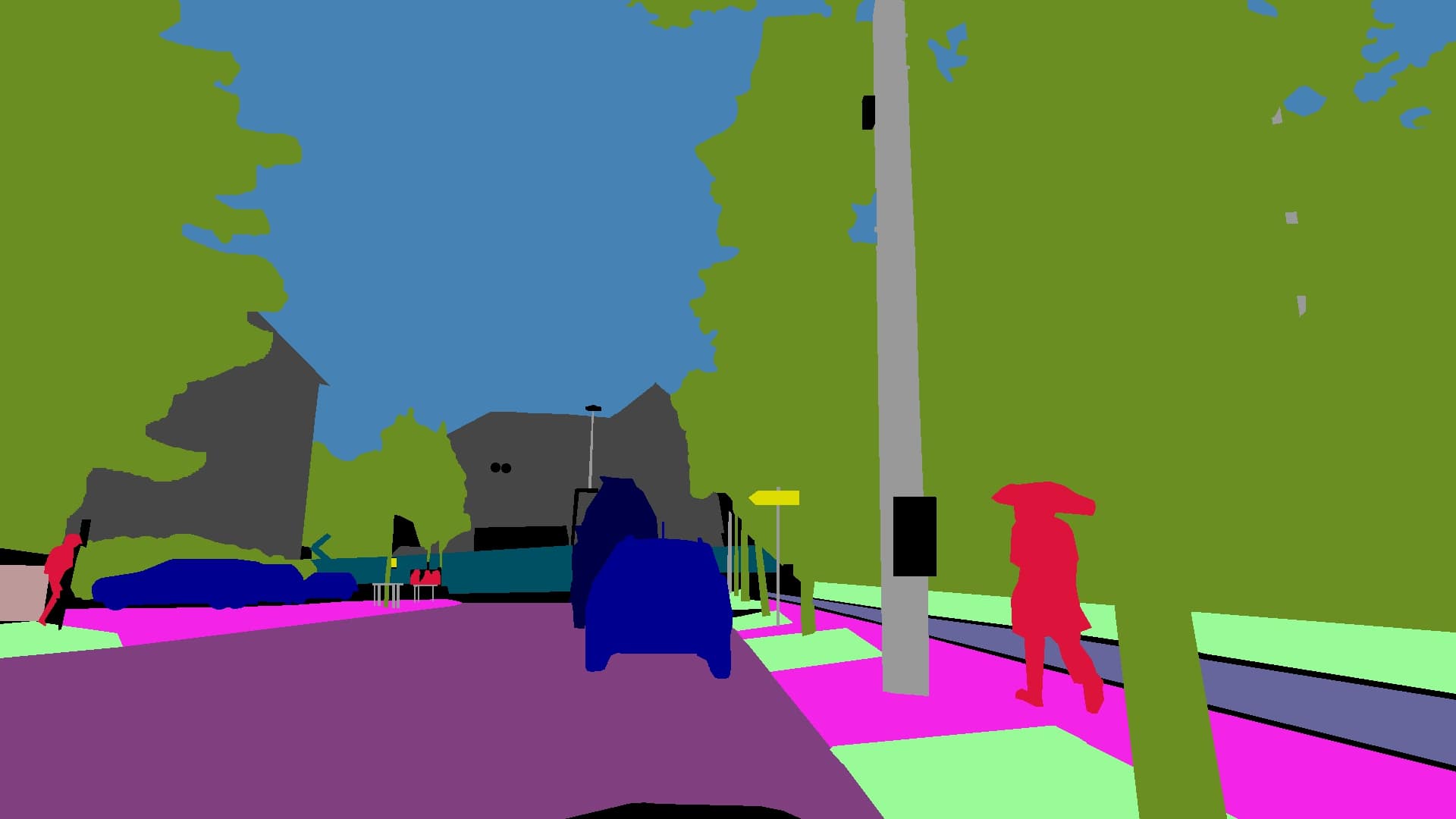}};
            \begin{scope}[
            x={($0.1*(image.south east)$)},
            y={($0.1*(image.north west)$)}]
            \draw[thick,green] (6.5,1.0) rectangle (8,4.5) ;
        \end{scope}
    \end{tikzpicture}
	   & {\footnotesize{}}
	   \begin{tikzpicture}
            \node [
	        above right,
	        inner sep=0] (image) at (0,0) {\includegraphics[width=0.241\textwidth,]{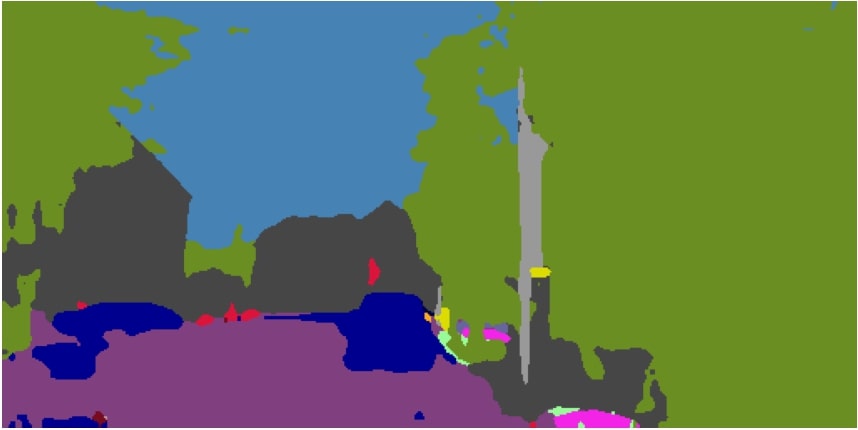}  };
            \begin{scope}[
            x={($0.1*(image.south east)$)},
            y={($0.1*(image.north west)$)}]
            \draw[thick,red] (6.5,1.0) rectangle (8,4.5) ;
        \end{scope}
    \end{tikzpicture}
		    & {\footnotesize{}}
	\begin{tikzpicture}
            \node [
	        above right,
	        inner sep=0] (image) at (0,0) {\includegraphics[width=0.241\textwidth,]{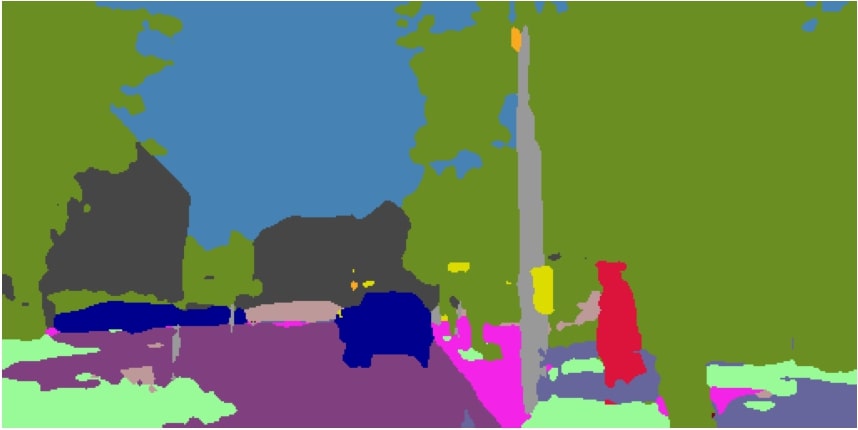}   };
            \begin{scope}[
            x={($0.1*(image.south east)$)},
            y={($0.1*(image.north west)$)}]
            \draw[thick,green] (6.5,1.0) rectangle (8,4.5) ;
        \end{scope}
    \end{tikzpicture}
	 \tabularnewline
	 
	\includegraphics[width=0.241\textwidth,height=0.1205\textwidth]{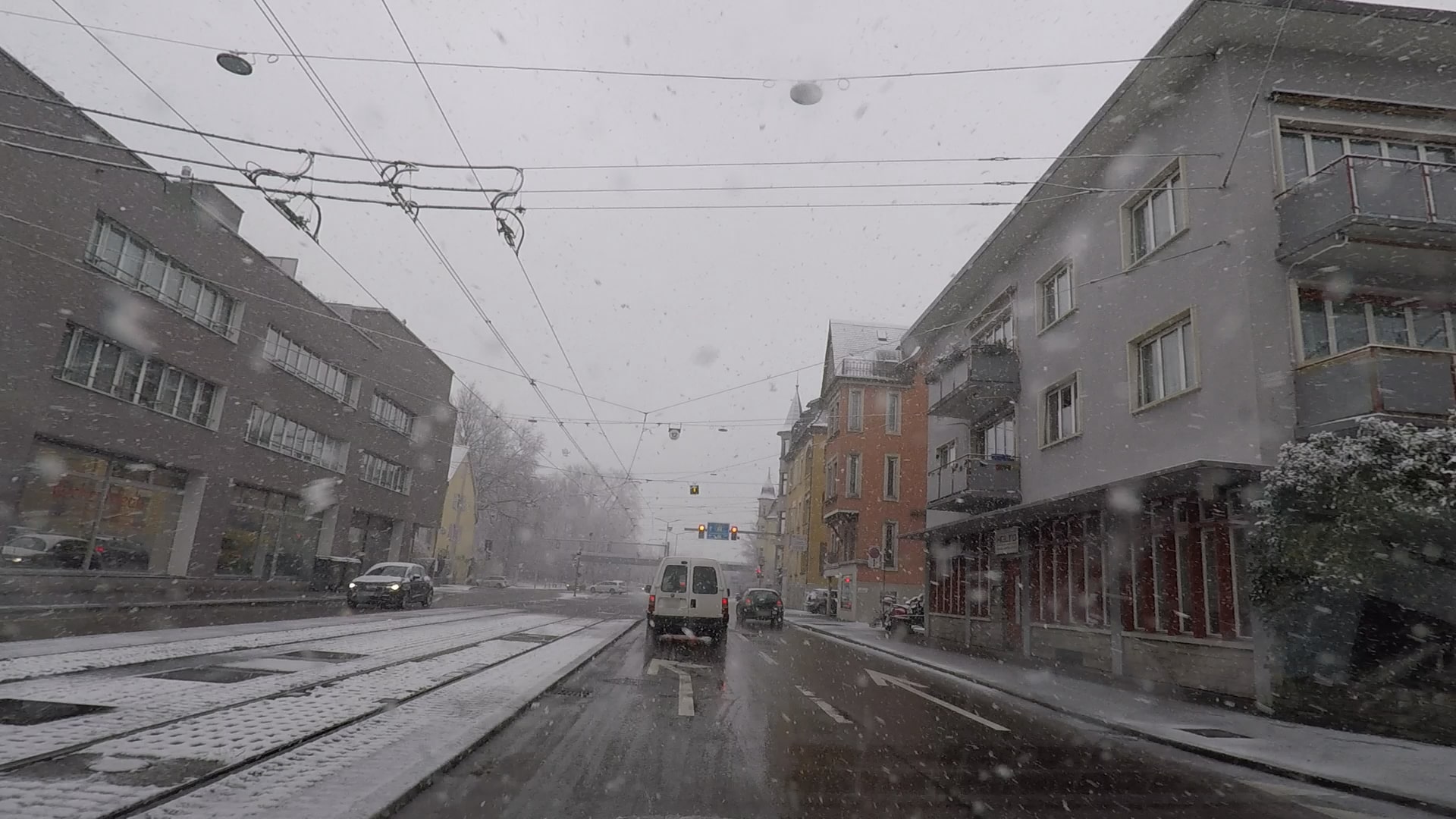} 
	        & {\footnotesize{}}
	   \begin{tikzpicture}
            \node [
	        above right,
	        inner sep=0] (image) at (0,0) {\includegraphics[width=0.241\textwidth,height=0.1205\textwidth]{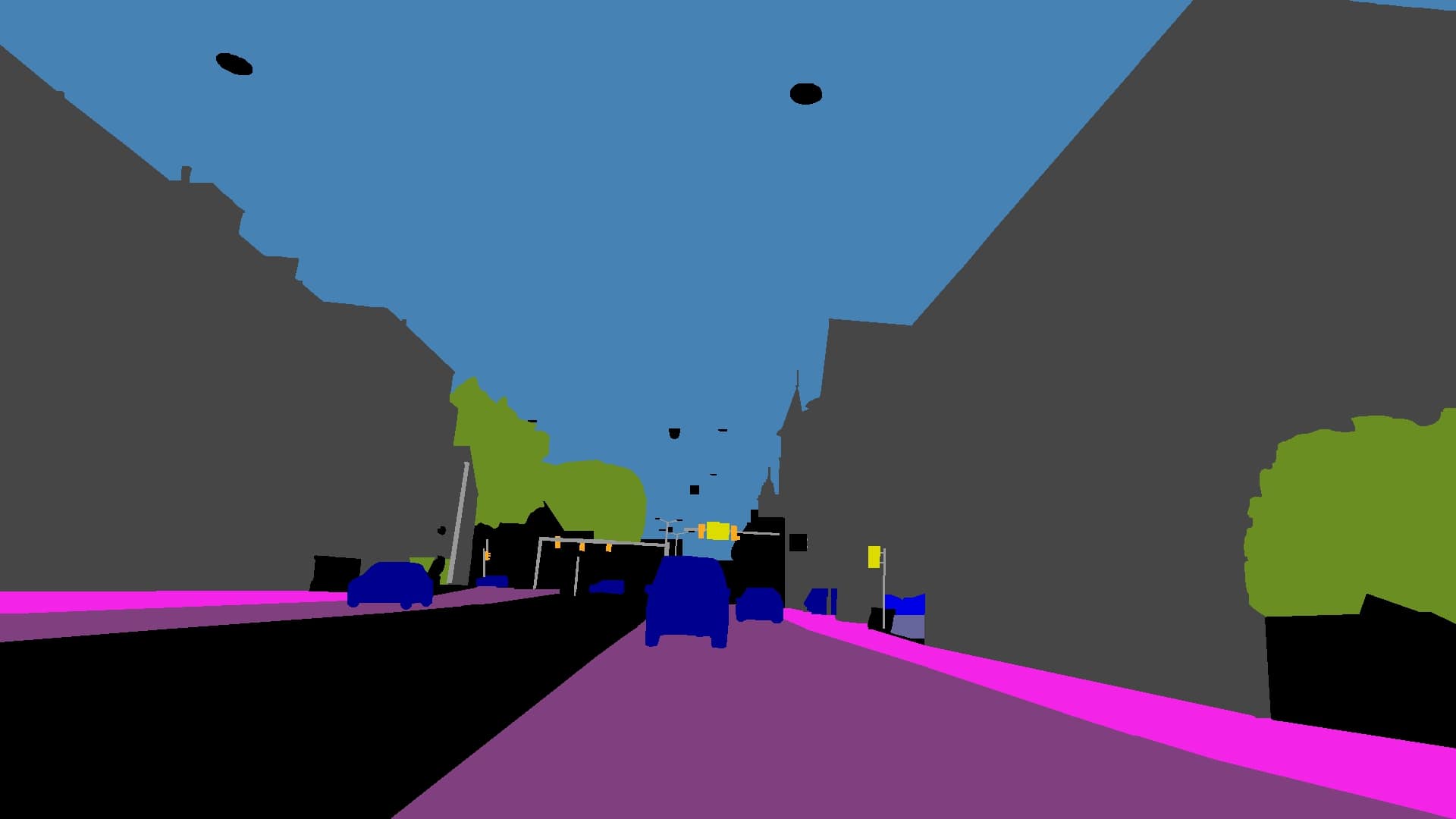}};
            \begin{scope}[
            x={($0.1*(image.south east)$)},
            y={($0.1*(image.north west)$)}]
           \draw[thick,green] (1.1,1.8) rectangle (4.5, 9.2) ;
        \end{scope}
    \end{tikzpicture}
	   & {\footnotesize{}}
	  \begin{tikzpicture}
            \node [
	        above right,
	        inner sep=0] (image) at (0,0) {\includegraphics[width=0.241\textwidth,]{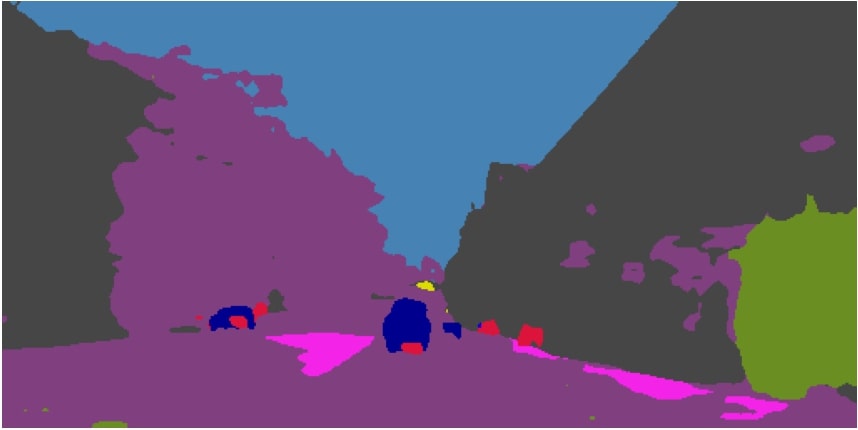}  };
            \begin{scope}[
            x={($0.1*(image.south east)$)},
            y={($0.1*(image.north west)$)}]
            \draw[thick,red] (1.1,1.8) rectangle (4.5, 9.2) ;
        \end{scope}
    \end{tikzpicture}
		    & {\footnotesize{}}
	\begin{tikzpicture}
            \node [
	        above right,
	        inner sep=0] (image) at (0,0) {\includegraphics[width=0.241\textwidth,]{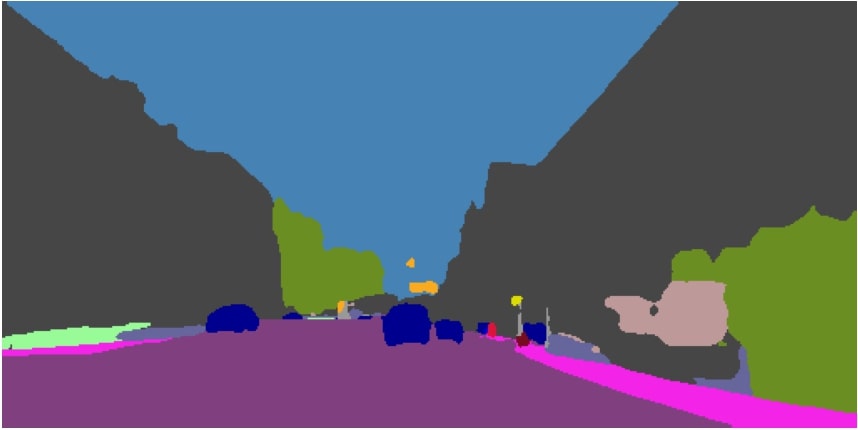} };
            \begin{scope}[
            x={($0.1*(image.south east)$)},
            y={($0.1*(image.north west)$)}]
           \draw[thick,green] (1.1,1.8) rectangle (4.5, 9.2) ;
        \end{scope}
    \end{tikzpicture}
	 \tabularnewline
	 
	 \includegraphics[width=0.241\textwidth,height=0.1205\textwidth]{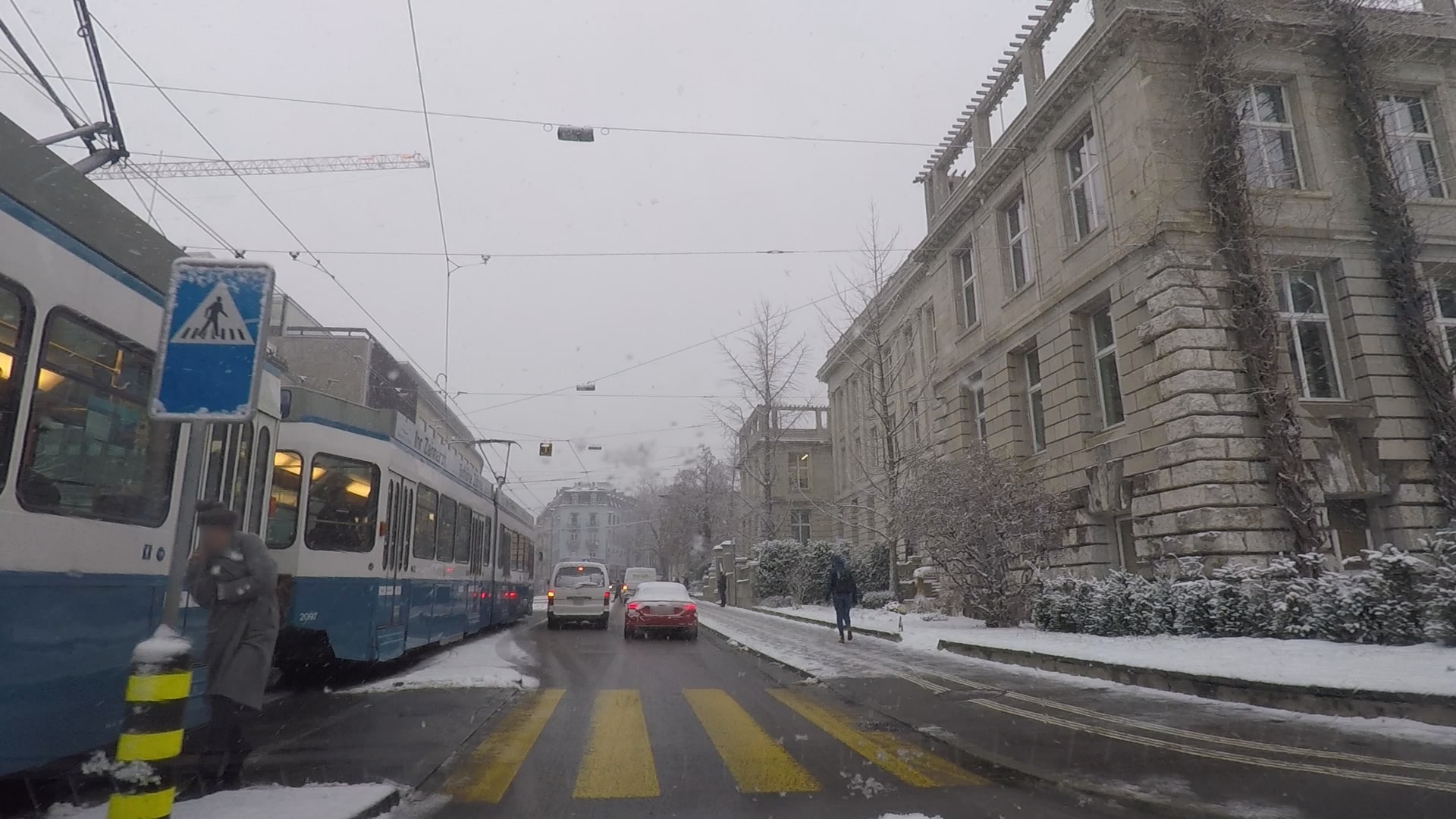} 
	        & {\footnotesize{}}
	        
	  \begin{tikzpicture}
            \node [
	        above right,
	        inner sep=0] (image) at (0,0) {\includegraphics[width=0.241\textwidth,height=0.1205\textwidth]{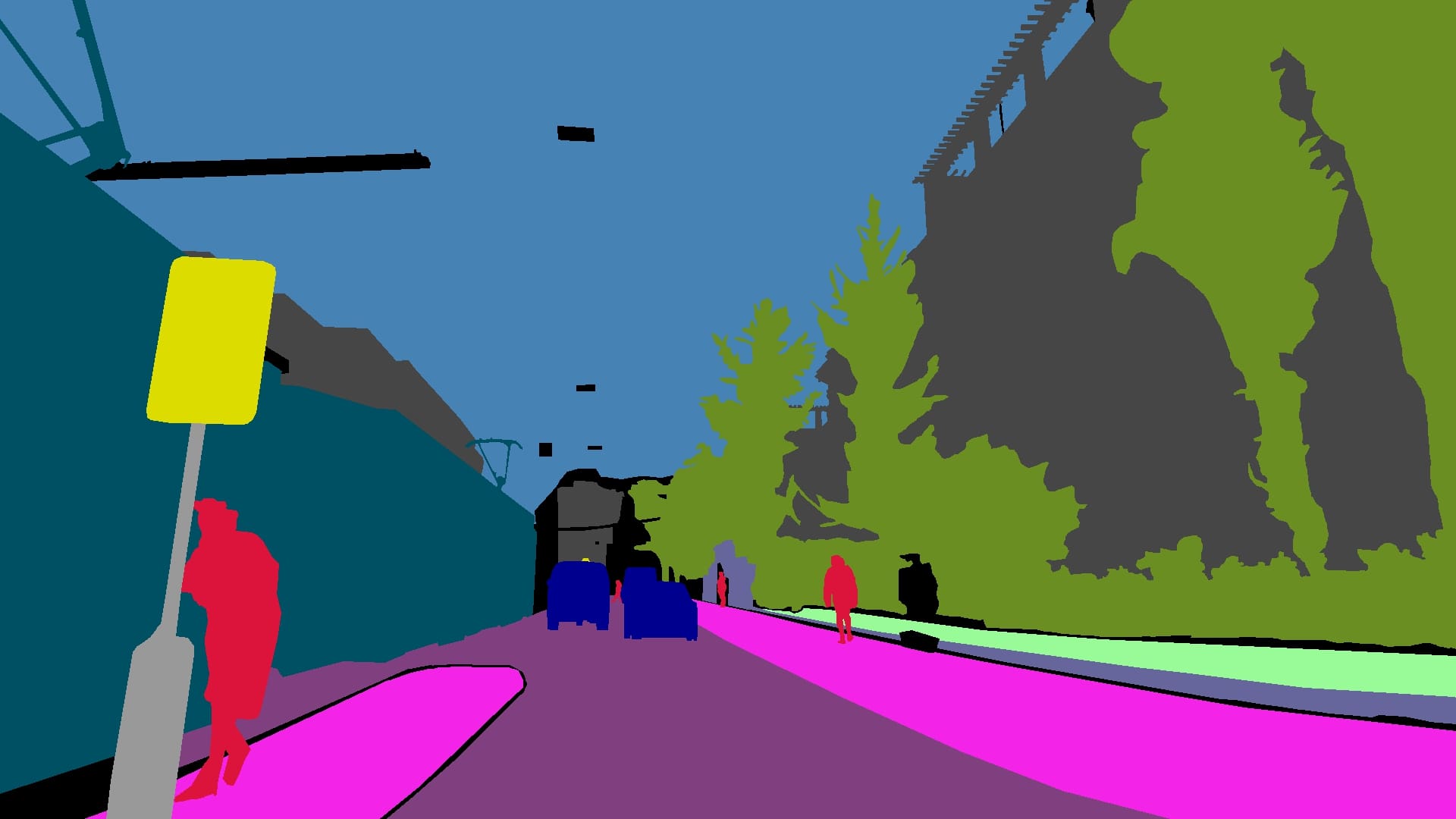}  };
            \begin{scope}[
            x={($0.1*(image.south east)$)},
            y={($0.1*(image.north west)$)}]
            \draw[thick,green] (0.5,0.3) rectangle (2.5, 7.2) ;
        \end{scope}
    \end{tikzpicture}      
	   & {\footnotesize{}}
	  \begin{tikzpicture}
            \node [
	        above right,
	        inner sep=0] (image) at (0,0) {\includegraphics[width=0.241\textwidth,]{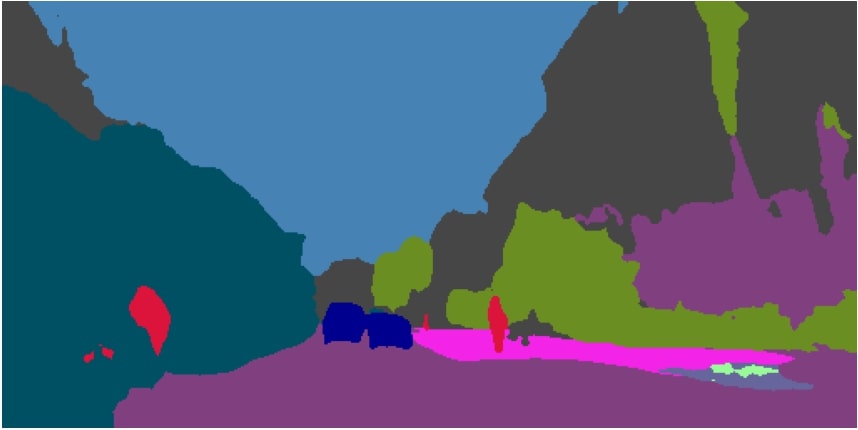}};
            \begin{scope}[
            x={($0.1*(image.south east)$)},
            y={($0.1*(image.north west)$)}]
           \draw[thick,red] (0.5,0.3) rectangle (2.5, 7.2) ;
        \end{scope}
    \end{tikzpicture}
		    & {\footnotesize{}}
    \begin{tikzpicture}
            \node [
	        above right,
	        inner sep=0] (image) at (0,0) {	\includegraphics[width=0.241\textwidth,]{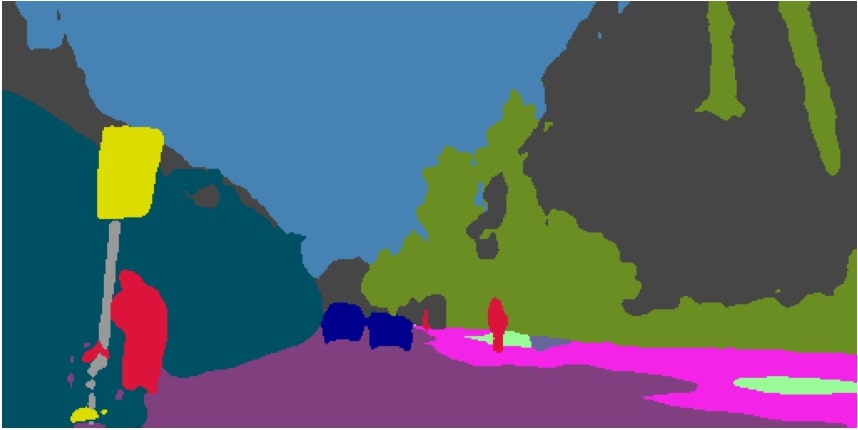}};
            \begin{scope}[
            x={($0.1*(image.south east)$)},
            y={($0.1*(image.north west)$)}]
           \draw[thick,green] (0.5,0.4) rectangle (2.5, 7.3) ;
        \end{scope}
    \end{tikzpicture}
	 \tabularnewline
	 
	\end{tabular}
\hfill{}
\par\end{centering}
\caption{Semantic segmentation results of Cityscapes to ACDC generalization using HRNet. The HRNet is trained on Cityscapes only. The segmenter trained with {\ourstyle} provides more reasonable prediction under adverse weather conditions. 
} 
\label{fig:dg-semseg}
\end{figure*}
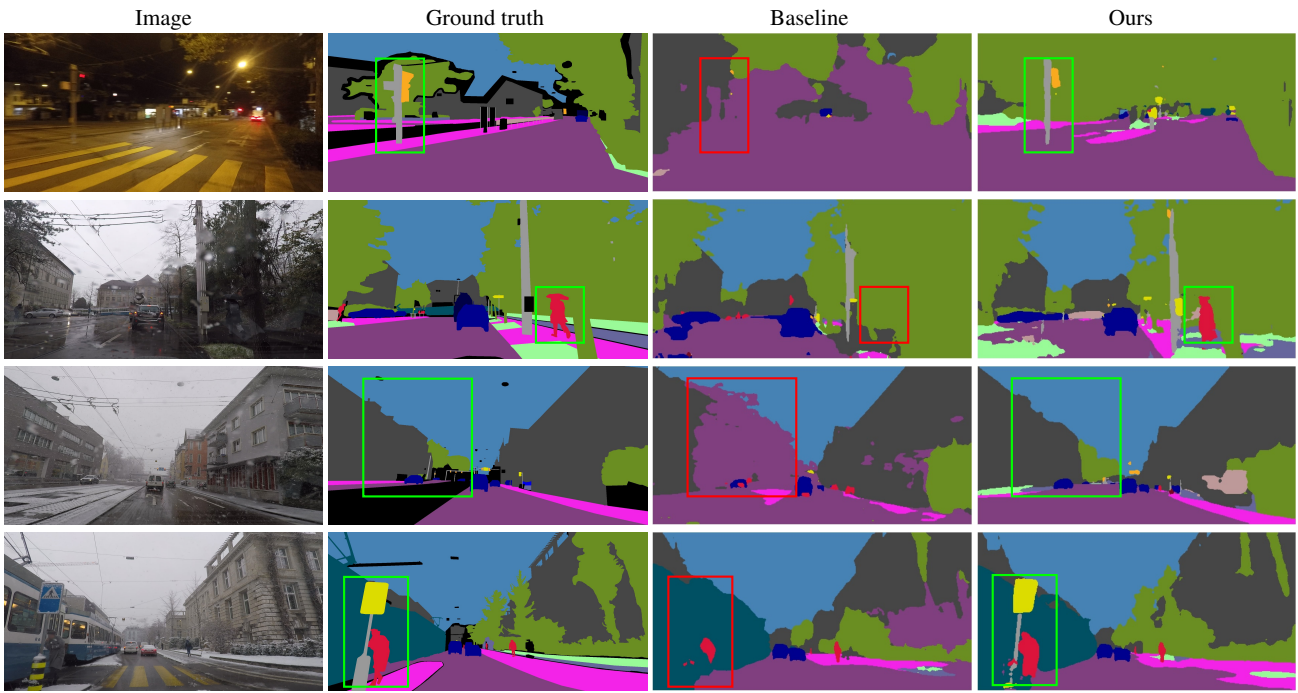

%% file: figs/compare_StyleMix.tex
\begin{figure}[t]
    \begin{centering}
    \setlength{\tabcolsep}{0.0em}
    \renewcommand{\arraystretch}{0}
    \par\end{centering}
    \begin{centering}
    \hfill{}
	\begin{tabular}{@{}c@{}c}
        \centering
		\small Content & \small Style  \tabularnewline
		\includegraphics[width=0.47\linewidth]{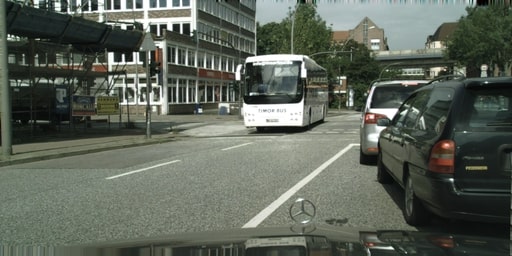} &
		{\footnotesize{}} \includegraphics[width=0.47\linewidth]{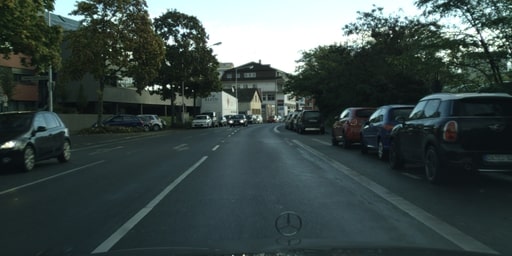}  \tabularnewline
		
		\small StyleMix\newcite{hong2021stylemix} & \small ISSA (Ours) \tabularnewline
		\includegraphics[width=0.47\linewidth]{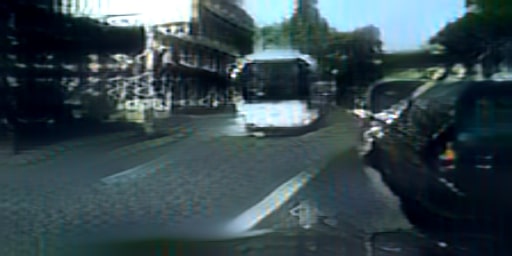} 
		&{\footnotesize{}} \includegraphics[width=0.47\linewidth]{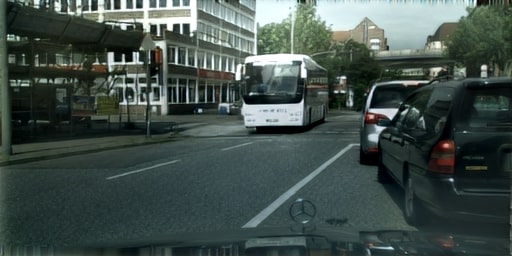}
		\tabularnewline
		\end{tabular}
\hfill{}
\par\end{centering}
\caption{Comparison of StyleMix\newcite{hong2021stylemix} and ISSA. StyleMix has rather low fidelity, while {\ourstyle} can preserve more details. }
\label{fig:stylemix-sample-1}
\end{figure}

%% file: tables/hrnet_bdd100k_daytime.tex
\begin{table}[t]
\begin{center}
{\small
    \begin{tabular}{@{}l|c|cccc@{}}
        Method & BDD100K & ACDC-Night & DarkZ\"urich   \\ \midrule
        Baseline
        & 52.97  & 23.52 & 23.63  \\
        \midrule
        CutMix & \textbf{54.03}  & 24.37 & 23.99  \\
        Weather  & 52.10  & 23.79 &  24.21 \\
        Digital  & 52.10   & 24.17 & 23.24 \\
        StyleMix 
        & 46.33 & 19.13 & 19.27 \\ 
        \ourstylebf \textbf{(Ours)} & 53.37 & \textbf{25.93} & \textbf{26.55} \\
    \end{tabular}
}
\end{center}
\caption{Comparison of data augmentation techniques for improving domain generalization using HRNet\hrnet, i.e., from BDD100K-Daytime to ACDC-Night and Dark Z\"urich. BDD100K-Daytime is a subset of BDD100K, which contains $2526$ images in daytime under various weather conditions, but not in dawn/nighttime. Here, we evaluate the domain generalization with respect to day to night.
}
\label{tab:hrnet-bdd-daytime-da}
\end{table}

%% file: tables/dg_bn_aug.tex
\begin{table}[t]
\begin{center}
{\small
    \begin{tabular}{@{}l|c|cccc@{}}
        Method  & CS & ACDC & BDD & DarkZ   \\ \midrule
        Baseline\newcite{chen2017deeplabv2} & 61.73  & 30.86 & 34.30  &  11.62 \\ 
    \midrule
        MixStyle\newcite{zhou2021mixstyle} & 59.01 & 36.97 & 36.27  & 9.38  \\
        DSU \newcite{li2022dsu}   & 59.59 & 38.31 & 35.53  & 12.29   \\
        \textbf{{\ourstylebf} (Ours)} & \textbf{62.20} & \textbf{43.21} & \textbf{42.60} & \textbf{21.56} \\ 
    \midrule
        MixStyle + \ourstyle  & 60.17 & 41.81  &42.17  &  20.56\\ 
        DSU + \ourstyle  & 60.20 & 43.31 & 42.24 & 24.63  \\ 
    \end{tabular}
}
\end{center}
\caption{Comparison with feature-level augmentation methods on domain generalization performance of Cityscapes 
as the source.
Following DSU\newcite{li2022dsu}, we conduct experiments using DeepLab v2\newcite{chen2017deeplabv2} as the baseline for fair comparison.
}
\label{tab:feature-aug-da}
\end{table}

%% file: tables/robustnet.tex
\begin{table}[t]
\begin{center}
{\small
    \begin{tabular}{@{}l|c|cccc@{}}
         Method & CS & ACDC & BDD & DarkZ   \\ \midrule
        Baseline\newcite{chen2018deeplabv3plus}  & 69.01  & 44.23  & 43.27 & 16.03  \\
        \midrule
        RobustNet\newcite{robustnet_2021}   & \textbf{69.47}  & 47.25 & 46.94 & 20.11  \\
        + \ourstyle  & 69.45 &  \textbf{47.55} & \textbf{48.44} & \textbf{23.09}  \\ 
        \midrule
        SHADE\newcite{shade-style-hall}  &  \textbf{64.24} & 47.30  & 46.44 & 25.37 \\ 
        + \ourstyle  & 63.79 & \textbf{47.64} & \textbf{47.76} & \textbf{25.58}
        %
        
    \end{tabular}
}
\end{center}
\caption{Combination of {\ourstyle} and RobustNet\newcite{robustnet_2021}. We adopt the experimental setting of RobustNet and use DeepLab v3+\newcite{chen2018deeplabv3plus} as the baseline. Our {\ourstyle} is complementary to RobustNet and further improves its generalization performance. 
}
\label{tab:robustnet-cs-da}
\end{table}

%% file: tables/uda_comparison.tex
\begin{table}[t]
\begin{center}
{\footnotesize  
    {
    \begin{tabular}{l|c|ccc}
        Method & Network & Use Target & mIoU \\
    \midrule
        Baseline & \multirow{10}{*}{DeepLabv2
        } & \textbf{---} & 30.9 
        \\ \midrule
        BDL\scriptsize{\newcite{li2019bdl}} & & \cmark & 32.7 \\
        CRST \scriptsize{\newcite{zou2019confidence}}  &  & \cmark & 32.8 \\
        AdaptSegNet\scriptsize{\newcite{tsai2018AdaptSegNet}}  &  & \cmark & 33.4 \\
        SIM\scriptsize{\newcite{wang2020differential}}  &  & \cmark & 34.6 \\
        MRNet\scriptsize{\newcite{zheng2021rectifying}}  &   & \cmark & 36.1 \\
        ADVENT\scriptsize{\newcite{tsai2019advent}}  &  & \cmark & 37.7 \\
        CLAN\scriptsize{\newcite{luo2019clan}}  &   & \cmark & 39.0 \\
        FDA\scriptsize{\newcite{yang2020fda}}  &   & \cmark & 45.7 \\
        \ourstyle(Ours) &   & \xmark & 43.2 \\
        \midrule
        DAFormer\scriptsize{\newcite{hoyer2022daformer}} & DAFormer
        & \cmark & 55.4\\
        \ourstyle(Ours) & SegFormer
        & \xmark & 52.5\\
    \end{tabular}
    }
}
\end{center}
\caption{Comparison with UDA methods on Cityscapes to ACDC generalization. Remarkably, our domain generalization method (without access to the target domain, neither images nor labels), is on-par or better than unsupervised domain adaptation (UDA) methods, which requires knowledge of the target domain during training. Results of UDA methods are from\acdc.}
\label{tab:uda-comparison}
\end{table}

%% file: figs/landscape_sample.tex
\begin{figure*}[t]
    \begin{centering}
    \setlength{\tabcolsep}{0.0em}
    \renewcommand{\arraystretch}{0}
    \par\end{centering}
    \begin{centering}
    \hfill{}%
	\begin{tabular}{@{}c@{}c@{}c@{}c@{}c}
        \centering
	Style & Content 1 & Mixed 1 & Content 2 & Mixed 2
	\tabularnewline
	\includegraphics[width=0.19\linewidth]{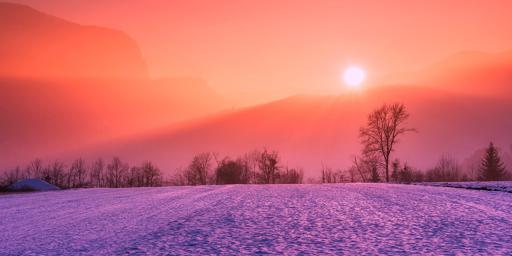} & {\footnotesize{}}
	\includegraphics[width=0.19\linewidth]{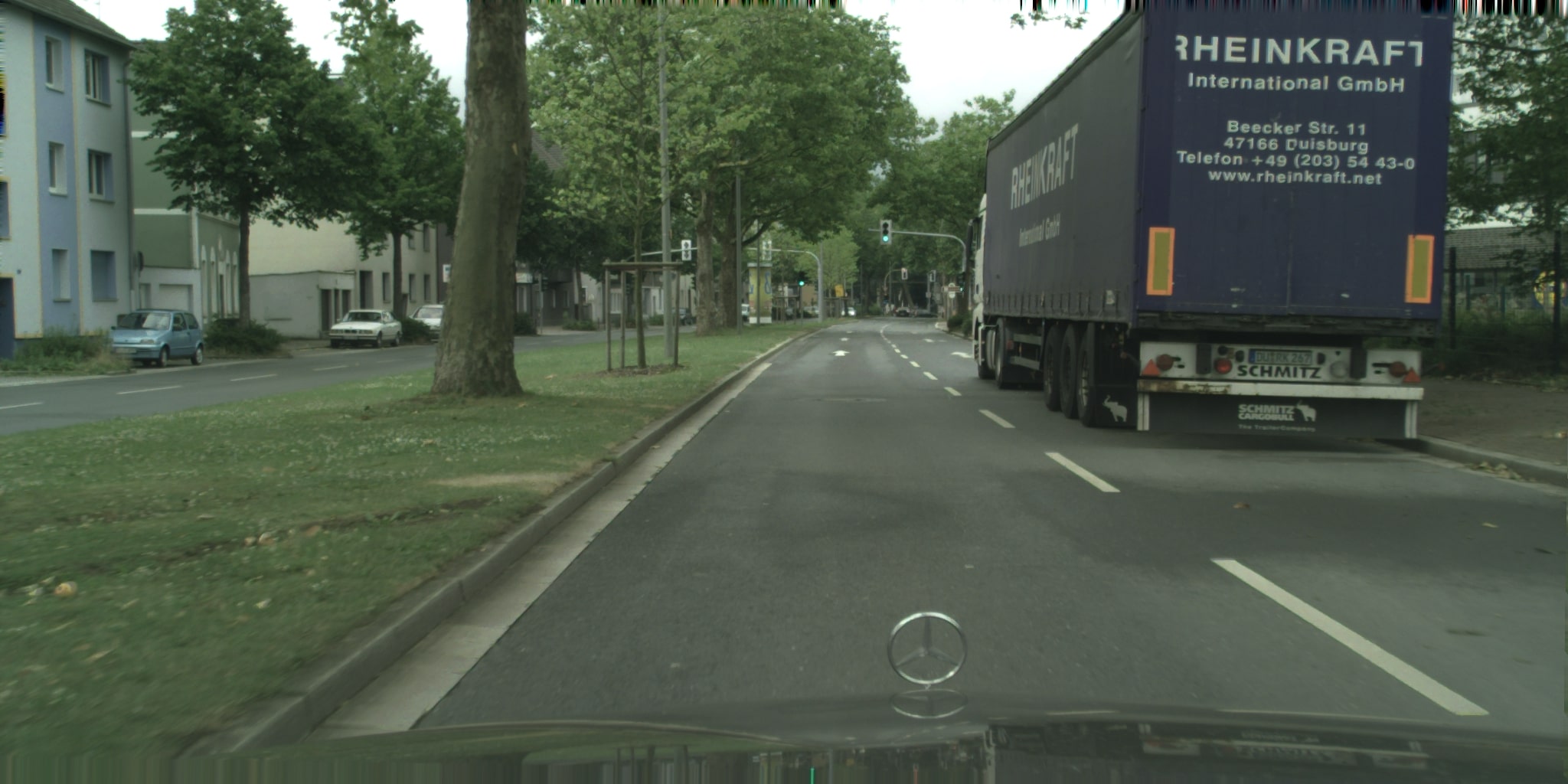} &
    {\footnotesize{}} 
	\includegraphics[width=0.19\linewidth]{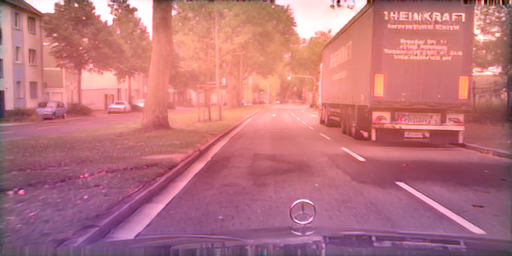} & {\footnotesize{}} 
	\includegraphics[width=0.19\linewidth]{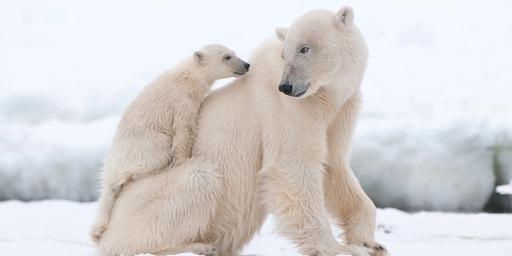} &
	{\footnotesize{}} 
	\includegraphics[width=0.19\linewidth]{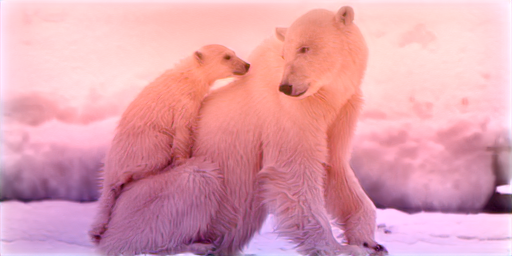}
	\tabularnewline
	
    \includegraphics[width=0.19\linewidth]{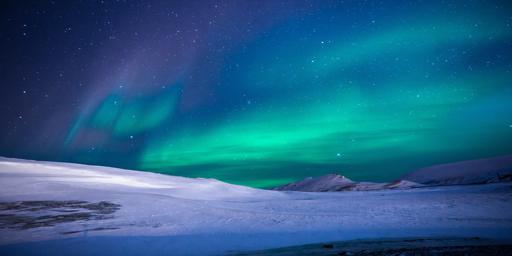} & {\footnotesize{}}
	\includegraphics[width=0.19\linewidth]{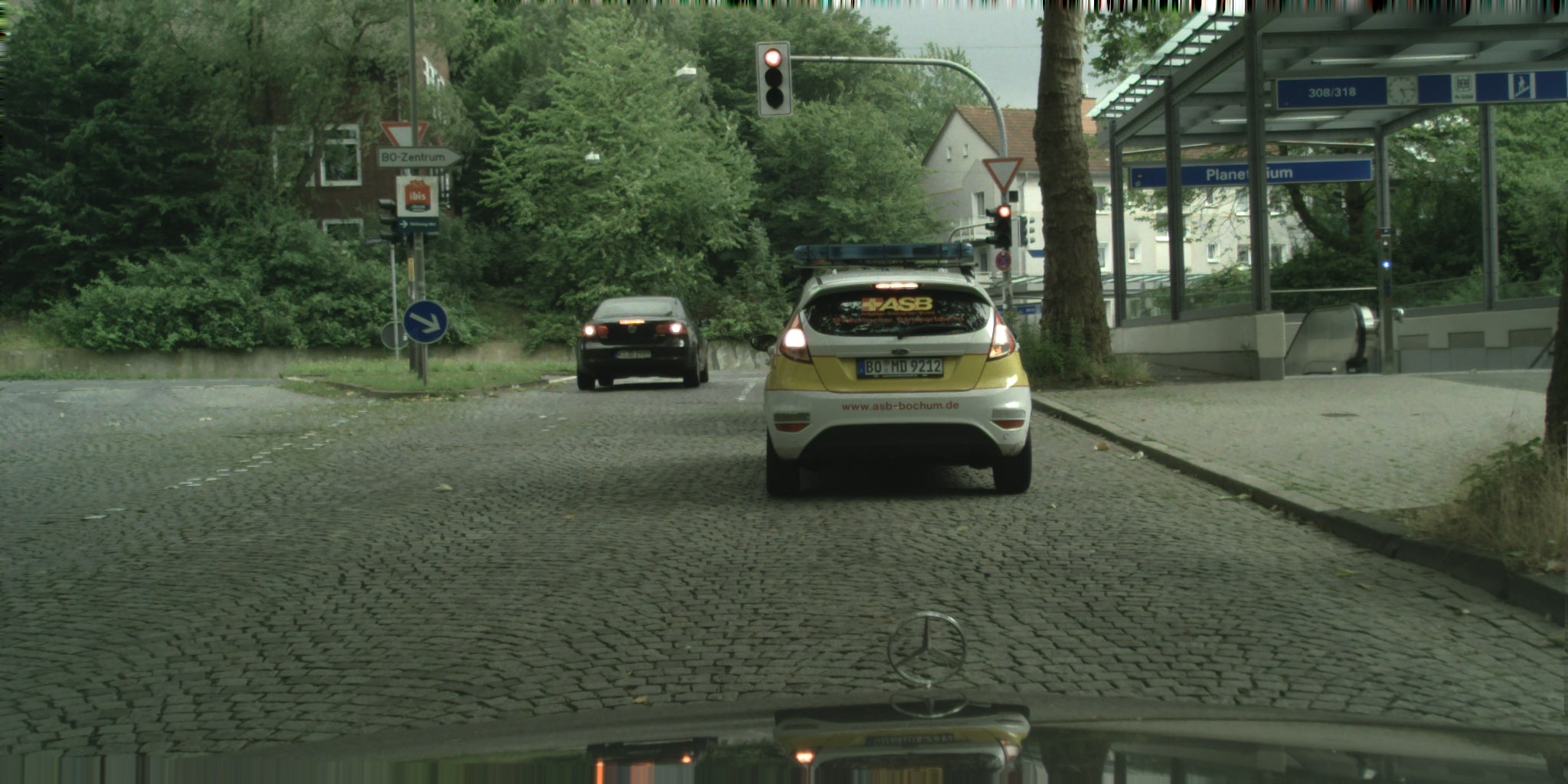} &
	{\footnotesize{}}
	\includegraphics[width=0.19\linewidth]{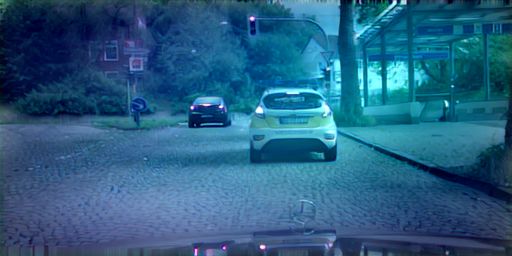} &
	{\footnotesize{}}
	\includegraphics[width=0.19\linewidth]{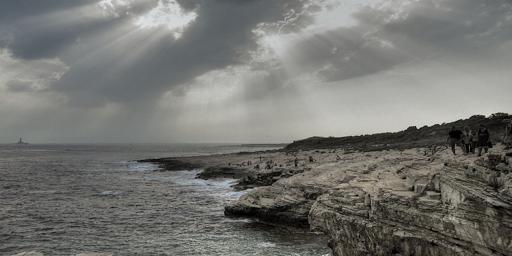} &
	{\footnotesize{}}
	\includegraphics[width=0.19\linewidth]{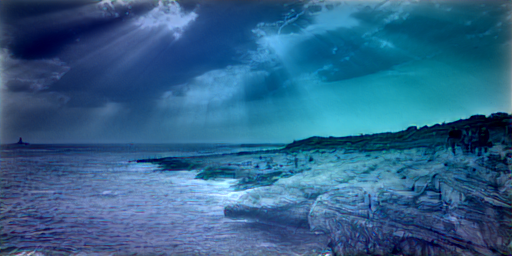} 
	\tabularnewline
	
    \includegraphics[width=0.19\linewidth]{gray_512x256.jpg} & {\footnotesize{}}
	\includegraphics[width=0.19\linewidth]{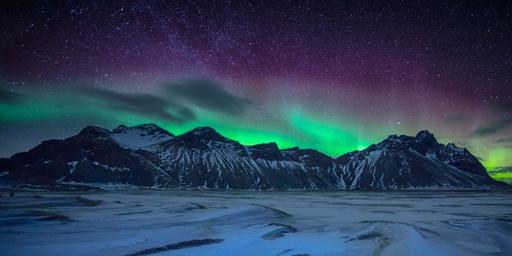} & {\footnotesize{}}
	\includegraphics[width=0.19\linewidth]{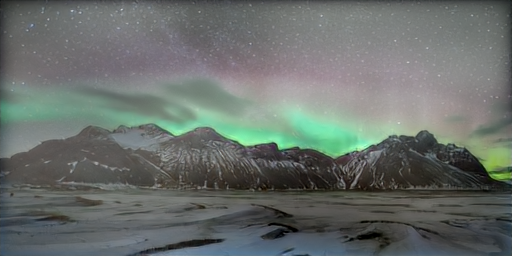} & {\footnotesize{}}
	\includegraphics[width=0.19\linewidth]{green_512x256.jpg} & {\footnotesize{}}
	\includegraphics[width=0.19\linewidth]{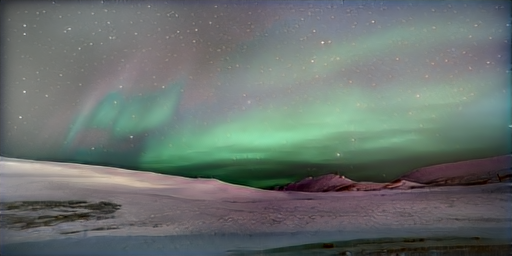} 
	\tabularnewline		
	
	\includegraphics[width=0.19\linewidth]{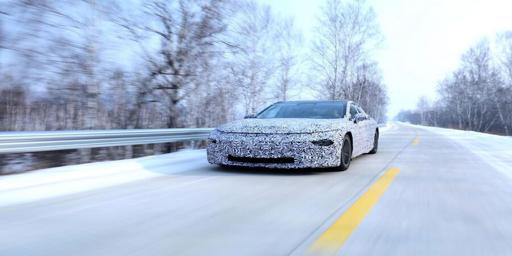} & {\footnotesize{}} 	
	\includegraphics[width=0.19\linewidth]{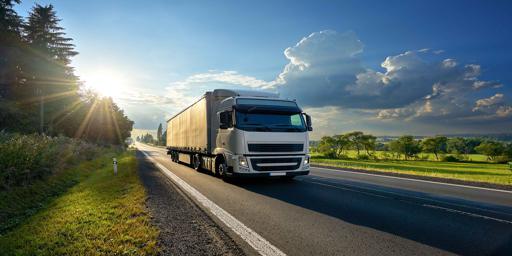} & {\footnotesize{}}
	\includegraphics[width=0.19\linewidth]{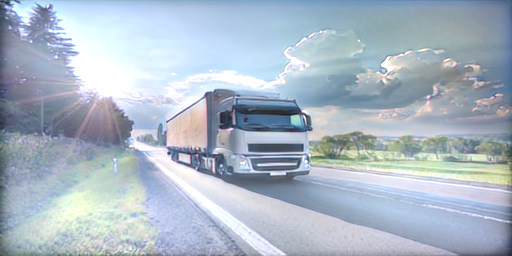} & {\footnotesize{}} 	
	\includegraphics[width=0.19\linewidth]{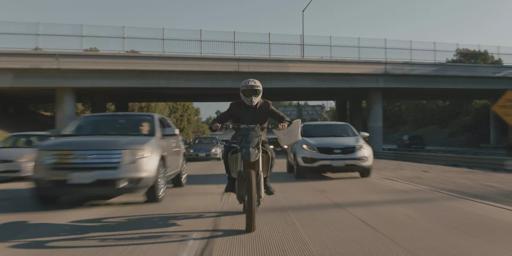} & {\footnotesize{}}
	\includegraphics[width=0.19\linewidth]{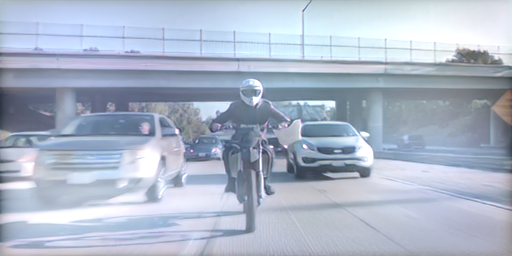}
	\tabularnewline
\end{tabular}
\hfill{}
\par\end{centering}
\caption{
Extra-source exemplar based style synthesis using web-crawled images, where
the generator and encoder are only trained on Cityscapes. Except for the Content 1 image of the first 2 rows, all the others are web-crawled images.
}
\label{fig:landscape-example}
\end{figure*}

%% file: figs/interpolation.tex
\begin{figure*}[t]
    \begin{centering}
    \setlength{\tabcolsep}{0.0em}
    \renewcommand{\arraystretch}{0}
    \par\end{centering}
    \begin{centering}
    \hfill{}%
	\begin{tabular}{@{}c@{}c@{}c@{}c@{}c@{}c}
        \centering
	Content &   &   &  & & Style
	\tabularnewline
	\includegraphics[width=0.16\linewidth]{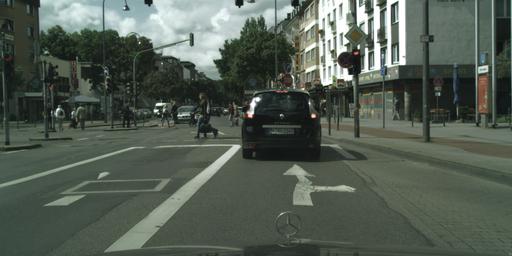} & {\footnotesize{}}
	\includegraphics[width=0.16\linewidth]{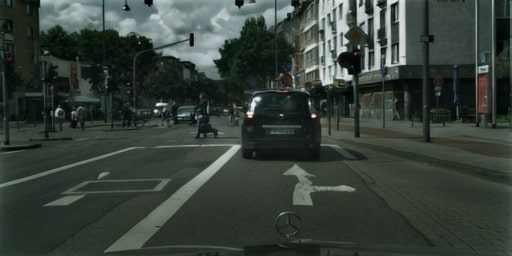} &
    {\footnotesize{}} 
	\includegraphics[width=0.16\linewidth]{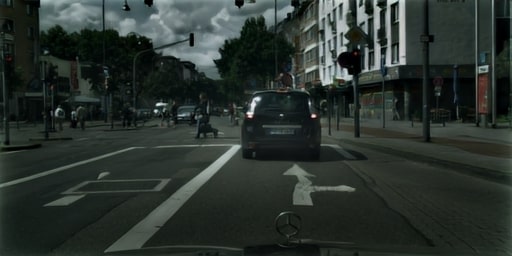} & {\footnotesize{}} 
	\includegraphics[width=0.16\linewidth]{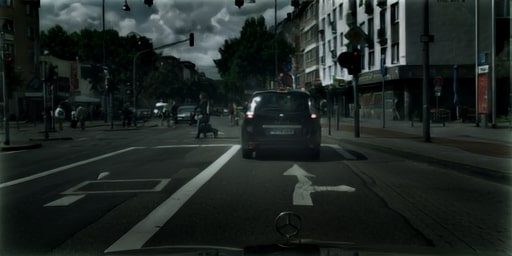} &
	{\footnotesize{}} 
	\includegraphics[width=0.16\linewidth]{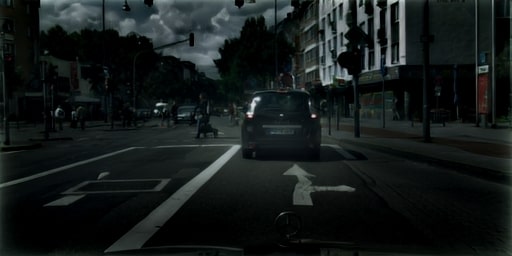} &
	{\footnotesize{}} 
	\includegraphics[width=0.16\linewidth]{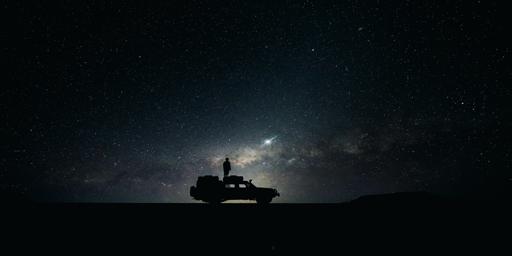}
	\tabularnewline
	
	\includegraphics[width=0.16\linewidth]{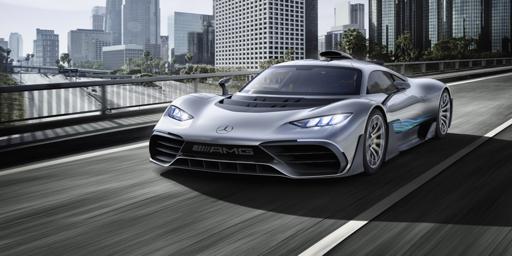} & {\footnotesize{}}
	\includegraphics[width=0.16\linewidth]{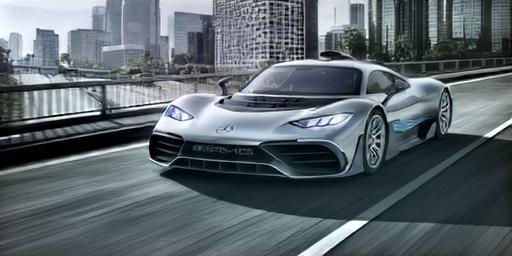} &
    {\footnotesize{}} 
	\includegraphics[width=0.16\linewidth]{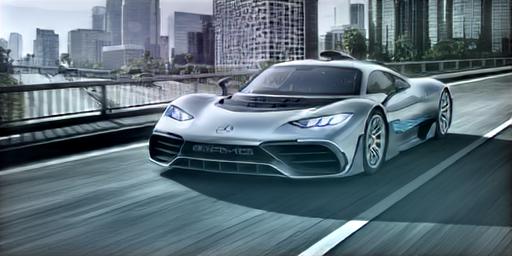} & {\footnotesize{}} 
	\includegraphics[width=0.16\linewidth]{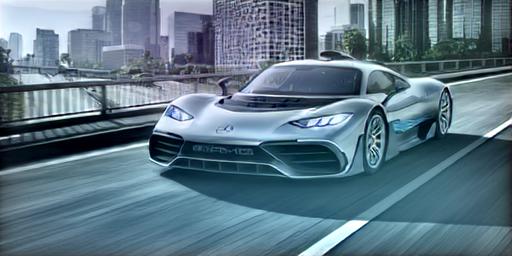} &
	{\footnotesize{}} 
	\includegraphics[width=0.16\linewidth]{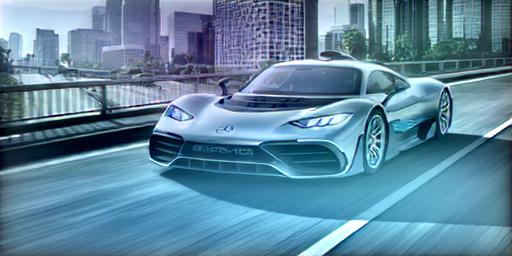} &
	{\footnotesize{}} 
	\includegraphics[width=0.16\linewidth]{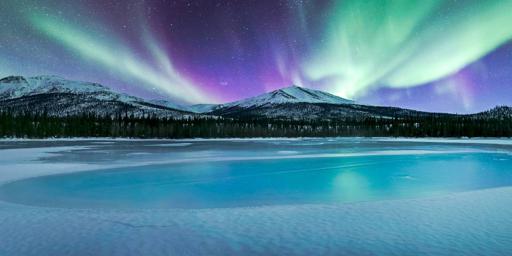}
	\tabularnewline
\end{tabular}
\hfill{}
\par\end{centering}
\caption{Visualization of interpolation in the style latent space. As illustrated, we can control the style mixing strength and achieve a smooth transition on both trained Cityscapes and unseen  web-crawled images. 
}
\label{fig:interpolation-new}
\end{figure*}

%% file: figs/acdc_stylized.tex
\begin{figure*}[t]
    \begin{centering}
    \setlength{\tabcolsep}{0.0em}
    \renewcommand{\arraystretch}{0}
    \par\end{centering}
    \begin{centering}
    \hfill{}
	\begin{tabular}{@{}l@{}c@{}c@{}c@{}c}
        \centering
		 & Snow & Night & Frog  & Rain \tabularnewline
        \hspace{0.075\textwidth} Content  \hspace{0.035\textwidth} \rotatebox{90}{\hspace{1.6em} Style}

		& {\footnotesize{}}
		\includegraphics[width=0.195\textwidth]{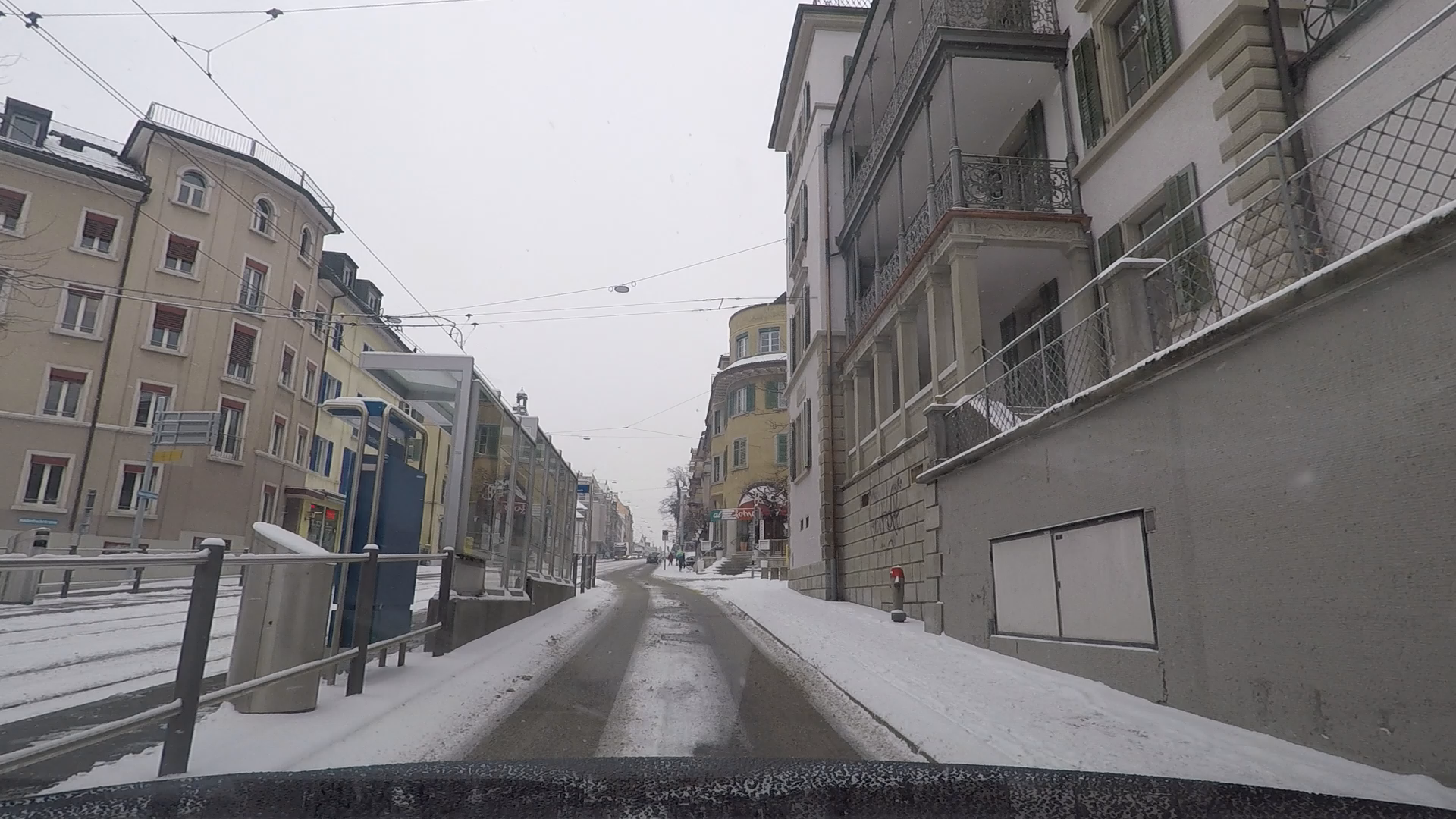} & {\footnotesize{}}
		\includegraphics[width=0.195\textwidth]{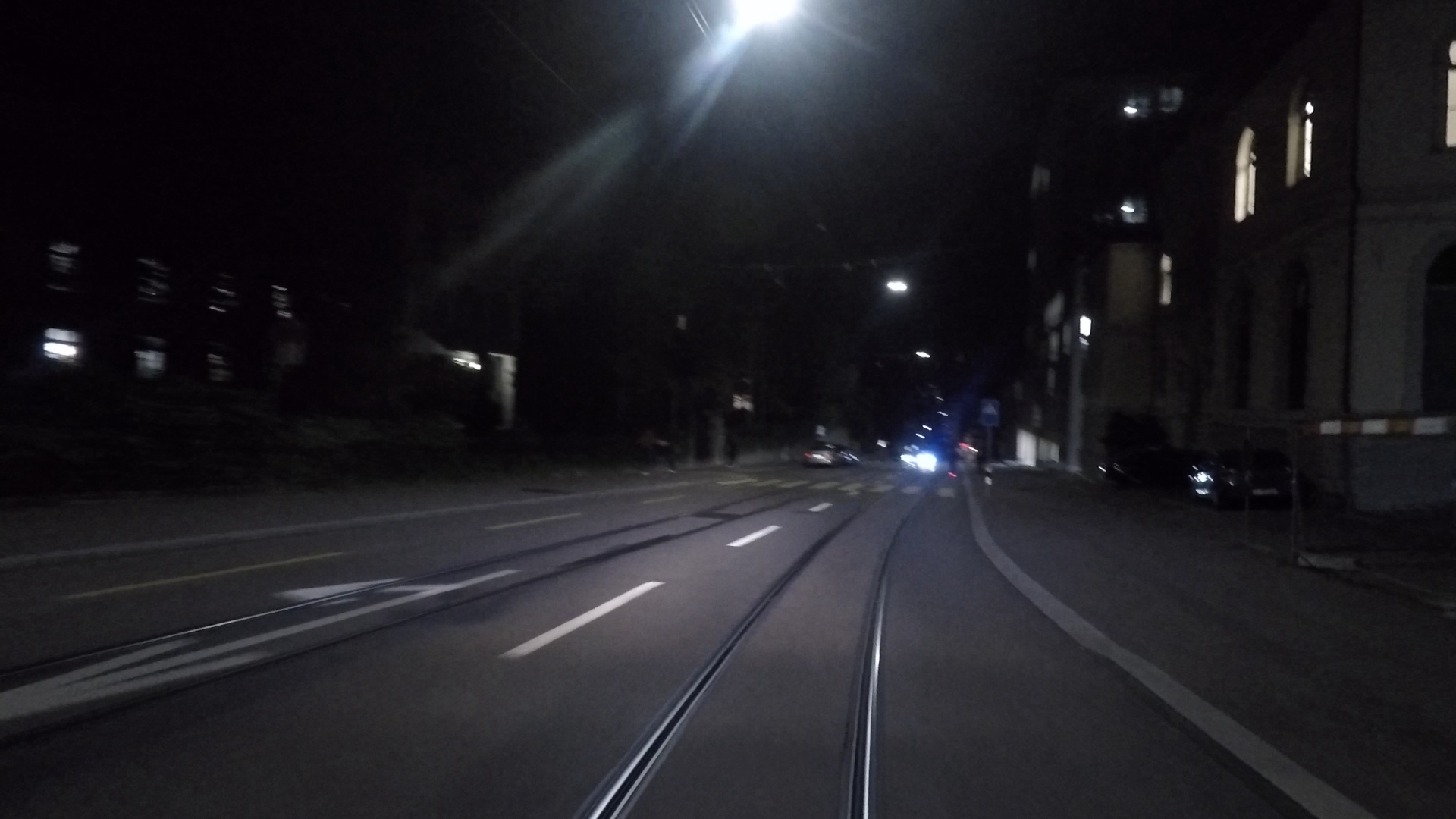} & {\footnotesize{}}
		\includegraphics[width=0.195\textwidth]{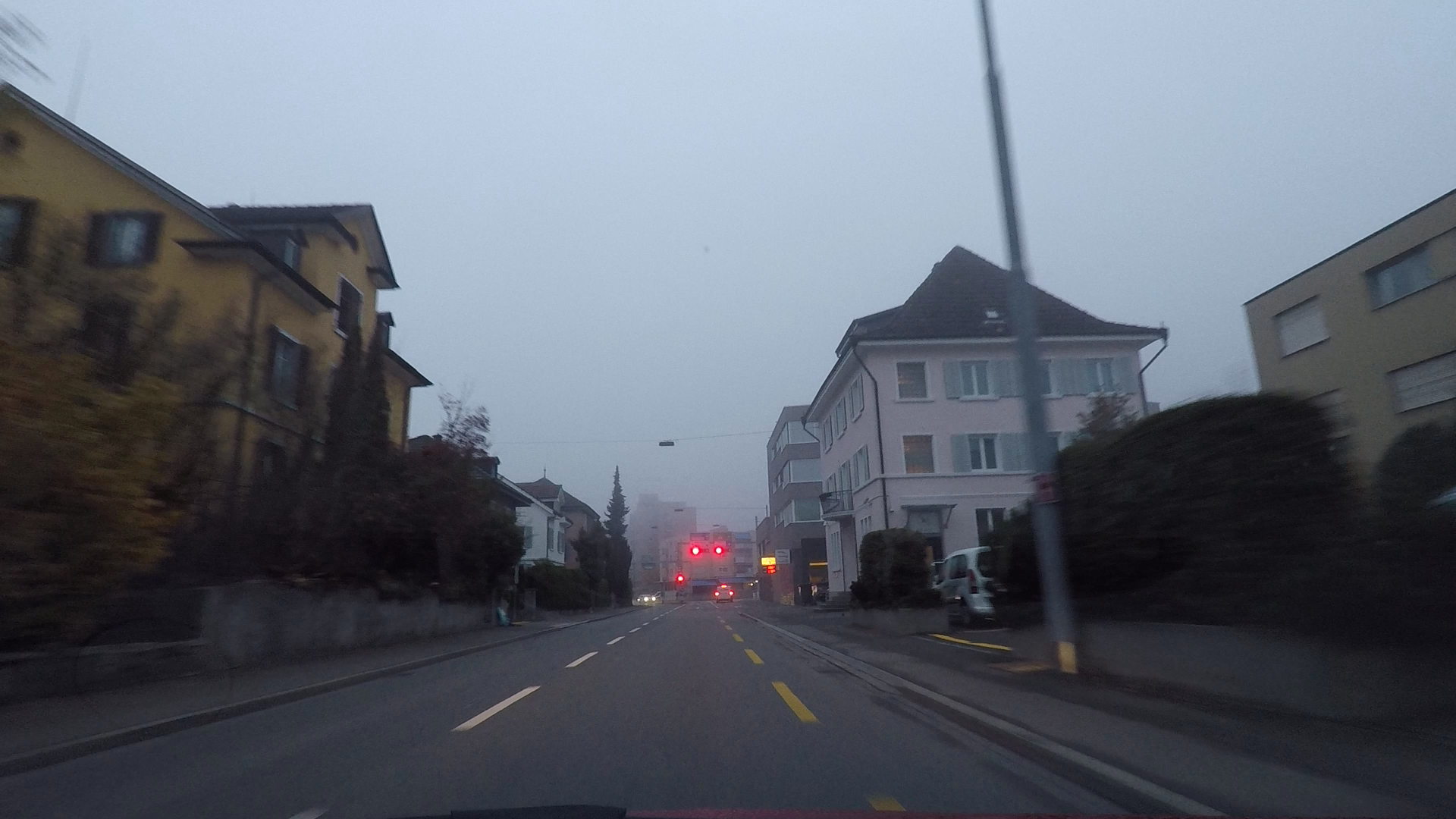} & {\footnotesize{}}
        \includegraphics[width=0.195\textwidth]{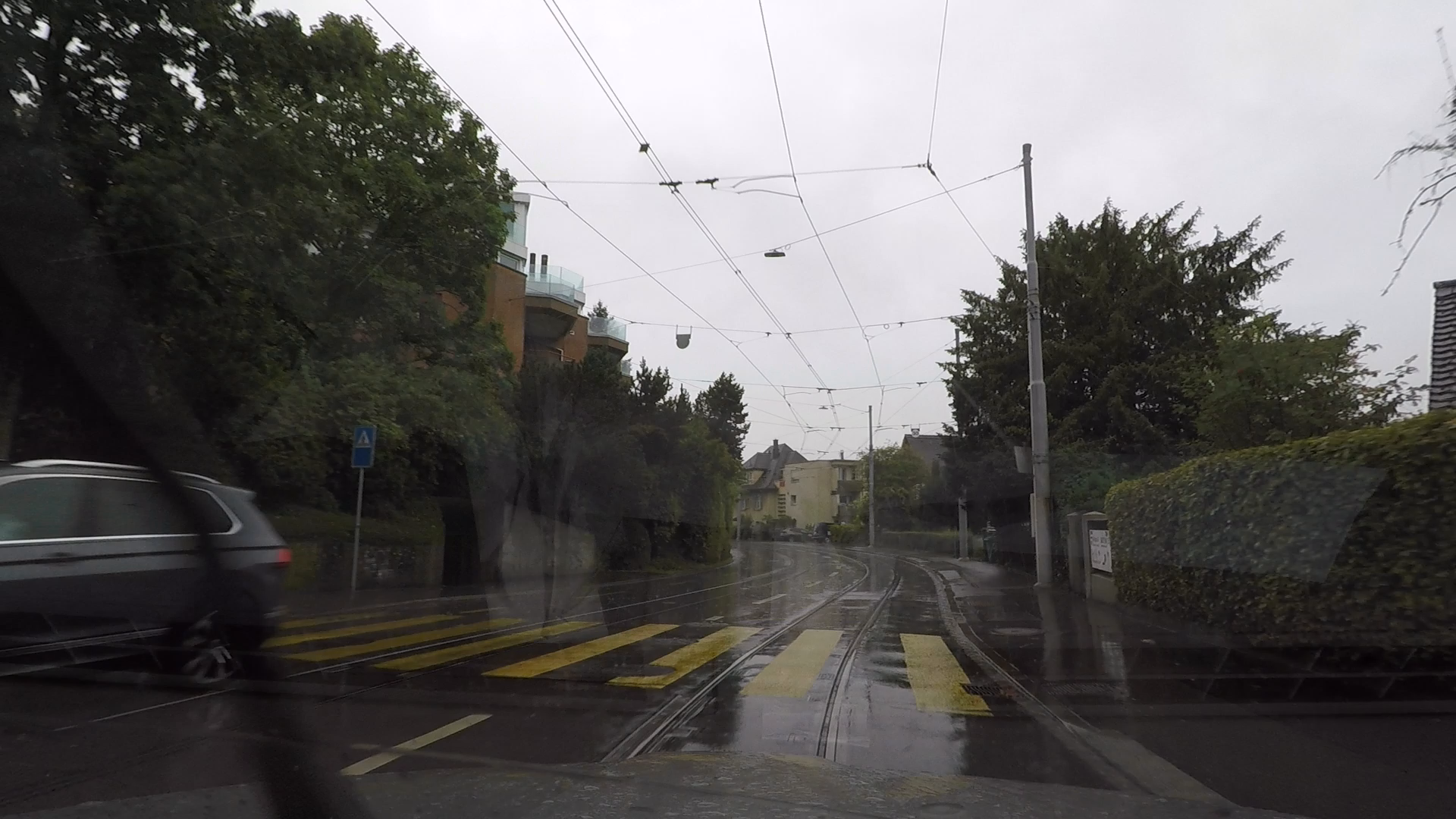} 
		\tabularnewline
		
		\includegraphics[width=0.195\textwidth]{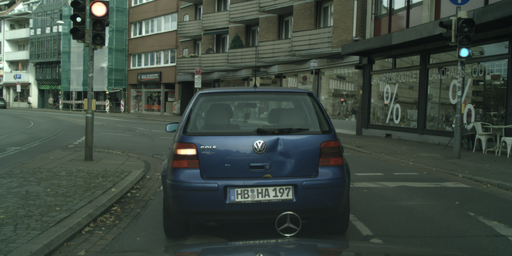} & {\footnotesize{}}
		\includegraphics[width=0.195\textwidth]{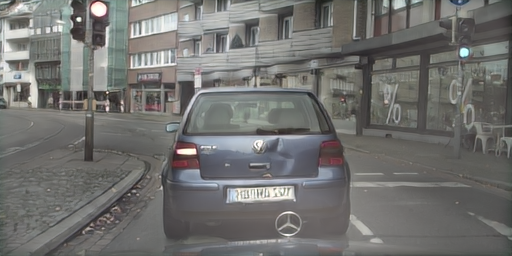} & {\footnotesize{}}
		\includegraphics[width=0.195\textwidth]{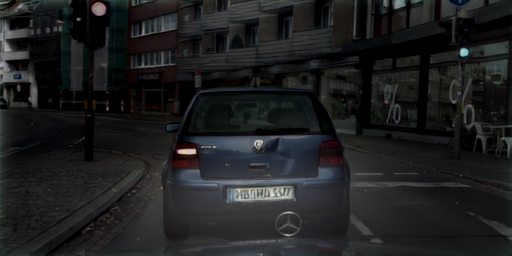} & {\footnotesize{}}
		\includegraphics[width=0.195\textwidth]{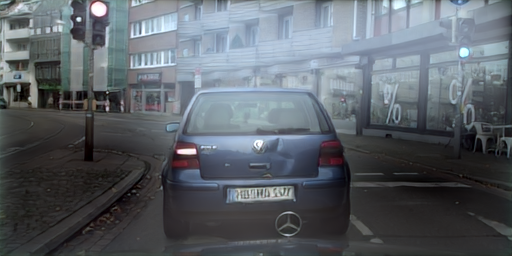} & {\footnotesize{}}
		\includegraphics[width=0.195\textwidth]{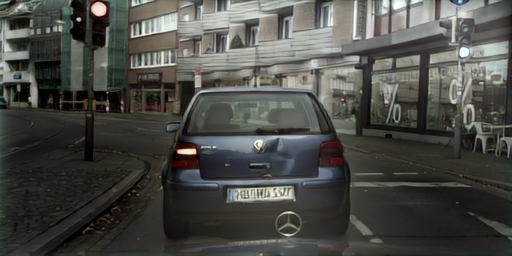} 
		\tabularnewline
		
		\includegraphics[width=0.195\textwidth]{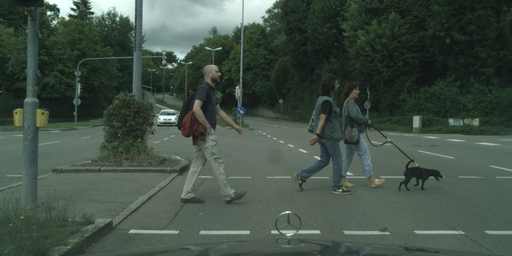} & {\footnotesize{}}
		\includegraphics[width=0.195\textwidth]{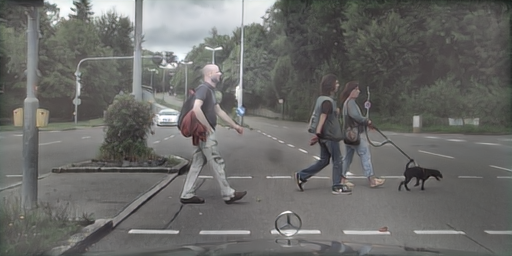} & {\footnotesize{}}
		\includegraphics[width=0.195\textwidth]{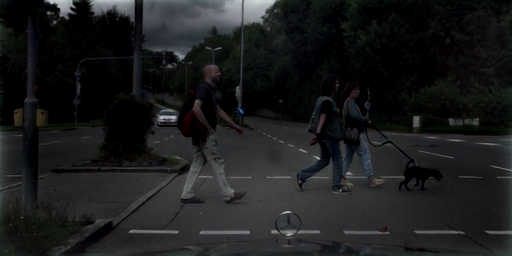} & {\footnotesize{}}
		\includegraphics[width=0.195\textwidth]{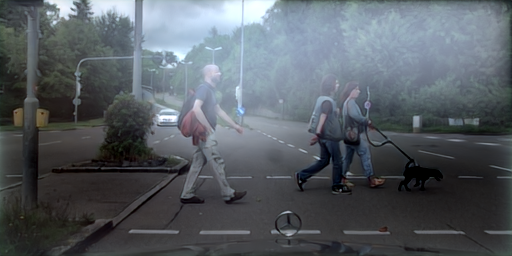} & {\footnotesize{}}
		\includegraphics[width=0.195\textwidth]{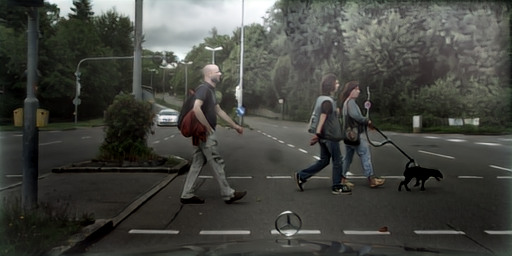}
		\tabularnewline
		
		\includegraphics[width=0.195\textwidth]{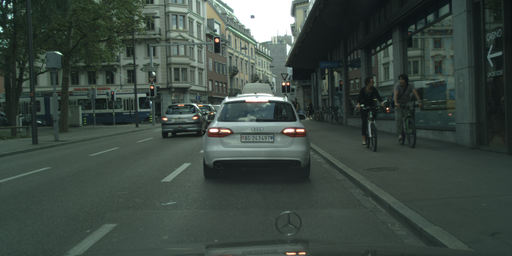} & {\footnotesize{}}
		\includegraphics[width=0.195\textwidth]{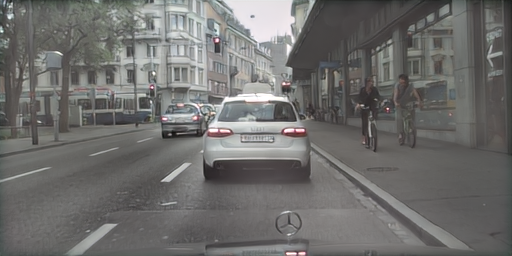} & {\footnotesize{}}
		\includegraphics[width=0.195\textwidth]{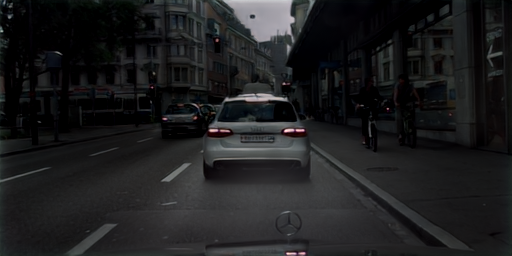} & {\footnotesize{}}
		\includegraphics[width=0.195\textwidth]{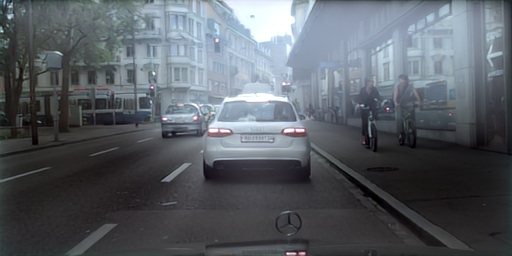} & {\footnotesize{}}
		\includegraphics[width=0.195\textwidth]{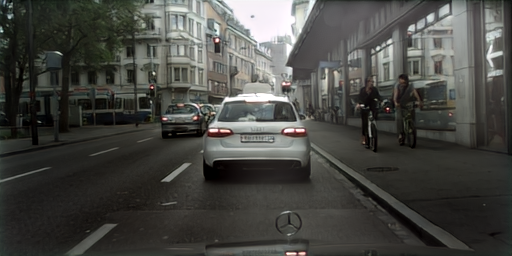}
		\tabularnewline
		\end{tabular}
\hfill{}
\par\end{centering}
\caption{
Visual examples of stylized data by transferring style from one unannotated ACDC sample (target domain) to Cityscapes (source domain). Best view in color. 
}
\label{fig:stylized_proxy}
\end{figure*}

%% file: tables/bddgan_dg.tex
\begin{table}[t]
\setlength{\tabcolsep}{0.5em}
\renewcommand{\arraystretch}{1.1}
\begin{center}
{\small
    \begin{tabular}{@{}l|c|ccccc@{}}
        Method  & CS & Rain & Fog & Snow & Night & Avg.   \\ \hline
        Baseline & \textbf{70.5} & 44.2 & 58.7  &44.2 &  18.9 & 41.5 \\ \hline
        ISSA: CS-G-E & 70.3 & 50.6 & 66.1 & \textbf{53.3} & 30.2 & 50.1  \\
        ISSA: BDD-G-E & 70.3 & \textbf{52.2} & \textbf{66.3} & 52.2 & \textbf{31.0} & \textbf{50.4}
    \end{tabular}
}
\end{center} 
\caption{Comparison on Cityscapes to ACDC generalization using ISSA with generator and encoder trained on Cityscapes (CS-G-E) and BDD100K (BDD-G-E), respectively.
Despite never seeing Cityscapes samples, ISSA with BDD-G-E is still highly effective.
}
\label{tab:bddgan-cityscapes-dg-small}
\end{table}

%% file: tables/landscape_style_dg.tex
\begin{table}[t]
\begin{center}
{\small
    \begin{tabular}{@{}l|c|cccc@{}}
        Method & CS & ACDC & BDD & DarkZ   \\ \midrule
        Baseline & 70.47 & 41.48 & 45.66 & 15.50 \\
        \midrule
        ISSA: CS-G-E   & 70.30  & 50.05  & 50.29 & 27.24  \\
        ESSA: CS-G-E  & 69.85  & \textbf{50.87} & \textbf{51.42} & \textbf{29.06} \\
    \end{tabular}
}
\end{center}
\caption{Utilizing Landscape Pictures as extra-source exemplars for style augmentation, where the generator and encoder are only trained on Cityscapes (CS-G-E). ESSA can further improve the generalization performance from Cityscapes to other unseen datasets.
}
\label{tab:ladnscape-dg}
\end{table}

%% file: figs/performance_validation_id.tex
\begin{figure}[t]
\centering
\includegraphics[width=0.85\linewidth]{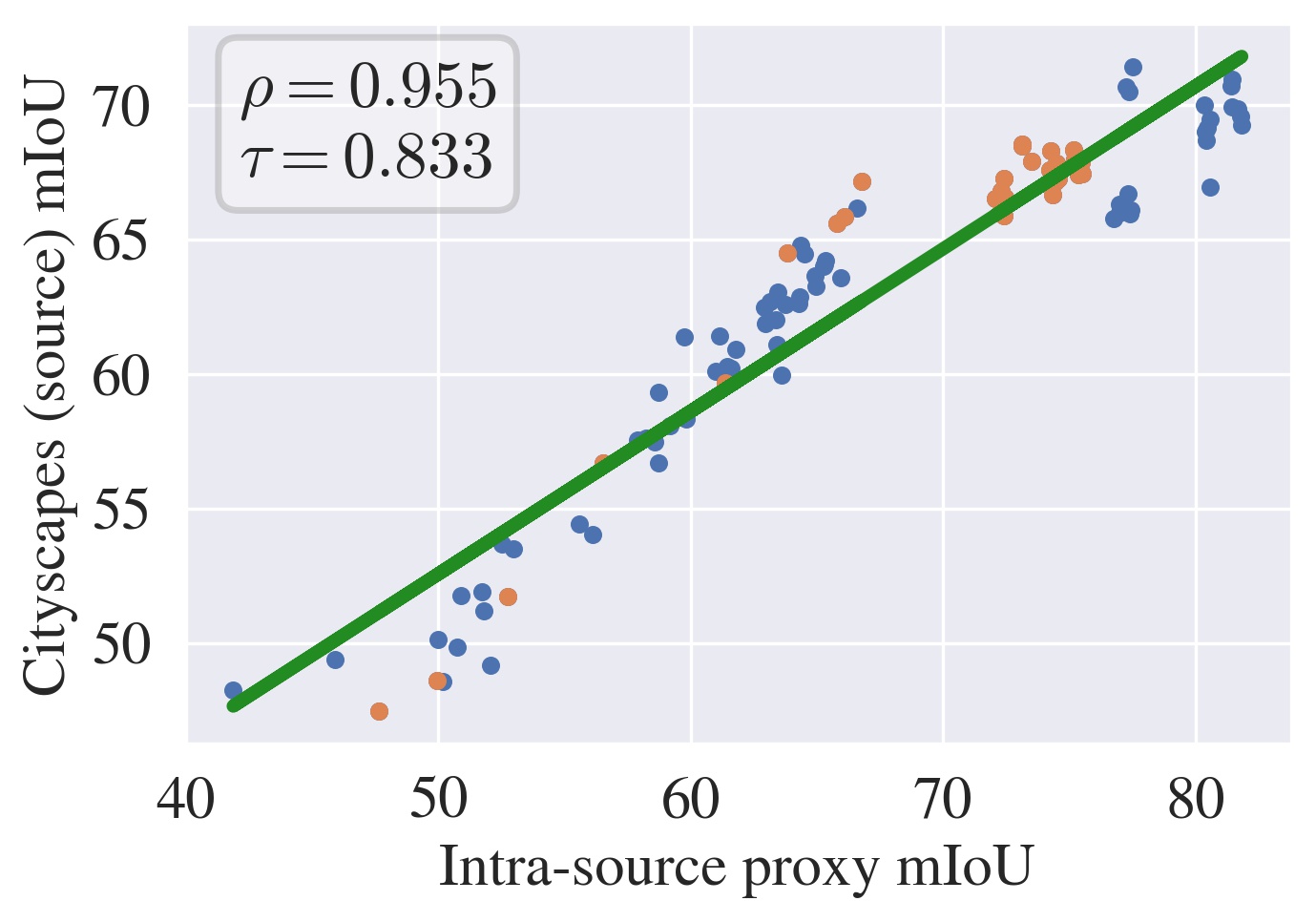}
\caption{Correlation between real Cityscapes  test performance and intra-source style augmented proxy performance for $95$ models.  Spearman’s Rank Correlation coefficient ($\rho$) and Kendall Rank Correlation Coefficient ($\tau$) are computed to quantitatively measure correlation strength. \textcolor{matplotlibBlue}{Blue} and \textcolor{matplotlibOrange}{orange} dots represent CNN- and transformer-based backbones, respectively. We observe that there is a strong correlation between the real test mIoU and proxy mIoU.
}
\label{fig:ID-correlation}
\end{figure}

%% file: figs/performance_validation.tex

\newlength{\twosubht}
\newsavebox{\twosubbox}
\newlength{\threesubht}
\newsavebox{\threesubbox}

\begin{figure*}[t]
\sbox\threesubbox{%
  \resizebox{\dimexpr.93\textwidth}{!}{%
    \includegraphics[height=2.9cm]{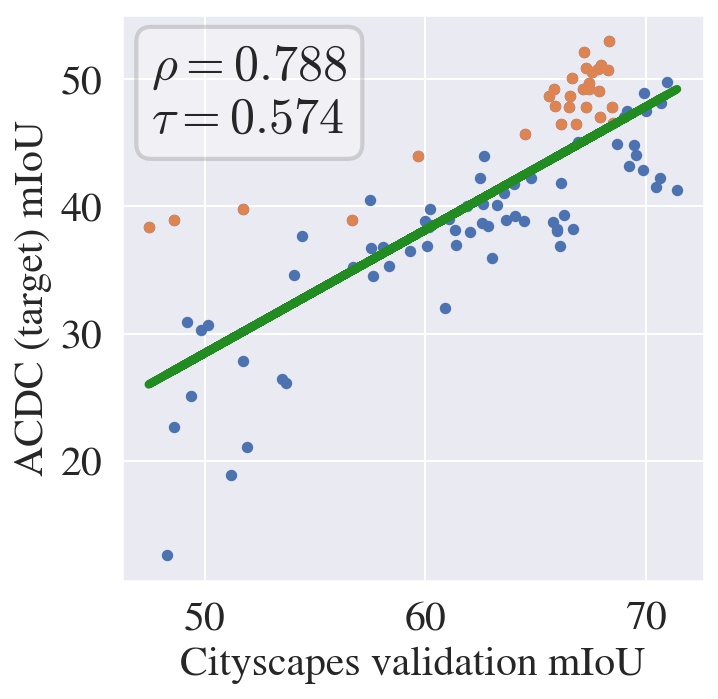}%
    \includegraphics[height=2.9cm]{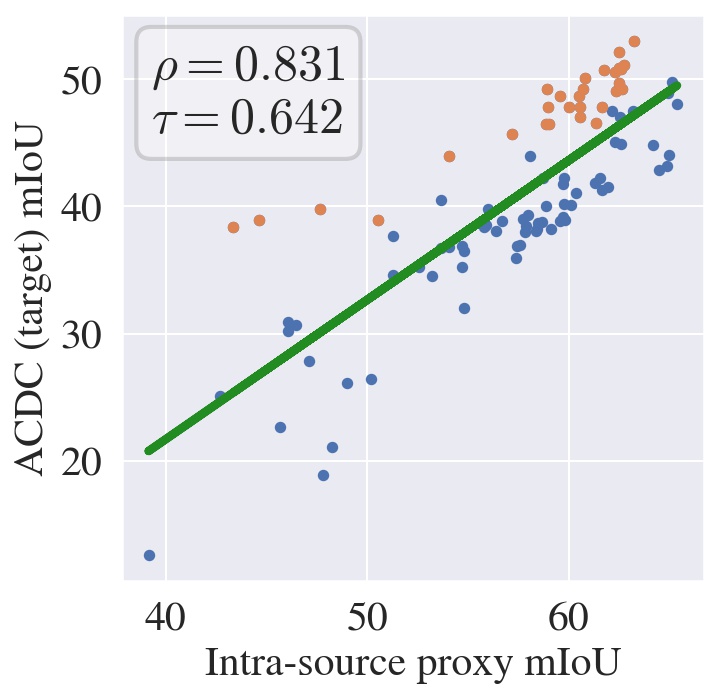}%
    \includegraphics[height=2.9cm]{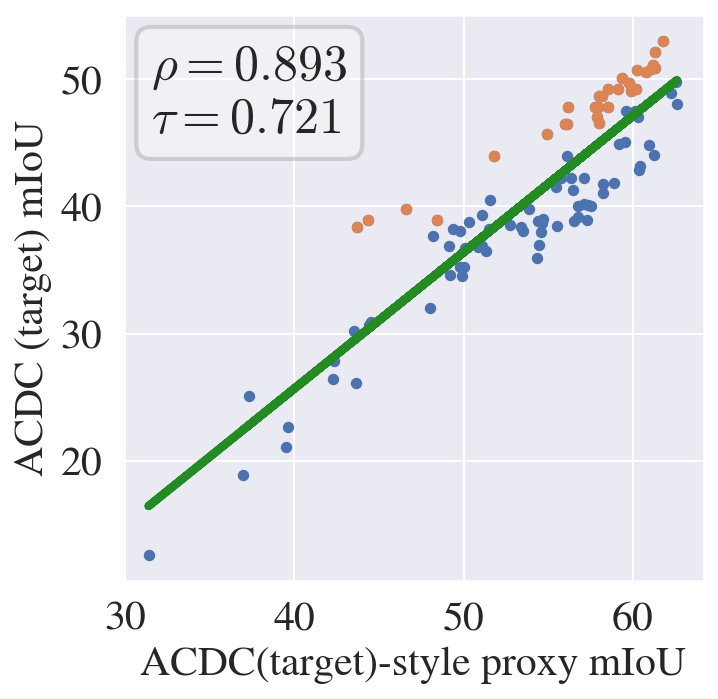}%
  }%
}
\setlength{\threesubht}{\ht\threesubbox}
\sbox\twosubbox{%
  \resizebox{\dimexpr.93\textwidth}{!}{%
    \includegraphics[height=3cm]{Intra-ACDC.jpg}%
    \includegraphics[height=3cm]{ACDC-ACDC.jpg}%
  }%
}
\setlength{\twosubht}{\ht\twosubbox}

\centering
\subcaptionbox{\label{csval-acdc}}{%
  \includegraphics[height=\threesubht]{CS-ACDC.jpg}
}\quad
\subcaptionbox{\label{intra-acdc}}{%
  \includegraphics[height=\threesubht]{Intra-ACDC.jpg}
}\quad
\subcaptionbox{\label{acdc-acdc}}{%
  \includegraphics[height=\threesubht]{ACDC-ACDC.jpg}
}
\\
\subcaptionbox{\label{csval-bdd}}{%
  \includegraphics[height=\threesubht]{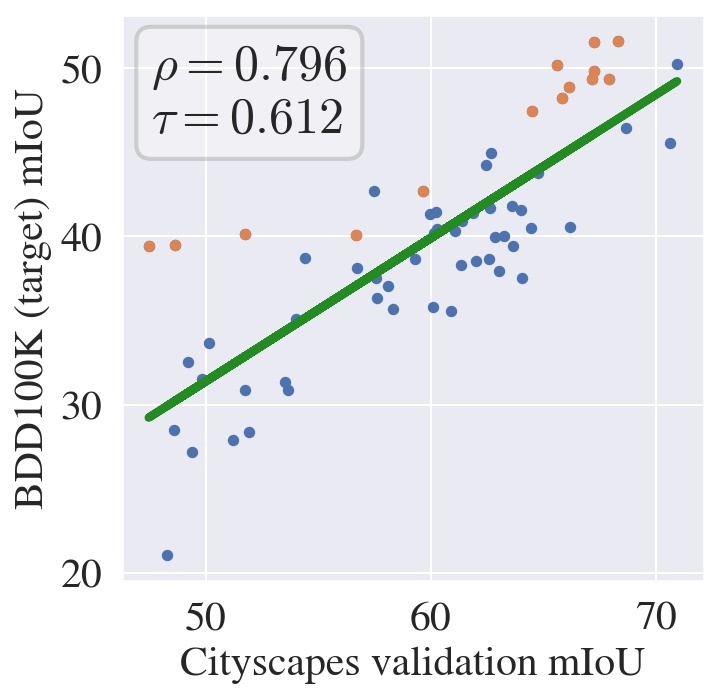}
}\quad
\subcaptionbox{\label{intra-bdd}}{%
  \includegraphics[height=\threesubht]{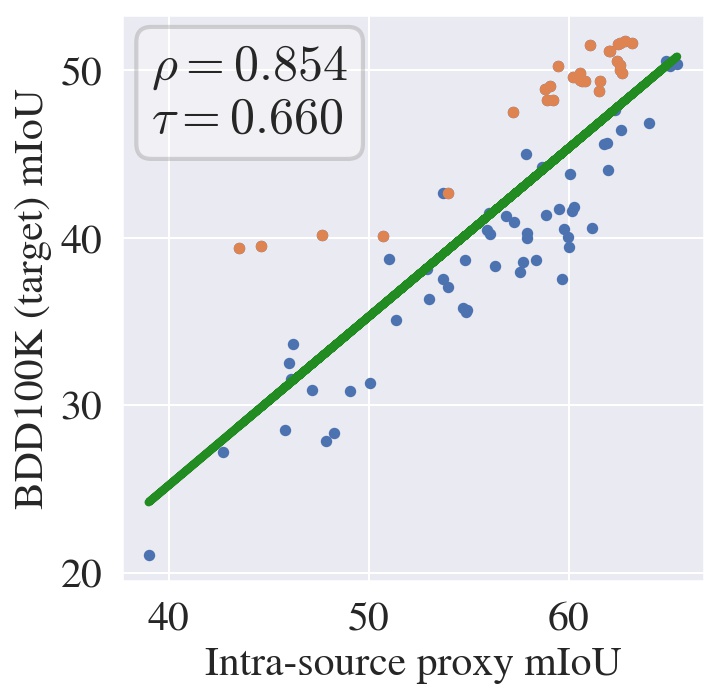}
}\quad
\subcaptionbox{\label{bdd-bdd}}{%
  \includegraphics[height=\threesubht]{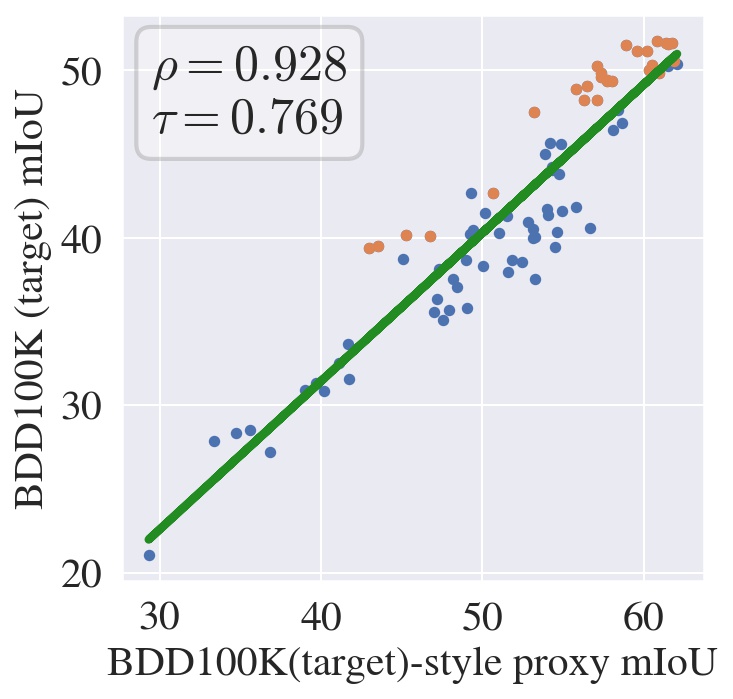}
}
\caption{Correlation between test performance and proxy performance for $95$ models. We compute Spearman’s Rank Correlation coefficient ($\rho$) and Kendall Rank Correlation Coefficient ($\tau$) to quantitatively measure correlation strength. \textcolor{matplotlibBlue}{Blue} and \textcolor{matplotlibOrange}{orange} dots represent CNN- and transformer-based backbones, respectively. In each row, we investigate the correlation between the real test performance, i.e., mIoU of ACDC and BDD100K, and mIoU of different proxy sets. We observe that \cref{acdc-acdc,bdd-bdd} achieve the strongest correlation for each scenario, which indicates that it is beneficial to build a proper proxy set using styles of the corresponding test dataset.
} 
\label{fig:performance_correlation}
\end{figure*}

%% file: tex/5_conclusion.tex
\section{Conclusion and Discussions}\label{sec:conclusion}

\newchange{In this paper, we propose a GAN inversion based style synthesis pipeline for domain generalization in semantic segmentation.
The key enabler for our pipeline is the masked noise encoder, which is capable of preserving fine-grained content details and allows style mixing between images without affecting the semantic content. 
In particular, we employ intra-source style augmentation ({\ourstyle}) for learning domain generalized semantic segmentation using restricted training data from a single source domain. }
Extensive experimental results verify the effectiveness of {\ourstyle} on domain generalization across different datasets and network architectures.
We further demonstrate the plug-n-play ability of the proposed pipeline. Without requiring retraining the encoder and generator, our model can be used directly on extra-source exemplars such as web-crawled images, enabling extra-source style augmentation (ESSA). It also opens up applications beyond data augmentation for improved domain generalization. Specifically, we show that the intra- \& extra-source exemplar-based style synthesis pipeline can be used for creating proxy validation sets to compare the generalization capability of diverse models on both the source and target domain without extra data annotation effort.

\paragraph{Limitation and future work} 
One limitation of {\ourstyle} is that our style mixing is a global transformation, which cannot specifically alter the style of local objects, e.g., adjusting vehicle color from red to black, though when changing the image globally, local areas are inevitably modified. Also compared to simple data augmentation such as color transformation, our pipeline requires higher computational complexity for training. It takes around $7$ days to train the masked noise encoder on $256 \times 512$ resolution using $2$ GPUs. A similar amount of time is required for the StyleGAN2 training. Nonetheless, for data augmentation, it only concerns the inference time of our encoder, which is much faster, i.e., 0.1 seconds, compared to optimization based methods such as PTI\newcite{roich2021pti} that takes 55.7 seconds per image.

In the future, it is challenging yet interesting to extend our work with more flexible local editing. Our proposed \newchange{intra- \& extra-source exemplar-based style synthesis} is a global transformation, which cannot specifically alter the style of local objects, e.g., adjusting vehicle color from red to black, though when changing the image globally, local areas are inevitably modified. 
One potential direction is by exploiting the pre-trained language-vision model, such as CLIP\newcite{radford2021clip}. We can synthesize styles conditioned on text rather than an image. For instance, by providing a text condition ``snowy road",  ideally we would want to obtain an image where there is snow on the road and other semantic classes remain unchanged. 
Recent works \newcite{bar2022text2live,hertz2022prompt2prompt,kawar2022imagic} studied local editing conditioned on text. However, CLIP exhibits a strong bias\newcite{bar2022text2live} and may generate undesirable results, and the edited image may suffer from insufficient alignment with the other parts of the image. Overall, there is still large room for improvement on  synthesizing images with more controls on both style and content.

\paragraph{Data Availability}
The datasets analysed during the current study are available at \href{https://www.cityscapes-dataset.com/}{Cityscapes}, \href{https://acdc.vision.ee.ethz.ch/}{ACDC}, \href{https://www.bdd100k.com/}{BDD100K}, \href{https://www.trace.ethz.ch/publications/2019/GCMA_UIoU/}{Dark Z\"urich}, \href{https://www.kaggle.com/datasets/arnaud58/landscape-pictures?resource=download}{Landscape Pictures} repository, respectively.